\documentclass[Afour,sageh,times]{sagej}

\usepackage{moreverb,url}
\usepackage{graphicx,amsmath,amssymb,array,fancyhdr,color,sidecap,paralist,eurosym,nicefrac,lastpage}
\usepackage{bm}
\usepackage{booktabs}

\usepackage[draft,colorlinks,bookmarksopen,bookmarksnumbered,citecolor=red,urlcolor=red]{hyperref}
\usepackage[linesnumbered,ruled]{algorithm2e}
\usepackage{subfigure}
\newcommand{\q}{\bm{q}}
\newcommand{\Q}{\bm{Q}}
\newcommand{\expnumber}[2]{{#1}\mathrm{e}^{#2}}
\newcommand{\up}{\tilde{p}}
\newcommand{\Dkl}[2]{D_{\mathrm{KL}}(#1||#2)}

\newcommand{\pmvn}{\mathcal{N}}
\newcommand{\pmn}{\mathcal{MN}}
\newcommand{\pbmf}{p_{\mathrm{BMF}}}
\newcommand{\vect}[1]{\mathrm{vec}(#1)}

\newcommand{\trans}[2]{\mathcal{T}_{#2}^{#1}}
\newcommand{\logm}[2]{\mathrm{Log}_{#2}(#1)}

\newcommand{\trsp}{{\scriptscriptstyle\top}}

\newcommand{\s}{\bm{s}}
\newcommand{\samples}[1]{\hat{#1}}

\newcommand{\xd}{\hat{\bm{p}}}

\def\BState{\State\hskip-\ALG@thistlm}

\newcommand\BibTeX{{\rmfamily B\kern-.05em \textsc{i\kern-.025em b}\kern-.08em
		T\kern-.1667em\lower.7ex\hbox{E}\kern-.125emX}}

\def\volumeyear{2020}

\runninghead{Pignat et al.}

\title{Learning from demonstration using products of experts: applications to manipulation and task prioritization}

\author{Emmanuel Pignat\affilnum{1}\affilnum{2}, Jo\={a}o Silv\'{e}rio\affilnum{1} and Sylvain Calinon\affilnum{1}\affilnum{2}}

\affiliation{\affilnum{1}Idiap Research Institute, Martigny, Switzerland.\\
	\affilnum{2}EPFL, Lausanne, Switzerland.}

\corrauth{Emmanuel Pignat,
	Idiap Research Institute,
	Rue Marconi 19, 
	1920 Martigny,
	Switzerland
}

\email{emmanuel.pignat@idiap.ch}
\begin{document}

	\begin{abstract}
		Probability distributions are key components of many learning from demonstration (LfD) approaches. 
		While the configuration of a manipulator is defined by its joint angles, poses are often best explained within several task spaces.
		In many approaches, distributions within relevant task spaces are learned independently and only combined at the control level.
		This simplification implies several problems that are addressed in this work. 
		We show that the fusion of models in different task spaces can be expressed as a product of experts (PoE), where the probabilities of the models are multiplied and renormalized so that it becomes a proper distribution of joint angles.
		Multiple experiments are presented to show that learning the different models jointly in the PoE framework significantly improves the quality of the model. 
		The proposed approach particularly stands out when the robot has to learn competitive or hierarchical objectives.
		Training the model jointly usually relies on contrastive divergence, which requires costly approximations that can affect performance.
		We propose an alternative strategy using variational inference and mixture model approximations.
		In particular, we show that the proposed approach can be extended to PoE with a nullspace structure (PoENS), where the model is able to recover tasks that are masked by the resolution of higher-level objectives.
	\end{abstract}

	
	\maketitle
	
	\section{Introduction}
	\label{sec:intro}
	\begin{figure*}
		\centering
		\includegraphics[width=1.\linewidth]{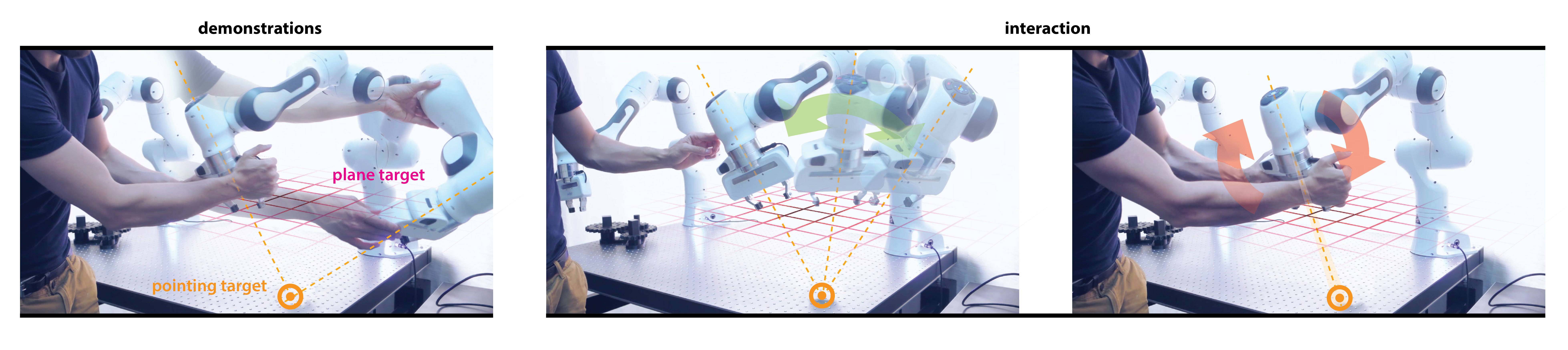}
		\caption{Product of experts can be used to set up virtual guides on robotic manipulators. A user demonstrates kinesthetically several configurations fulfilling the task objectives (\emph{left}). In this example, the end-effector of the robot should stay close to a horizontal plane, while pointing to a desired target. The objectives are then inferred by the robot as target distributions under several transformations (task spaces). By exploiting the torque control capabilities of recent robots, the different objectives are tracked. Using an optimal control strategy, feedback gains can be computed according to the precision of the different objectives. As a resulting behavior, the robot is free to move along directions that are not constrained by the objectives (\emph{center}), while preventing deviations from the objectives (\emph{right}).}
		\label{fig:poe_manipulator}
	\end{figure*}
	
	Adaptability and ease of programming are key features necessary for a wider spread of robotics in factories and everyday assistance.
	Learning from demonstrations (LfD) is an approach to address this problem. 
	It aims to develop algorithms and interfaces such that a non-expert user can teach the robot new tasks by showing examples. 
	In LfD, as well as in other robot learning techniques, probability distributions are a key component of the proposed models.
	
	In this work, we propose a framework to represent and learn distributions of robot configurations $p(\q)$, which can be used within many LfD models.
	As the desired configurations should be adapted to external parameters $\s$, such as position and orientation of objects, we also address the problem of learning conditional distributions $p(\q | \s)$. As acquiring data by manipulating the robot is often costly, we focus on problems where only small datasets are provided and in which generalization capabilities with respect to the external parameters $\s$ are important.
	
	In contrast, many of the current works in machine learning rely on big datasets. It enables complex distributions to be learned with little or no prior knowledge about the structure of the data. Our approach aims to exploit at best the existing robotic knowledge, while keeping the possibility to learn more complex distributions. Following Occam's razor principle, our approach aims to find simple explanations for complex distributions. We show that simpler explanations not only lead to more interpretable models but also increase the generalization capabilities and reduce the need for data.
	
\begin{figure}
		\centering
		\includegraphics[width=1.\linewidth]{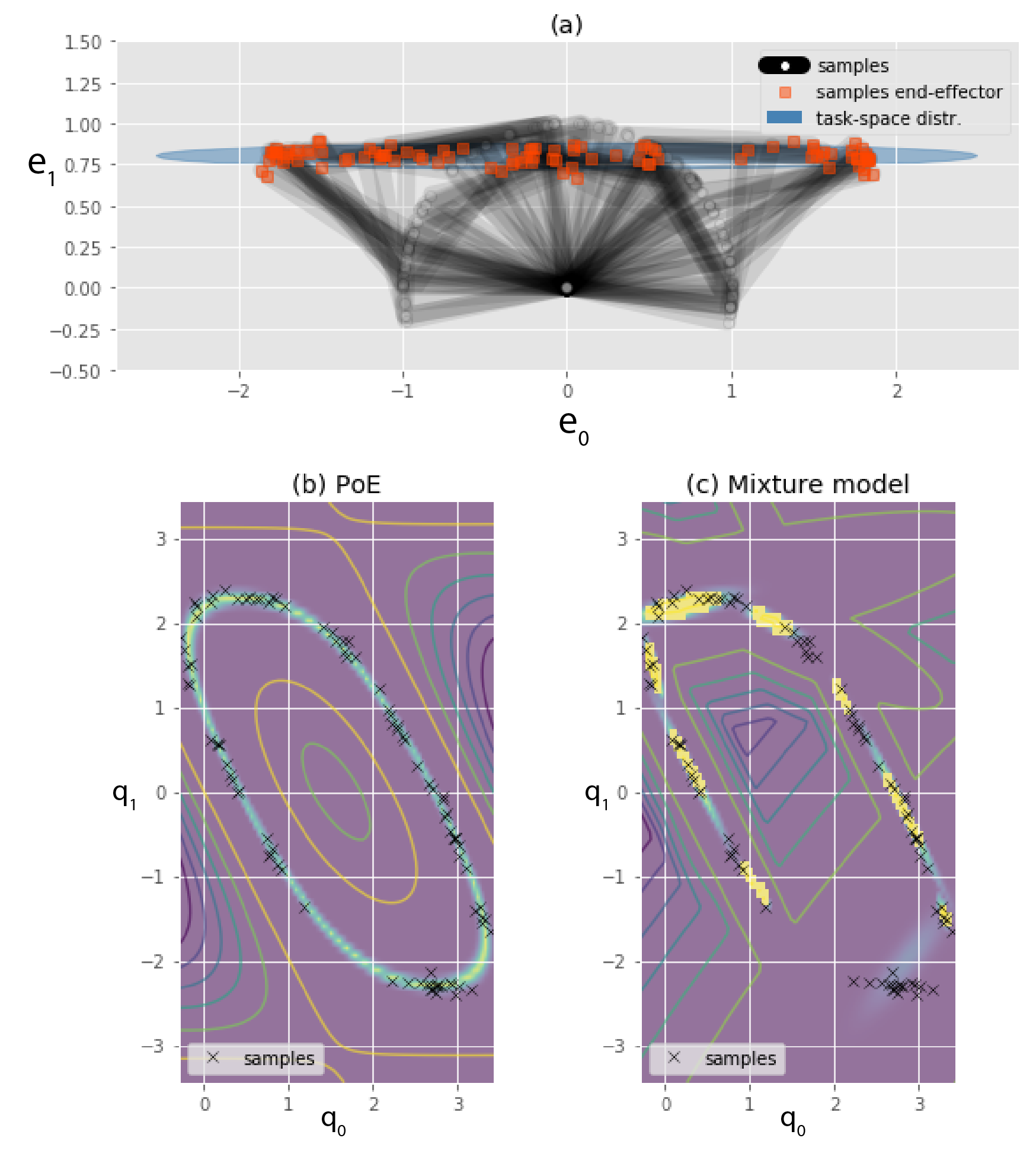}
		\caption{
			Product of experts can learn complex distributions of joint angles as a fusion of simple distributions within several task spaces.
			\textit{(a)} 30 samples of the configuration of a 2-DoF planar robot are displayed. The end-effector follows an elongated Gaussian distribution. The distribution in configuration space has a more complex shape, which it is difficult to represent. \textit{(b)} With our approach, we learn configuration-space distributions as a product of simple densities in different task spaces. It is thus possible to represent this sharp distribution easily with few parameters. \textit{(c)} A standard approach like a Gaussian mixture model, directly applied in configuration space, needs much more samples and parameters to represent this distribution. Following Occam's razor principle, finding simpler explanations (fewer parameters of the model) to understand complex dataset leads to better generalization.
		}
		\label{fig:poevsgmm}
	\end{figure}

	In robotics, these explanations often correspond to transformations providing distributions of simpler shapes. For example, a Gaussian distribution of the end-effector might result in a very complex distribution of joint angle configurations, as shown in Fig.~\ref{fig:poevsgmm}. The configuration $\q$ is often not of primary interest; poses in different task spaces, distances to objects, pointing directions or bimanual correlations are often more important. Hence, many approaches in LfD only learn distributions of these transformed quantities. These distributions are often learned independently and combined only at the control level.
	In this article, we show that this approach has several drawbacks. First, it is unaware of the kinematic structure of the robot and of the limited range of values that can be reached within each task space. It results in a confusion between the characteristics of the task and the capabilities of the robot. Secondly, when distributions under several task spaces are learned independently, the relation between them is ignored. It then becomes impossible to properly understand each task and the required precision. Moreover, when some tasks are prioritized, secondary objectives can only be recovered if the dependencies between the task spaces are considered. They are indeed masked by the resolution of the tasks of higher importance.
	
	The main contribution of this work is to apply the products of experts (PoE) approach in \cite{hinton1999products} to robotics. This approach can combine the interpretability, compactness and precision of task-space distributions with the kinematic awareness of the configuration space. Particularly, the main detailed contributions are: 
\begin{enumerate}
	\itemsep-0.em 
	\item \textbf{Training PoEs (Section \ref{sec:poe})} An approach to train PoEs using variational inference, better suited to the targeted robotic applications than the original approach from \cite{hinton2002training}.
	\item \textbf{PoE with prioritization (Section \ref{sec:poe_nullspace})} A novel technique that leverages the PoE formulation in combination with null space operators to learn task priorities from demonstrations with minimal prior knowledge compared to the state of the art (e.g. no need to know task references a priori, recover secondary masked tasks).
	\item  \textbf{New perspectives for PoEs in robotics (Section \ref{sec:experts}, \ref{sec:applications})} An extensive formulation of experts describing various relevant manipulation skills in robotics (e.g. position/orientation/joint space/manipulability, movement primitives, ergodic control, prioritization) whose fusion can be harmoniously learned using our approach.
	\item \textbf{Experiments (Section \ref{sec:experiments})} A detailed analysis of the proposed approach, highlighting the capabilities of learning from few data, as well as the ability to learn an arbitrary number of prioritized tasks.
\end{enumerate}


\paragraph{\textbf{Organization of the article}}
The article starts with an overview of the related work in Section \ref{sec:intro}. After presenting the general approaches to learn distributions, we concentrate on PoE models and their applications. The use of distributions in robotics and its relation with inverse optimal control are then discussed.

In Section \ref{sec:poe}, PoEs are formally presented and the practical implications of the normalization constant for robotics are discussed. In the second part of the section, a method to approximate and train PoEs using variational inference is proposed.

In Section \ref{sec:poe_nullspace}, we propose an extension of PoEs with nullspace filter (PoENS). Classical nullspace approaches for inverse kinematic are first presented. Then, this approach is used to redefine the derivative of the log-likelihood of the PoE such as to induce a hierarchy. 

Several distributions and transformations are presented in Section \ref{sec:experts}. They help the practitioner to tackle a wide range of robotic problems. 

In Section \ref{sec:applications}, two different control strategies compatible with PoEs are then presented (Fig.~\ref{fig:poe_manipulator} illustrates one of them). 

Finally, in Section \ref{sec:experiments}, several experiments are presented to compare the proposed models with other density estimation techniques.

	\subsection{Related work}
	\label{sec:related_work}
	\paragraph{\textbf{Estimating distributions}}
	Estimating probability density functions have been for long a major field in statistics and machine learning. Estimating complex densities using a sum of simpler densities has been proposed with kernels in \cite{parzen1962estimation} or mixture models in
	\cite{dempster1977maximum}. Unfortunately, these techniques can be very inefficient in high-dimensional spaces, such as the configuration of a robot. With the rising popularity of neural networks, deep generative models have been used to learn distributions of higher dimensions, such as images. Restricted Boltzmann machine (\cite{hinton2006fast}) and deep Boltzmann machine \cite{salakhutdinov2009deep} are popular models. Like products of experts (PoE), proposed in \cite{hinton1999products}, these models are trained by maximizing an intractable likelihood function. Several approximations are necessary to compute the gradient of the log-likelihood.
	
	These complex approximations motivated the development of various models that do not represent the likelihood but allow sampling from the distribution like generative networks \cite{bengio2014deep}, generative adversarial nets (GAN) \cite{goodfellow2014generative} and variational autoencoders (VAE) \cite{kingma2013auto}. Unlike previous techniques, they do not require approximations when computing the gradient of the cost to optimize. However, once the model is trained, computing the likelihood is either impossible or requires approximations. These generative machines have been very popular recently and are used to learn complex, high-dimensional distributions from huge datasets. 
	
	Models with unnormalized likelihood like PoEs are of interest in robotics for inverse optimal control (IOC) problems, see e.g., \cite{finn2016guided}. In \cite{ziebart2008maximum}, the maximum entropy principle was introduced for IOC as an intractable density on paths. However, the direct use of PoEs has been largely overlooked in robotics. Compared to modern generative models, it might appear at first sight as a step backward. However, the direct use of PoE has multiple advantages the will be highlighted in this article. For the learning process, it enables to use of robotics-related knowledge to decrease the amount of data required. Particularly, we show that the computational overhead induced by the approximation in the gradient is largely compensated by the advantages of these models in terms of small datasets requirements. Also, for control applications, the direct access to the (at least unnormalized) likelihood is important. Moreover, these computational overheads are very limited when used with small datasets; most of the training procedures considered in the experiments in Sec.~\ref{sec:experiments} take between 10 seconds to a minute.
	
	\paragraph{\textbf{Product of experts}}
	Products of experts have been proposed in \cite{hinton1999products} as an alternative to mixture models to compensate for their poor efficiency in high-dimensional space. The combination of the distributions (called experts) is achieved by a product instead of a summation, which provides much sharper distributions. If computing the normalizing constant of a sum of distributions is straightforward, the major problem with PoEs is to compute this quantity and its derivative, which requires approximate methods. Sampling from a PoE is also more difficult than from a mixture model.
	
	Many works have used the term "product of experts" to express the fusion of several models. Many of these are not considering the joint training of the models as originally proposed in \cite{hinton2002training}. For example, two long short-term memory (LSTM) models are combined in \cite{johnson2017learning} to generate jazz melodies. In robotics, the fusion of multiple sources of data (e.g. sensors) was expressed as a PoE in \cite{pradalier2003expressing} for the localization of mobile robots and obstacle avoidance. In \cite{calinon2016tutorial}, the fusion of trajectory models learned independently in multiple coordinate systems is also referred to as a product of Gaussians (PoG).
	
	Jointly training the experts, as proposed in \cite{hinton2002training}, has been used in several applications.
	In robotics, a product of contact models is proposed in \cite{kopicki2017learning} to predict the motion of manipulated objects.
	In speech processing, PoEs have been used for vowel classification \cite{dixon2006speech}. In \cite{zen2012product}, speech sequences are modeled using a product of multiple acoustic models. These acoustic models consist of transformations, both linear (discrete cosine transform, summation) and non-linear (quadratic), by using various distributions such as Gaussian, Gamma or log-Gaussian.
	The authors claim that distortions of the generated speech are often due to the different acoustic models being learned separately and only combined later, at the synthesis stage.
	The authors proposed to use the PoE framework, enabling the training of multiple acoustic models cooperatively. Our claim is that, for robotics applications, learning the different models separately and combining them later, also induce heavy distortions. Many works learn different models (in configuration space, in task space, manipulability measures, ...) separately and only combine them at the control stage. In this article, we show the advantages of learning these models collaboratively in a way that the dependencies between the different features are kept, while taking the kinematic structure of the robot into account.
	
	\paragraph{\textbf{Distributions for learning in robotics}}
	In Learning from demonstrations (LfD), as well as in other robot learning techniques, probability distributions are a key component of the proposed models. They are often used to define motions by introducing a dependence to time \cite{calinon2009statistical}, as observation models in hidden Markov models \cite{calinon2016tutorial}, or transformed by a time-dependent basis matrix \cite{paraschos2017probabilistic}. Most of these works learn the different models separately and combine them at the control stage.
	In \cite{calinon2009statistical}, objectives from multiple task spaces and configuration space are considered. Multiple time-dependent Gaussian distributions of end-effector positions and joint angles are learned separately, with an assumption of independence. These competing objectives are only combined at the control stage, by exploiting a product of Gaussians in the velocity space.
	In \cite{calinon2016tutorial}, multiple Gaussian models of motion are learned independently within different task spaces. Using properties of linear transformations and product of Gaussians, the different models are combined in closed-form before synthesizing sequences.  
	In \cite{zeestraten2017approach}, this approach is extended to also encode orientation in multiple coordinate systems.
	In \cite{paraschos2017probabilistic}, the authors propose a combination of multiple Gaussian distributions of trajectories (ProMP). The different models are also defined both in configuration space and in different task spaces, again learned separately. They are then combined at the control level as acceleration commands.

	Similarly, in \cite{silverio2018learning}, we proposed to fuse multiple independent Gaussian models in different task spaces and configuration space as a product of Gaussians. The fusion is approximated locally using a linearization of the forward kinematics. We also proposed to learn a hierarchy of tasks using linear transformations with nullspace filters. That approach is limited to velocity commands and tasks where the target is known. The approach that we now propose can cope with static configurations samples, which requires more advanced tools to take into account the non-linearities. The different tasks do not need to be known beforehand and our method can uncover secondary tasks that are masked by primary ones.
	The secondary tasks can also be discovered in \cite{Towell10} and \cite{Lin15}. However, one needs to know the control variables during the demonstrations, which are not always easily accessible. In our approach we can rely on static configurations solely.
	
	In \cite{niekum2015learning}, segments of trajectories are analyzed in different coordinate systems, in a similar fashion to \cite{calinon2016tutorial}. A unique coordinate system is then selected for each segment, by looking at the similarity between end-points. This idea of selecting relevant task spaces is explored in several other works. 
	In \cite{muhlig2009automatic}, the demonstrations are projected in a set of task spaces. Different criteria, such as the variance of the data and the attention of the demonstrator, are used.
	In \cite{lober2015variance}, the variance of several distributions learned separately is mapped to weights modulating the task prioritization, where the variances are learned from several demonstrations. 
	
	A common point of these works is that the different models are learned independently and their prioritization or importance is related to their variance. Indeed, secondary tasks do exhibit a higher variance but the experiments of this article will show the necessity to distinguish between the variance and the prioritization. Indeed, we will demonstrate that competing tasks can be understood only if the models are trained jointly, especially to understand the characteristics of secondary tasks.
	
	Outside the field of LfD, learning distributions of robot configurations is also a topic of interest for sampling-based path planning \cite{amato1996randomized}. Sampling-based path planning methods require to sample valid configurations (e.g. collision-free) to connect them and create paths. Randomly sampling the configuration space and keeping valid samples is not efficient because of a possible low acceptance rate. Some works focus on learning the distribution to sample from.
	In \cite{ichter2018learning}, conditional distributions are learned using a conditional variational autoencoder \cite{sohn2015learning}. The distribution is conditioned on some external factors to provide adaptation, such as obstacles occupancy grids.
	In \cite{lehner2017repetition}, the authors use a similar approach with Gaussian mixture models learned from previous collision-free configurations.

	\paragraph{\textbf{Inverse optimal control}}
	Our approach shares similarities with inverse reinforcement learning (IRL) or inverse optimal control (IOC) \cite{ng2000algorithms}. In these approaches, the robot is trying to infer the cost function shown by the expert demonstrator. Maximum entropy inverse reinforcement learning has been proposed in \cite{ziebart2008maximum}. It frames the problem as a maximum likelihood estimation of an intractable (unnormalized) density over trajectories $\bm{\tau}$
	\begin{align}
	p(\bm{\tau}) \propto \exp\big(-c_{\bm{\theta}}(\bm{\tau})\big).
	\end{align}
	Our approach can be interpreted as inferring a cost that is minimized by inverse kinematics. In our case, the multiple costs in the different task spaces are defined by the log-likelihood of expert distributions. Instead of ``PoE'', our work could also have been called ``inverse inverse kinematics''.
	
	Learning cost functions for inverse kinematics was already discussed in \cite{kalakrishnan2013learning}. However, their approach is limited to learning a weight vector of different cost features (joint limits, manipulability measure, elbow position).
	In \cite{jetchev2011task}, the cost is learned on a feature vector of the proposed task space. Sparsity is ensured using $\ell_1$ regularization which allows retrieving and selecting important task spaces.
	These two latter approaches formulate the problem as inferring a cost that acts on different task spaces. The relation between the different task spaces and the dependencies between the different tasks are better taken into account than in the probabilistic models presented in the previous paragraph.
	
	In \cite{finn2016guided}, a more expressive cost function is considered by using neural networks. While our approach considers simpler static tasks, we propose more complex approximations and a hierarchical structure. With a more structured and interpretable cost, our approach also targets smaller datasets.
	
	\section{Product of Experts}
	\label{sec:poe}
	Products of experts, proposed in \cite{hinton1999products}, are models in which several densities $p_m$ (which are called experts) are multiplied together and the product renormalized. Each expert can be defined on a different view or transformation of the data $\trans{}{m} (\q)$, and the resulting density expressed as
	\begin{align}
	p(\q|\bm{\theta}_1,\ldots,\bm{\theta}_M) &= 
	\frac{\prod_m p_m(\trans{}{m} (\q)\,| \bm{\theta}_m)}
	{\int_z \prod_m p_m(\trans{}{m} (\bm{z})\,| \bm{\theta}_m)}.
	\end{align}
	For compactness, we will later refer to the unnormalized product as 
	\begin{align}
	\up(\q) &= 
	\prod_m p_m(\trans{}{m} (\q)\,| \bm{\theta}_m),
	\end{align}
	where we drop the parameters of the experts $\bm{\theta}_1,\ldots,\bm{\theta}_M$ in the notation.
	In this work, the transformations $\trans{}{m} (\q)$  correspond to different task spaces. They can either be given, such as the forward kinematics of a known manipulator, or parametrized (subject to estimation). Several transformations that can be used in robotic problems are presented in Sec.~\ref{sec:transformations}.
	
	As an example, $\q$ can be the configuration of a humanoid (joint angles and a floating base 6-DoF transformation). The different transformations can consist of the forward kinematics of several links, like the feet and the hands. The densities of the corresponding experts can be multivariate Gaussian distributions, defining where these links should be located. 
	
	This way of combining distributions is very different from more traditional mixture models. In mixture models, each distribution is in configuration space, which can be of high dimensions ($>30$) for humanoids. Many components and a laborious tuning are necessary to cover this space. Often, the distribution is also very sharp, creating a low-dimensional manifold in a higher-dimensional space, as illustrated in Fig.~\ref{fig:poevsgmm}. These sharp distributions are hard to be represented with mixtures, even in low-dimensional spaces. In PoEs, each expert can constrain a subset of the dimensions or a particular transformation of the configuration space. The product is the intersection of all these constraints, which can be very sharp. It has far fewer parameters to train than a mixture. A PoE density is also smoother than a mixture, which is by definition multimodal. This feature is particularly beneficial for applications in control, as it will be presented in Sec.~\ref{sec:applications}.
	Moreover, if the model needs to encode time-varying configurations, it can easily be extended to a mixture of PoEs. Great flexibility is then allowed to set up such model; some experts can be shared among mixture components to encode global constraints (preferred posture, static equilibrium), while some others can have different parameters for each mixture component.
	
	\subsection{Importance and implications of the normalization constant}
	The renormalization is an important aspect that distinguishes this approach from the ones in which the models are learned independently. In these approaches, the recorded data is first transformed into the quantities of interest $\trans{}{m} (\q)$, or directly recorded in this form. Then, either the different quantities are stacked to learn one model, or different models on the different quantities are trained separately. In the latter case, the models are combined at the control level. This approach corresponds to a PoE in which the normalization constant has been dropped. The renormalization might seem to be a mathematical preoccupation, but it has several practical implications.
	\begin{itemize}
		\item Renormalizing the product allows some experts to give constant probabilities to all configurations. If the normalization was done experts-wise, this would imply close to zero probability everywhere, and result in a very low likelihood. Experts which do not influence the product can be dropped without penalizing the overall likelihood. Hence, explaining the data with sparse models (a small number of experts) is possible. It often leads to better generalization. In Sec.~\ref{sec:experiments:multimodal_distribution}, an experiment is presented in which the model should distinguish between a configuration-space or a task-space target distribution.
		\item When using a mixture of PoEs, as in \cite{calinon2016tutorial}, not considering the renormalization further reduces generalizations capabilities. It prevents mixture components to identify patterns appearing only in a subset of the task spaces, which would have been sufficient to explain the full configuration. Instead, the mixture components are allocated such that the representation is compact under each transformation, preventing good generalization capability. 
		\item The proper support (set of possible values that $\trans{}{m}(\q)$ can take) of each expert distribution is taken into account. Renormalizing on $\q$ makes sure that the support considered by each expert is defined by the set of possible configurations and its transformation in the different task spaces. For example, let us consider an expert as a non-degenerate multivariate Gaussian defining the position of the end-effector of a fixed manipulator. The support of a Gaussian is normally $\mathbb{R}^3$. By using a PoE, which takes into account $\q$ and the forward kinematics transformation, the support turns into the actual workspace of the robot.
		
		When training the model, it means that the high probability regions of each expert are not necessary where the data is. With a proper support, the model is trained such that the transformed data is in a region of higher probability than the remaining values that it can take.
		Practically, it means that the model can clearly distinguish between the targeted task, represented by the PoE, and the kinematic capability of the robot. The importance of the support is best noticed when transferring models between robots with different kinematic chains, or when tasks should be understood even if their realization is prevented by kinematic constraints.
		\item The realization of tasks is sometimes prevented by complementary or competitive objectives. In Sec.~\ref{sec:poe_nullspace}, an extension to PoE is presented to admit strict hierarchies between the tasks.
		As with kinematic constraints, considering the proper support and renormalization allows recovering the masked tasks. This possibility will be illustrated by several experiments in Sec.~\ref{sec:experiments:hierarchical_tasks}, also shown in Fig.~\ref{fig:bimanualnullspacecomp}.
		\item The use of unnormalized expert densities is enabled. Especially for orientation statistics, some interesting distributions have intractable normalizing constant, which sometimes dissuade their use. In a PoE, these distributions can be used with no overhead, as the normalization only occurs at the level of the product. Also, the normalization constant is typically easier to compute on joint angles than on orientation manifolds. An experiment is presented in Sec.~\ref{sec:experiments:pos_rot_correlation} in which a joint distribution of positions and rotation matrices is learned.
	\end{itemize}
	When the experts are learned independently, nothing ensures that the PoE matches the data distribution. It only becomes similar in some particular cases. For example, when the demonstrated data has an important variance under all-but-one transformations. Moreover, no kinematic constraint should prevent the resolution of the task (e.g., a task-space target distribution with all its mass inside the robot workspace).

	\subsection{Estimating PoE}
	\subsubsection{Markov Chain Monte Carlo (MCMC)}
	In the general case, the renormalized product $p(\q)$ has no closed-form expression.
	A notable exception is Gaussian experts with linear transformations. In this case, the product $p(\q)$ is Gaussian. Many techniques such as Kalman filtering or linear quadratic tracking (LQT) can be reformulated as a product of Gaussians under linear transformations. In our case, this is of limited interest, due to the nonlinearities of the transformations we are considering.
	
	Approximation methods are required to work with PoEs. An adequate approximation method should enable several elements: we should be able to draw samples from it, to estimate the normalizing constant, to estimate its gradient with respect to the model parameters, and to identify the modes of the distribution.
	Such methods are found in Bayesian statistics, where the same problem of approximating an unnormalized density occurs. Indeed, in Bayesian statistics, the posterior distribution of model parameters $\bm{\theta}$ can be viewed as a product of two experts: a likelihood and a prior.
	
	Markov chain Monte Carlo (MCMC), see \cite{andrieu2003introduction} for an introduction, is a class of methods to approximate $p(\q)$ with samples. If MCMC methods can represent arbitrarily complex distributions, they suffer from some limitations. 
	
	For example, they are known not to scale well to high-dimensional spaces, which is particularly constraining for our application. A lot of samples are required to cover high-dimensional distributions.
	The key component of many MCMC methods is the definition of a proposal move, which makes the random walker (chain) more likely to visit areas of high probabilities. The design of this proposal move is algorithmically restrictive, because it needs to ensure that the distribution of samples at equilibrium is proportional to the unnormalized density. Furthermore, it is difficult to obtain good acceptance rates in high dimension, especially with very correlated $\up(\q)$, as shown in Fig.~\ref{fig:poevsgmm}. 
	
	By representing the distribution only through samples, it is also difficult to assess if this latter is well covered. A huge part of the space and distant modes might remain undiscovered. It is also difficult to know the granularity of the approximated distribution.
	
	Except for some particular approaches, such as \cite{chen2014stochastic}, MCMC methods require an exact evaluation of $\up(\q)$. In contrast, stochastic variational inference (SVI) only requires a stochastic estimate of the gradient of $\log\up(\q)$. There are many advantages to this unconstraining requirement. Experts transformations that are too costly to compute exactly can be approximated. If the PoE is conditional, batches of conditional values can be used. Also, the gradient can be redefined such as a hierarchy between the task can be set up, as will be presented in Sec.~\ref{sec:poe_nullspace}.
	
	Finally, MCMC methods struggle with multimodal distributions. Chains are unlikely to cross big regions of small density. If multiple chains are run in parallel, the respective mass of each mode is difficult to assess. A proper approach of this problem is to design particular proposal steps to move between distant modes, as proposed in \cite{sminchisescu2003mode}, which is algorithmically restrictive.
	
	\newcommand{\approxq}{\tilde{q}}
	\subsubsection{Variational inference}
	\label{sec:variational_inference}
	Variational inference (VI) (\cite{wainwright2008graphical}) is another popular class of methods that recasts the approximation problem as an optimization. VI approximates the \textit{target density} $\up(\q)$ with a tractable density $\approxq(\q; \bm{\lambda})$ called the \textit{variational density}. $\bm{\lambda}$ are the \textit{variational parameters} and are subject to optimization. A density is called tractable if drawing samples from it is easy and that the density is properly normalized. VI tries to minimize the intractable KL-divergence between the renormalized density $p(\q)$ and the variational density $\approxq(\q; \bm{\lambda})$
	\begin{align}
	\Dkl{\approxq}{p} = \int_{\q} \approxq(\q; \bm{\lambda}) \log \frac{\approxq(\q; \bm{\lambda})}{p(\q)} d\q.
	\end{align}
	Given that $p(\q) = \up(\q)/\mathcal{C}$ where $\mathcal{C}$ is the normalizing constant, we can rewrite the previous divergence as
	\begin{align}
	\Dkl{\approxq}{p} = \int_{\q} \approxq(\q; \bm{\lambda}) \log \frac{\approxq(\q; \bm{\lambda})}{\up(\q)} d\q + \log \mathcal{C},
	\label{equ:vi_kl_cost}
	\end{align}
	which can be evaluated up to a constant and minimized. The first term is the negative \textit{evidence lower bound} (ELBO). This term can be estimated by sampling as
	\begin{align}
	\mathcal{L}(\bm{\lambda}) &= \int_{\q} \approxq(\q; \bm{\lambda}) \log \frac{\approxq(\q; \bm{\lambda})}{\up(\q)} d\q\\
	\label{equ:elbo_1}
	&= \mathbb{E}_{\approxq}[\log \approxq(\q; \bm{\lambda}) - \log \up(\q)]\\
	&\approx \frac{1}{N}\sum_{n=1}^{N}\big(\log \approxq(\q^{(n)}; \bm{\lambda})-\log \up(\q^{(n)}) \big),\\
	\notag
	&\mathrm{with}\quad \q^{(n)} \sim \approxq(\,\cdot\,|\bm{\lambda}).
	\end{align}
	The reparametrization trick proposed in \cite{salimans2013fixed} and \cite{ranganath2014black}
	allows a noisy estimate of the gradient $\mathcal{L}(\bm{\lambda})$ to be computed. It is compatible with stochastic gradient optimization like Adam \cite{kingma2014adam}.
	For example, if $\approxq$ is Gaussian, this is done by sampling $\bm{\eta}^{(n)}\sim \mathcal{N}(\bm{0}, \bm{I})$ and applying the continuous transformation $\q^{(n)} = \bm{\mu} + \bm{L}\bm{\eta}^{(n)}$, where $\bm{\Sigma} = \bm{L}\bm{L}^\trsp$ is the covariance matrix. $\bm{L}$ and $\bm{\mu}$ are the \textit{variational parameters} $\bm{\lambda}$. More complex mappings as normalizing flows can be used as in \cite{rezende2015variational}.
	
\begin{figure}
		\centering
		\includegraphics[width=1.\linewidth]{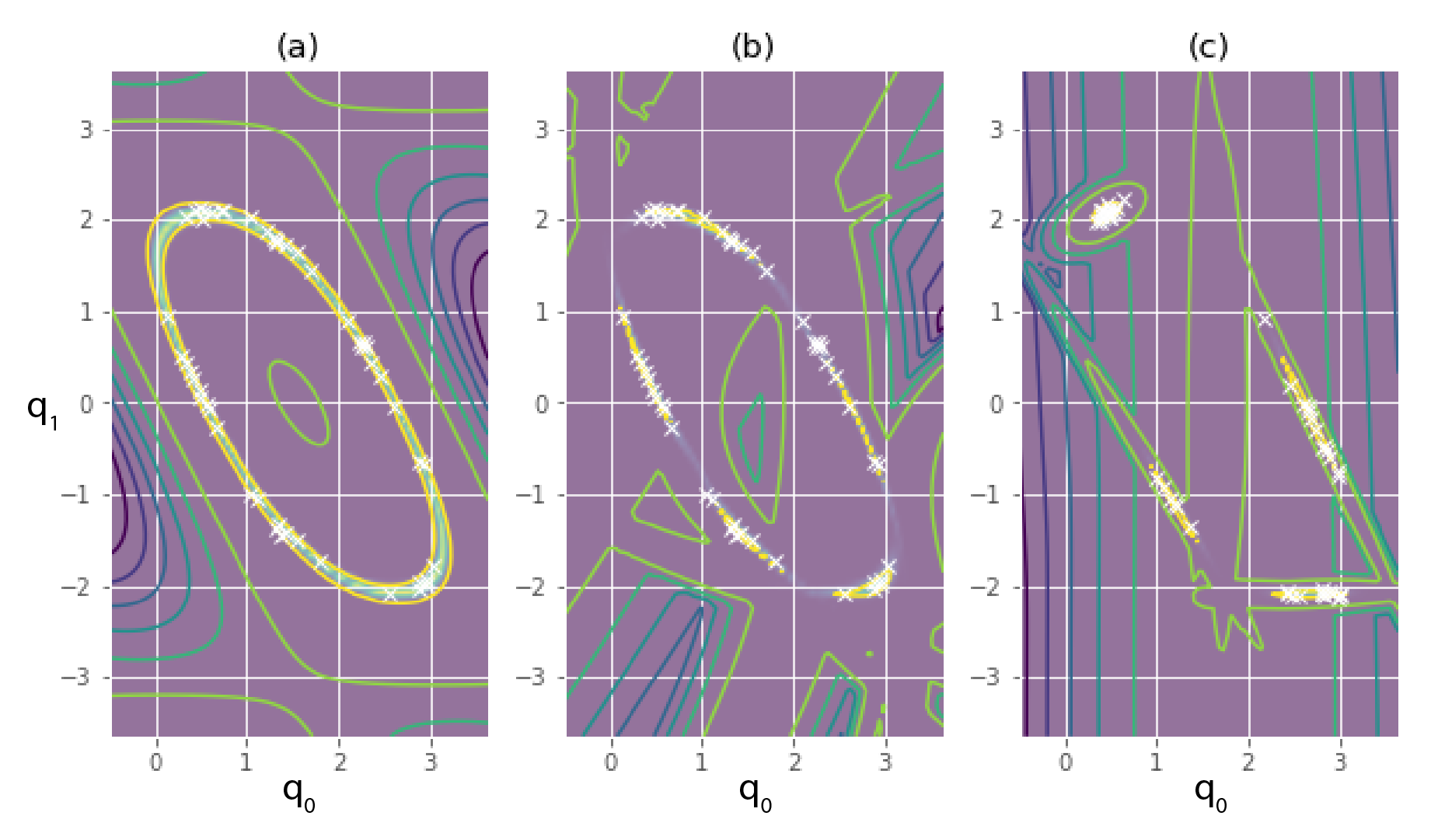}
		\caption{The product of experts is an unnormalized density. As in Fig.~\ref{fig:poevsgmm}, the density shown in \textit{(a)} results from an elongated Gaussian defined on the end-effector of a 2-DoF planar robot. For control applications, such as keeping the robot in this distribution, accessing the unnormalized density is sufficient. For other applications, such as training a PoE, sampling from this distribution, evaluating the normalizing constant and its gradient might be needed. Using variational inference, the PoE can be approximated by a tractable distribution, as a mixture of banana-shaped distributions \textit{(b)} or Gaussians \textit{(c)}. On purpose, only 5 mixture components were used in the illustrations. A more precise approximation can be achieved with a higher number.}
		\label{fig:estimatingpoe2d003}
	\end{figure}

	\paragraph{Zero forcing properties of minimizing $\Dkl{q}{p}$}
	\label{par:zero_avoid}
	It is important to note that due to the objective $\Dkl{\approxq}{p}$, $\approxq$ is said to be zero forcing.
	If $\approxq$ is not expressive enough to approximate $\up$, it would miss some mass of $\up$ rather than giving a high probability to locations where there is no mass (see Fig.~\ref{fig:estimatingpoe2d003} for an illustration).
	
	\paragraph{Mixture model variational distribution}
	For computational efficiency, the \textit{variational density} $\approxq(\q; \bm{\lambda})$ is often chosen as a factorized distribution, using the mean-field approximation (\cite{wainwright2008graphical}). 
	Correlated distributions can be approximated by a Gaussian distribution with full covariance as in \cite{opper2009variational}. 
	These approaches fail to capture the multimodality and arbitrary complexity of $\up(\q)$. The idea to use a mixture for greater expressiveness as \textit{approximate distribution} was initially proposed in \cite{bishop1998approximating}, with a recent renewal of popularity \cite{miller2017variational,guo2016boosting,arenz2018efficient}.
	
	A mixture model is built by summing the probability of $K$ mixture components
	\begin{align}
	\approxq(\q\arrowvert\bm{\lambda}) = \sum_{k=1}^{K} \pi_k\, \approxq_k(\q\arrowvert\bm{\lambda}_k), \quad \sum_{k=1}^{K} \pi_k = 1,
	\label{equ:mm}
	\end{align}
	where $\pi_k$ is the total mass of component $k$. The components $\approxq_k$ can be of any family accepting a continuous and invertible mapping between $\bm{\lambda}$ and the samples.
	The discrete sampling of the mixture components according to $\pi_k$ has no such mapping. Instead, the variational objective can be rewritten as
	\begin{align}
	\mathcal{L}(\bm{\lambda}) &=\mathbb{E}_{\approxq}[\log \approxq(\q; \bm{\lambda}) - \log \up(\q)]\\
	&= \sum_{k=1}^{K} \pi_k \mathbb{E}_{\approxq_k}[\log \approxq(\q; \bm{\lambda}) - \log \up(\q)],
	\end{align}
	meaning that we need to compute and get the derivatives of expectations only under each component distribution $\approxq_k(\q\arrowvert\bm{\lambda}_k)$.
	
	\subsection{Training PoEs}
	\label{sec:training_poe}
	From a given dataset of robot configurations $\Q$, maximum likelihood (or maximum a posteriori) of the intractable distribution $p(\q|\bm{\theta}_1,\ldots,\bm{\theta}_M)$ should be computed. 
	
	It can be done using gradient descent, as proposed in \cite{hinton1999products}. The derivative of the log-likelihood of the PoE can be separated into the derivative of the unnormalized expert and the normalizing constant
	\begin{align}
	\begin{split}
	\frac{\partial \log p(\q|\bm{\theta}_1,\ldots,\bm{\theta}_M)}{\partial \bm{\theta_m}} =
	&{}\frac{\partial\log p_m(\q|\bm{\theta}_m)}{\partial \bm{\theta_m}} -\\
	&{}\frac{\partial \log \mathcal{C}(\bm{\theta}_1,\ldots,\bm{\theta}_M)}{\partial \bm{\theta}_m}.
	\label{equ:poe_gradient}
	\end{split}
	\end{align}
	The derivative of the normalizing constant is intractable and requires approximation methods. It can be written as
	\begin{align}
	\begin{split}
	\frac{\partial \log \mathcal{C}(\bm{\theta}_1,\ldots,\bm{\theta}_M)}{\partial \bm{\theta}_m}&{}= \\
	\int_c p(\bm{c}|&{}\bm{\theta}_1,\ldots,\bm{\theta}_M) \, \frac{\partial\log p_m(\bm{c}|\bm{\theta}_m)}{\partial \bm{\theta_m}} \, d\bm{c},
	\label{equ:poe_norm_gradient}
	\end{split}
	\end{align}
	which is the expected derivative of the unnormalized expert log-likelihood under the current PoE distribution. 
	The averaged derivatives over the dataset $\Q$ and with respect to the parameters of all experts $\bm{\theta} = \begin{bmatrix}
	\bm{\theta}_1^\trsp,\ldots,\bm{\theta}_M^\trsp
	\end{bmatrix}^\trsp$ can be written as
	\begin{align}
	\Big<\frac{\partial \log p(\q)}{\partial \bm{\theta}} \Big>_{\Q}
	&= \Big<\frac{\partial\log \up(\q)}{\partial \bm{\theta}}\Big>_{\Q}\, -
	\Big<\frac{\partial\log \up(\q)}{\partial \bm{\theta}}\Big>_{p(\q)},
	\label{equ:poe_dataset_gradient}
	\end{align}
	where $<\cdot>_{p(\cdot)}$ denotes the expectation over the distribution $p$.
	Intuitively, it means that we compare the expected gradient of the unnormalized density over the dataset $\Q$ with the expected gradient over the current density $p(\q)$. At convergence, the expected difference should be zero; it means that the distribution of the PoE $p(\q)$ matches the data distribution $\Q$.
	Maximizing the log-likelihood of the data is also equivalent to minimizing the KL divergence between the data distribution and the equilibrium distribution.	
	
	In all but a few cases, the expectation under the current PoE density has no closed-form. Worse, estimating this integral with samples is not trivial since drawing samples from the current PoE is difficult. In \cite{hinton2002training}, it is proposed to use a few sampling steps initialized at the data distribution $\Q$. These chains have not reached their equilibrium distribution (the PoE) but can still provide a biased estimation of the gradient. Unfortunately, this approach fails when $p(\q|\bm{\theta}_1,\ldots,\bm{\theta}_M)$ has multiple modes. The few sampling steps never jump between modes, resulting in the incapacity of estimating their relative mass. Wormholes have been proposed as a solution in \cite{welling2004wormholes}, but are algorithmically restrictive.
	
	As an alternative, we propose to use VI with the proposed mixture distribution to approximate $p(\q)$. Throughout the training process, this approximation is kept updated to be able to draw samples from the PoE. The process thus alternates between minimizing $\Dkl{\approxq}{p}$ with current $p$ and using current $\approxq$ to compute the gradient \eqref{equ:poe_gradient}. The approximate distribution $\approxq$ can either be used as an importance sampling distribution or directly (if expressive enough to represent $p$). The expected gradient over the current density $p(\q)$ becomes
	\begin{align}
	\Big<\frac{\partial\log \up(\q)}{\partial \bm{\theta}}\Big>_{p(\q)} \approx \sum_{n=1}^{N} w_n \frac{\partial\log \up(\q^{(n)})}{\partial \bm{\theta}}/\sum_{n=1}^{N} w_n\\ 
	\mathrm{with}\quad \q^{(n)} \sim \approxq(\,.\,|\bm{\lambda}) \quad \mathrm{and} \quad w_n = \frac{\up(\q^{(n)})}{\approxq(\q^{(n)}|\bm{\lambda})}
	\end{align}
	where $w_n$ are the importance weights.
	
	It is also possible to estimate this expectation by using a mixture of samples from the variational distribution $q$ and from Markov chains initialized on the data. It combines the advantages of the two methods. Samples from $q$ are necessary to estimate the relative mass of distant modes, while the few steps of the Markov chain reduce the variance of the derivative. Empirically, we noticed that they tend to stabilize the learning procedure.
	
	In the derivation, only the experts parameters $\bm{\theta} = \begin{bmatrix}
	\bm{\theta}_1^\trsp,\ldots,\bm{\theta}_M^\trsp
	\end{bmatrix}^\trsp$ were trained. It is also possible to train the parameters of their associated transformation without any modification to the procedure.
	
	\paragraph{Initializing PoEs}
	As the training procedure is iterative, a good initialization of $\bm{\theta_m}$ speeds up the learning process. For the experts with simple forms of maximum likelihood estimation (MLE) and known transformation $\trans{m}{}$, we propose independent initializations with maximum likelihood. As detailed in the related work, many works in robotics train the model independently. Thus, we propose to initialize the PoE that way and improve it by considering the proper renormalization.

	\section{Product of Experts with nullspace filter (PoENS)}
	\label{sec:poe_nullspace}
	We often encounter tasks where some objectives are more important than others. The problem of controlling a robot when the different objectives and their hierarchy are known has already been addressed in several works, see e.g. \cite{nakamura1987task}.
	It is usually done by filtering commands minimizing secondary objectives with nullspace projection operators. 
	It ensures that these commands stay within the subspace of commands minimizing higher level objectives.
	
	In our work, considering a hierarchy has an additional interest. The realization of some subtasks might be prevented and masked by the resolution of higher-level subtasks. 
	Thus, it is not trivial to identify secondary objectives in the dataset.
	Ignoring their lower priority when training the model will lead to problems (at best, minimizing their importance, and at worst, completely missing them). 
	These tasks can be understood only by considering their lower priority and their entanglement with higher priority subtasks.
	To this end, we propose to extend the PoE framework to the use of nullspace filters. 

	Illustratively, the idea is to define a hierarchy between the experts such that secondary experts can express their opinions only in subspaces that primary experts do not care about. 
	For example, let us consider that $\q \in \mathbb{R}^7$ are the joint angles of a 7-DoF manipulator and $p_1(\trans{}{1}(\q))$ is a Gaussian distribution of the end-effector position. The system has still 4 DoFs in which secondary experts can express their opinion.
	
	Our approach consists of redefining the derivative of the log-likelihood of the PoE with a nullspace filter. This filter cancels the derivative of secondary experts in the space of primary experts, making sure that secondary experts have no power in the area of expertise of primary experts.
	Each expert acts on a transformation of the configuration $\q$
	\begin{align}
	\bm{y}_m = \trans{}{m}(\q),
	\end{align}
	and have the differential relationship
	\begin{align}
	\dot{\bm{y}}_m = \bm{J}_m(\q)\,\dot{\q},
	\label{equ:diff_jac}
	\end{align}
	where $\bm{J}_m(\q) = \frac{\partial \trans{}{m}(\q)}{\partial \q}$ is the Jacobian matrix of the transformation $\trans{}{m}$. When computing inverse kinematics with a priority, as in \cite{nakamura1987task}, the general solution of \eqref{equ:diff_jac} is
	\begin{align}
	\dot{\q} = \bm{J}_m^{\dagger}(\q)\ \dot{\bm{y}}_m + \big(\bm{I} - \bm{J}_m^{\dagger}(\q)\bm{J}_m(\q)\big)\bm{z},
	\end{align}
	where $\bm{J}_m^{\dagger}$ is the pseudoinverse of $\bm{J}_m$ and $\bm{z}$ is an arbitrary vector. The nullspace filter  $\bm{N}_m(\q)=\bm{I} - \bm{J}_m^{\dagger}(\q)\bm{J}_m(\q)$ makes sure that the arbitrary vector has no effect in the transformation $\trans{}{m}$ as 
	\begin{equation*}
	\bm{J}_m(\q)\big(\bm{I} - \bm{J}_m^{\dagger}(\q)\bm{J}_m(\q)\big)\,\bm{z} 
	= \big(\bm{J}_m(\q)- \bm{J}_m(\q)\big)\,\bm{z}
	= \bm{0}.
	\end{equation*}
	
	By applying the chain rule, the derivative of the log-likelihood with respect to the configuration $\q$ becomes
	\begin{align}
	\frac{\partial \log \up(\q|\bm{\theta}_1,\ldots,\bm{\theta}_M)}{\partial \q} 
	&= \sum_m \frac{\partial \log p(\trans{}{m}(\q)|\bm{\theta}_m)}{\partial \q}\nonumber\\
	&= \sum_m \frac{\partial \log p(\bm{y}|\bm{\theta}_m)}{\partial \bm{y}}  
	\bm{J}_m(\q).
	\end{align}
	
	The nullspace filter $N_m(\q)$ is used to filter the derivative coming from other experts such that they do not affect the space where the expert $m$ is acting.
	For example, if we have two experts, with expert $m=1$ the primary the derivative becomes
	\begin{flalign}
	\frac{\partial \log p(\trans{}{m}(\q)|\bm{\theta}_1, \bm{\theta}_2)}{\partial \q} \qquad
	= \notag \\
	\frac{\partial \log p(\bm{y}|\bm{\theta}_1)}{\partial \bm{y}}  
	\bm{J}_1(\q) + \frac{\partial \log p(\bm{y}|\bm{\theta}_2)}{\partial \bm{y}} \bm{J}_2(\q) \bm{N}_1(\q). 
	\end{flalign}

	\begin{figure}
		\centering
		\includegraphics[width=1.\linewidth]{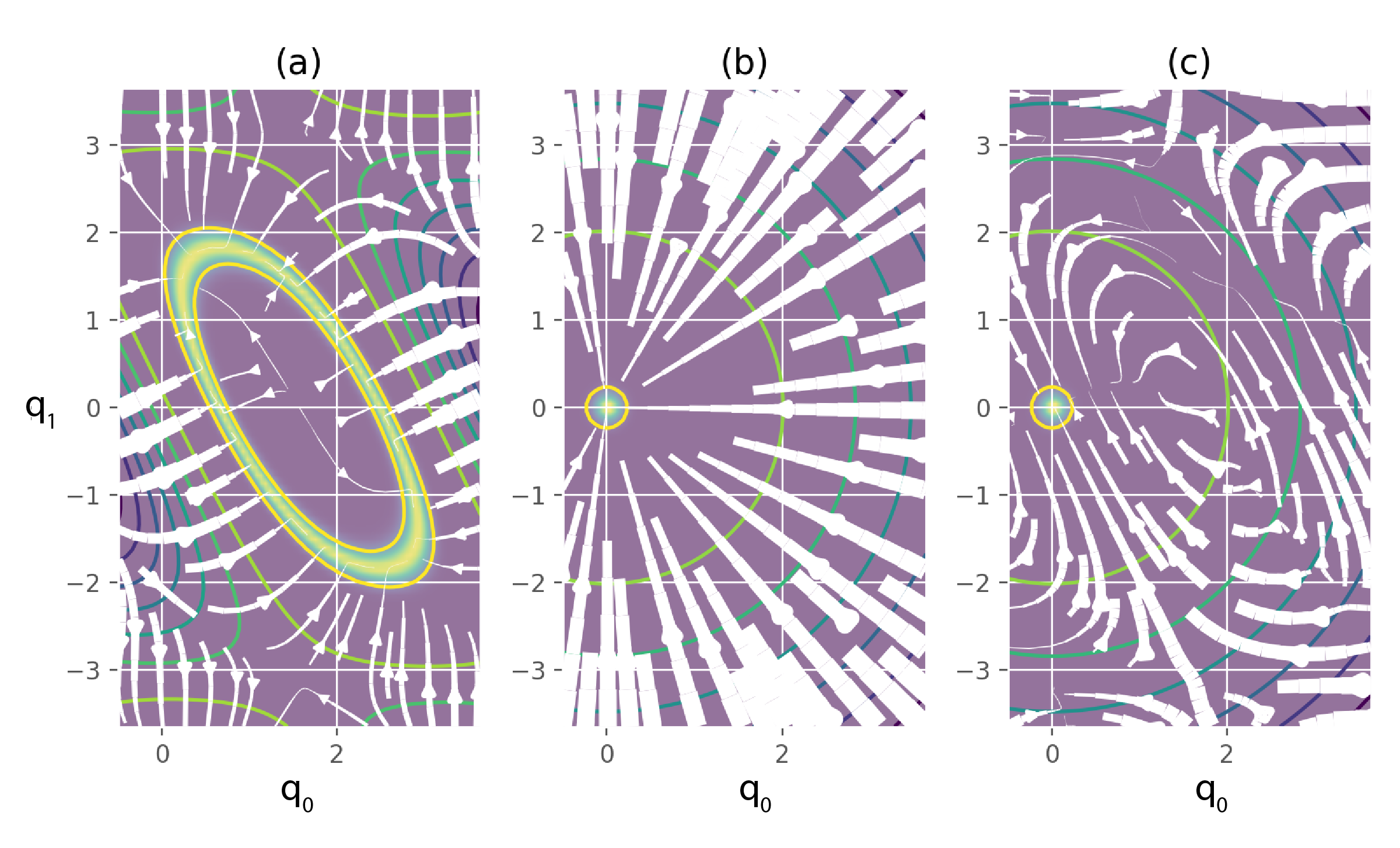}
		\vspace{-1.2em}
		\caption{Densities and derivatives of a two-expert problem. A 2-DoF planar robot is considered as in Fig.~\ref{fig:poevsgmm}. The first expert is an elongated Gaussian distribution on its end-effector. The second expert acts in joint angles and defines a preferred configuration. \textit{(a)} The density of the expert $\log p_1(\trans{}{1}(\q)|\bm{\theta}_1)$ is shown as a colormap. The derivative $\partial \log p(\bm{y}|\bm{\theta}_1)/\partial \bm{y}\,
			\bm{J}_1(\q)$ is displayed as streamlines.
			\textit{(b)} The same is done with the second objective in configuration space.
			\textit{(c)} The derivative of the second experts is this time filtered as $\partial \log p(\bm{y}|\bm{\theta}_2)/\partial \bm{y}\, \bm{J}_2(\q) \bm{N}_1(\q)$. }
		\label{fig:nullspace_derivative}
	\end{figure}
	
	Fig.~\ref{fig:nullspace_derivative} provides an example where the terms of this derivative are displayed with and without the nullspace filter for the secondary objectives.
	We note that when using automatic differentiation libraries such as TensorFlow \cite{abadi2016tensorflow}, gradients can be easily redefined with this filter.
	
	While in standard PoEs, it was possible to evaluate the unnormalized log-likelihood, the PoENS is defined only by the gradient of this quantity. It is thus not possible to evaluate the unnormalized log-likelihood $\up$. It constrains the class of methods to approximate the density. Stochastic variational inference can be employed, as it only requires stochastic evaluation of the gradient. This characteristic is shared with only a very few Monte Carlo methods such as \cite{chen2014stochastic}. In this mini-batch variant of Hamiltonian Monte Carlo method \cite{duane1987hybrid}, no corrective Metropolis-Hastings steps are used as they are too costly.
	
\begin{figure}
		\centering
		\includegraphics[width=1.\linewidth]{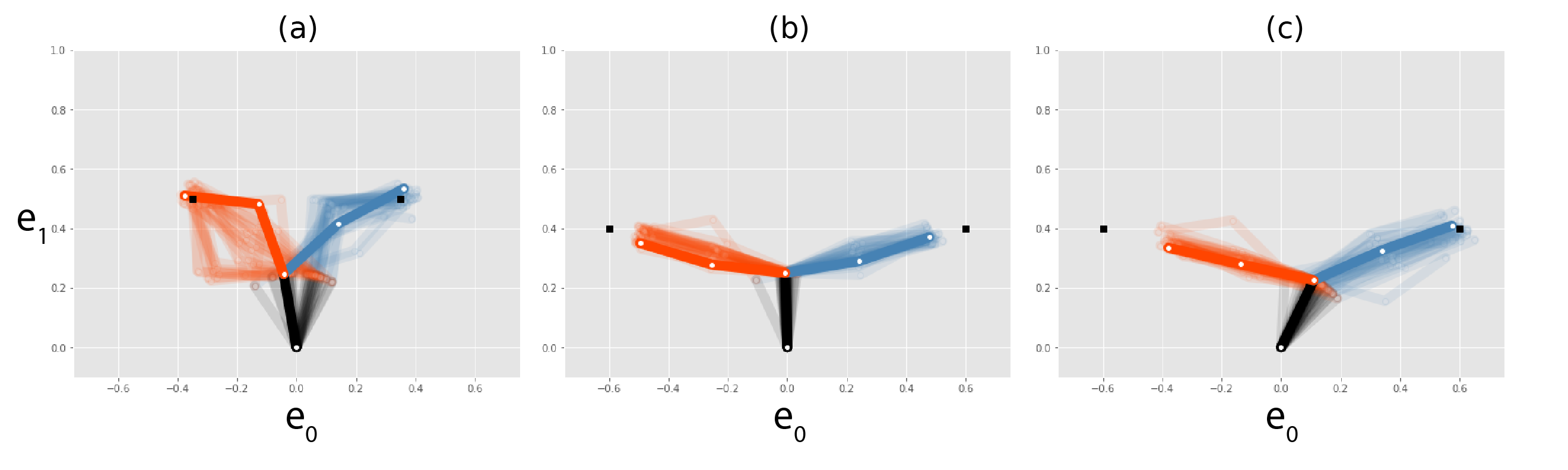}
		\vspace{-1.2em}
		\caption{5-DoF bimanual planar robot with two forward kinematics objectives. Variational inference is used to generate the displayed samples. \textit{(a)} The two tasks are compatible and the distribution of solution is approximated. \textit{(b)} No nullspace, the two tasks are of the same importance. \textit{(c)} The gradient of the objective of the orange arm is projected onto the nullspace of the Jacobian of the forward kinematics of the blue arm. This is achieved with stochastic variational inference that only requires to evaluate the gradient of the unnormalized density. }
		\label{fig:nullspace}
	\end{figure}

	Fig.~\ref{fig:nullspace} shows a 5-DoF bimanual planar robot with two forward kinematics objectives. When the tasks are compatible, the filtering does not affect the solutions. The gradient of the objective of the orange arm can be projected onto the nullspace of the Jacobian of the forward kinematics of the blue arm, resulting in a prioritization.

	\section{Useful distributions and transformations for robotics}
	\label{sec:experts}
	In this section, several transformations $\trans{m}{}$ and experts models $p_m$ related to common robotic problems are presented with a practitioner perspective. This can be used as a toolkit to unify various problems into the PoE framework.
	
	\subsection{Transformations}
	\label{sec:transformations}
	Several common transformations are presented. These transformations can be known and fixed, as the forward kinematics of the end-effector. They can also be partially known, for example the position of an object held by the known end-effector, which constitutes a new end-effector. Fully unknown transformations can also be learned with neural networks.
	
	\paragraph{\textbf{Forward kinematics (FK)}} One of the most common transformations used in robotics is forward kinematics, computing poses (position and orientation) of links given the robot configuration $\q$. Forward kinematics can be computed in several task spaces associated with objects of interest, as in \cite{calinon2016tutorial,niekum2015learning,muhlig2009automatic}. In these works, analyzing movements from several coordinate systems allows for generalizations with respect to movements of their associated objects. These works only consider cases where the transformation is fully known. The approaches in these works, based on separately transforming the data and learning the models, are not compatible with partially known transformations, as opposed to ours. In particular, two types of unknowns can be considered with PoE: unknown coordinate systems or free parameters in the kinematic chain. These two cases will be considered in the experiments of Sec.~\ref{sec:experiments:generalization}.
	
	In the first case, the robot could, for example, have to track an object that can move. We could have a dataset split into several subparts in which the pose of this object is constant. The unknown displacement of the objects can be subject to optimization, as well as the parameters of the distribution representing the target within its associated coordinate system.
	
	The second case can occur if a new end-effector is added. For example, it can be a tool grasped by the robot, whose position is given by
	\begin{align}
	\trans{}{m}(\q)= \bm{F}_{R}(\q) \bm{d} + \bm{F}_{x}(\q),
	\end{align}
	where $\bm{F}_{x}(\q)$ and $\bm{F}_{R}(\q)$ are respectively the position and rotation matrix of the known end-effector, and $\bm{d}$ the displacement of the tool. The parameters $d$ can be optimized as well when training the PoE with maximum likelihood. The results is that a new end-effector is found with which the distribution of configurations $\q$ is best explained and compact.

	\paragraph{\textbf{Manipulability}}
	Measures of manipulability are other interesting transformations in robotics. The most simple form is a scalar defined as
	\begin{align}
	\trans{}{m}(\q) = \sqrt{\det(\bm{J}(\q)\bm{J}(\q)^\trsp)},
	\end{align}
	which can be used to explain the avoidance of singular configurations in the dataset, as in \cite{kalakrishnan2013learning}.
	
	For a more precise description of the configuration, velocity and force ellipsoids can be used \cite{yoshikawa1985manipulability}. The velocity ellipsoid is defined by the matrix
	\begin{align}
	\trans{}{m}(\q) = \big(\bm{J}(\q)\bm{J}(\q)^\trsp\big)^{-1},
	\end{align}
	and the force by its inverse. If we consider a zero-centered and unit-covariance Gaussian distribution of joint velocity, the velocity ellipsoid corresponds to the precision matrix (inverse of the covariance) of the task-space velocities. It thus relates to the feasible distribution of task-space velocities (respectively forces).
	For example, this full matrix can be used to define the distribution of configurations in which the transfer of velocities or forces along a direction should be maximized.
	
	Such matrix is positive semi-definite, which should be taken into account for the choice of the associated expert distribution. An option is to work with its Cholesky decomposition. A Wishart distribution is not expressive enough to define variances along different directions. In \cite{Jaquier20IJRR}, 
it is proposed to use Gaussian distributions in the tangent space of manifolds. This choice is motivated by the geometry of symmetric positive definite matrices (SPD). However, this approach ignores the geometry of the robot (the manipulability is a function of the joint angles and the kinematic structure). This comes with two important limitations. First, the proper support of the manipulability ellipsoid is ignored. By considering a distribution on the SPD manifold, it is assumed that this space can be covered by the robot, while the robot might only cover a subspace of it (which can also vary among robots). A second limitation is that manipulability is often employed as secondary objective. For example, when playing golf, hitting the ball at the right place is of higher importance than replicating a desired manipulability ellipsoid. The subspace of SPD matrices is further limited when it lies in the nullspace of more important tasks. A simple experiment with a manipulability measure will be presented in Sec.~\ref{sec:experiments:hierarchical_tasks} to show that this should be considered to learn the targeted manipulability. 

	Similarly, such consideration is important to transfer manipulability objectives between robots with different kinematic chains. In \cite{Jaquier20IJRR}, 
manipulability ellipsoid distributions are expressed on the set of symmetric positive definite matrices. We will show in the experiments of Sec.~\ref{sec:experiments:hierarchical_tasks} that manipulability-related tasks would be better transferred by considering that the distributions are expressed on subsets of these matrices. These experiments will motivate that a better support of the distributions can be considered by taking into account the differences in the kinematic chain configurations.
	
	\paragraph{\textbf{Relative distances}} A \textit{relative distance space} is proposed in \cite{yang2015real}. It computes the distances from multiple virtual points on the robot to other objects of interest (targets, obstacles). It can, for example, be used in environments with obstacles, providing an alternative or complementing standard forward kinematics.
	
	\paragraph{\textbf{Center of Mass (CoM)}}
	From the forward kinematics of the center of mass of each link and their mass, it is possible to compute the center of mass (CoM) of the robot. When considering mobile or legged robots, the CoM should typically be located on top of support polygons to satisfy static equilibrium on flat surfaces. 
	
	\paragraph{\textbf{Jacobian pseudoinverse iterations}}
	The following transformation is more interesting for the problem of sampling configurations from a given PoE than for the one of maximum likelihood from a dataset.
	Precise kinematics constraints imply a very correlated $\up(\q)$, as shown in Fig.~\ref{fig:poevsgmm}. 
	In the extreme case of hard kinematics constraints, the solutions are on a low dimensional manifold embedded in configuration space. 
	With most of the existing methods, it is very difficult to sample from this correlated PoE and to approximate it.
	Dedicated methods address the problem of representing a manifold as \cite{voss2017atlas+} or sampling from it, as \cite{zhang2013unbiased}.
	In \cite{berenson2011task}, a projection strategy is proposed. Configurations are sampled randomly and projected back using an iterative process. We propose a similar approach where the projection operator $\mathcal{P}_N$ would be used as transformation $\trans{m}{}$.
	Inverse kinematics problems are typically solved iteratively with
	\begin{align}
	\mathcal{P}(\q) &=\q + J(\q)^\dagger \Big(\xd - \bm{F}(\q)\Big),
	\label{equ:ik_iteration}
	\end{align} 
	where $\hat{\bm{p}}$ is the target and $J(\q)^\dagger$ is the Moore-Penrose pseudo-inverse of the Jacobian.
	This relation is derivable and can be applied recursively with
	\begin{align}
	\mathcal{P}_{0}(\q) &= \mathcal{P}(\q),\\
	\mathcal{P}_{n+1}(\q) &= \mathcal{P} \Big(\mathcal{P}_{n-1}(\q)\Big).
	\end{align} 
	
	
	Then, the distribution
	\begin{align}
	p_m(\q)\propto\mathcal{N}\Big(\mathcal{P}_{N}(\q) \Big| \, \xd, \sigma \bm{I}\Big)
	\end{align}
	is the distribution of configurations which converges in $N$ steps to $\mathcal{N}(\xd, \sigma \bm{I})$. 
	Thanks to the very good convergence of the iterative process \eqref{equ:ik_iteration}, $\sigma$ can be set very small.
	However, this approach has a similar (but less critical) problem as in \cite{berenson2011task}. The resulting distribution is slightly biased toward regions where the forward kinematics is close to linear (constant Jacobian), which are those where more mass converges to the manifold.
	
	With high DoFs robots, it might be computationally expensive to run iteration steps inside the stochastic gradient optimization and propagate the gradient. Another approach would be to define heuristically (or learn) $\bm{\Sigma}_h$ such that $\mathcal{N}(\q|\, \xd, \sigma \bm{I} + \bm{\Sigma}_h)$ is close to
	$\mathcal{N}(\mathcal{P}_{N}(\q)|\, \xd, \sigma \bm{I})$.

\begin{figure}
		\centering
		\includegraphics[width=1.\linewidth]{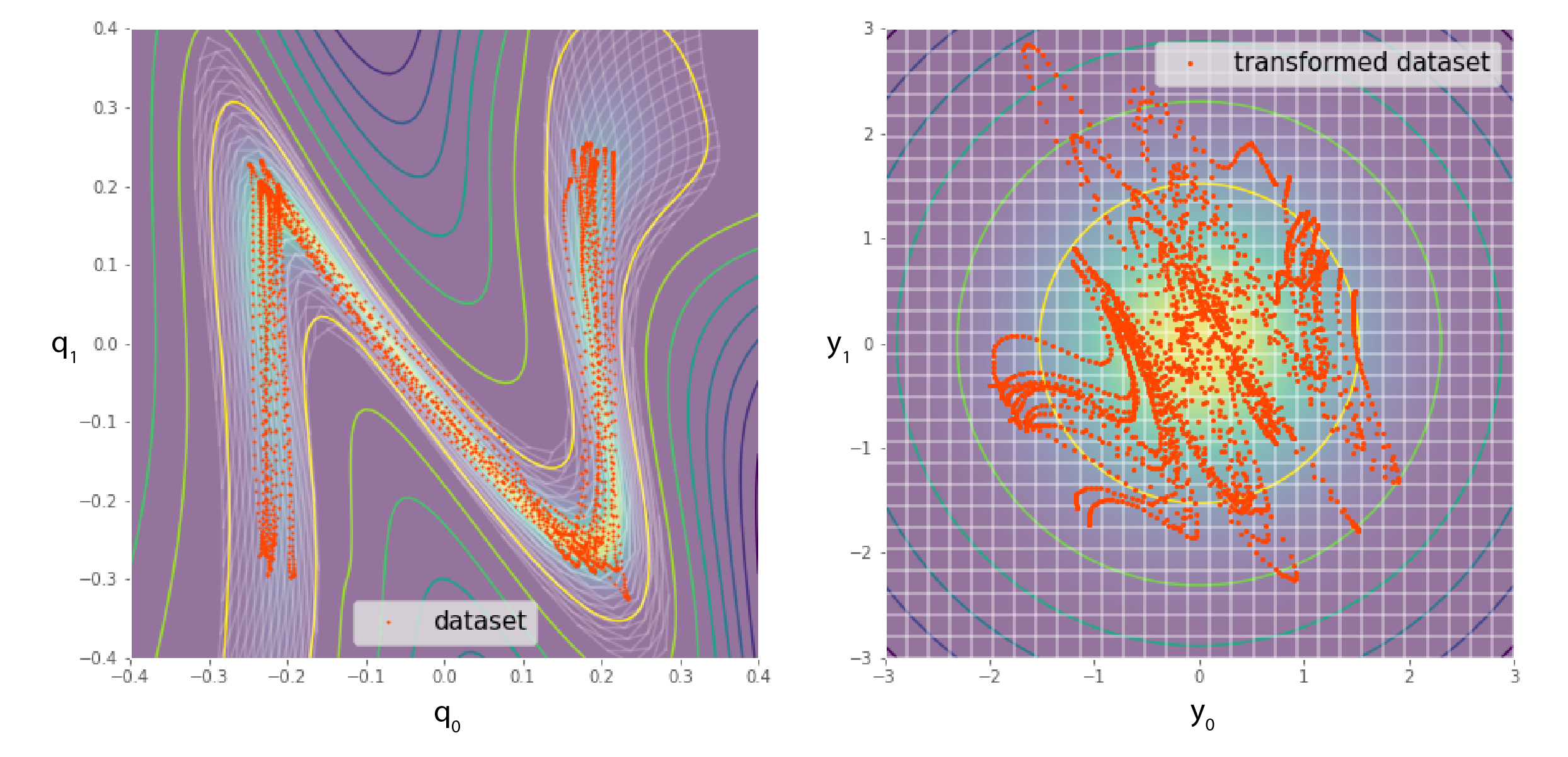}
		\caption{For greater expressiveness, an invertible neural network can be used as an expert transformation $\trans{}{m}$. Instead of optimizing the parameters of the expert density $p_m$, which can be set as a zero-centered and unit covariance Gaussian (right), the parameters of the network can be trained. The neural network becomes a transformation under which the dataset (N shape, \emph{left}) becomes a simple distribution (Gaussian, \emph{right}). The densities are shown as a colormap with isolines. A grid that undergoes the inverse transformation is also shown. This approach allows complex attractors or guiding distributions to be learned, where it is easy to control the robot in the transformed space.}
		\label{fig:transformed_density}
	\end{figure}

	\paragraph{\textbf{Neural network}}
	When more data are available, more general and complex transformations can be considered, such as neural networks. For example, it can be used to define a complex distribution in which the end-effector should be. Coupled with a control strategy (see Sec.\ref{sec:applications}), it can provide virtual guides as in \cite{raiola2015co} to constraint the robot on a trajectory or within a shape, as illustrated in Fig.~\ref{fig:transformed_density}.
	
	Particularly, we recommend using invertible neural networks for practical reasons of initializing the training procedure and for its interesting properties of global maximum when controlling the robot.
	Training a PoE with contrastive divergence is not as efficient as training models with tractable likelihood. Therefore, we propose to initialize the PoE by training all the experts that have tractable likelihood independently.
	Thus, we propose to treat an expert composed of a neural network transformation as a distribution with a change of variable. 
	This provides a tractable likelihood as in \cite{dinh2017density}, that can be used for initializing.
	Given that $\trans{}{m}$ is bijective, the tractable likelihood is given by the change of variable 
	\begin{align}
	p(\q) = p_m\big(\trans{}{m}(\q)\big)\, \Big| \det\Big( \frac{\partial\trans{}{m}(\q)}{\partial \q^\trsp}\Big)\Big|,
	\end{align}
	where $p_m$ is a simple distribution. $p_m$ can be chosen as a zero-centered and unit covariance Gaussian. Actually, a full covariance Gaussian can be described in this framework with $\trans{}{m}$ being a linear transformation.
	If we were to train the neural network without considering the renormalization, $\trans{}{m}$ will try to push all the configuration of the dataset to the mode of $p_m$. The term $\big| \det\big(\partial\trans{}{m}(\q)/\partial \q^\trsp\big)\big|$ can be seen as a cost preventing the contraction around the mode.
	
	To avoid overfitting, we propose a strategy to penalize abrupt change in the transformation $\trans{-1}{m}$. We propose to minimize the expectation of a measure of local change of the Jacobian of the inverse of the transformation $\partial\trans{-1}{m}$ under a distribution $p_r$
	\begin{align}
	\mathbb{E}_{p_r(\bm{y})}\Bigg[\int_{\bm{r}} \, \Big|
	\frac{\partial\trans{-1}{m}(\bm{y})}{\partial \bm{y}^\trsp} - 
	\frac{\partial\trans{-1}{m}(\bm{y}+\alpha \bm{w})}{\partial (\bm{y} + \alpha \bm{w})^\trsp} 
	\Big| \, d\bm{w} \Bigg],
	\end{align}
	where $\bm{w}$ is a unit vector on $\mathcal{S}^{n_m}$, $n_m$ being the dimensionality of $\bm{y}$, and $\alpha$ a small positive scalar. The distribution $p_r$ can be chosen as a $p_m$ but with up to 3 to 10 times bigger standard deviations, which would make sure that the transformation is smooth even further from the dataset. This cost can be optimized with stochastic gradient descent by evaluating the derivative on samples of $\bm{y}^{(n)} \sim p_r(\cdot)$ and of unit vectors $\bm{w}^{(n)}$.
	
	Another interesting property of using an invertible network is that the density under the transformation keeps a unique global maximum if the expert density $p_m$ has only one. It is justified because invex functions (functions that have only one global minimum and no local minimum) are still invex under a diffeomorphism \cite{pini1991invexity}. It is especially advantageous when we want to control the robot to track the density of the PoE, as explained in Sec.~\ref{sec:applications}.
	
	\subsection{Distributions}
	\label{subsec:distributions}
	Several common distributions are now presented to handle the transformations presented in the above. 

	\paragraph{\textbf{Multivariate normal distribution (MVN)}}
	An obvious choice for forward kinematics objectives is the Gaussian or multivariate normal distribution (MVN). Its log-likelihood is quadratic, making it compatible with standard inverse kinematics and optimal control techniques,
	\begin{equation}
	\pmvn(\q |\bm{\mu}, \bm{\Sigma})\propto 
	\exp\Big(-\frac{1}{2}(\q-\bm{\mu})^{\trsp}
	\bm{\Sigma}^{-1}(\q-\bm{\mu})\Big),
	\end{equation}
	where $\bm{\mu}$ is the location parameter and $\bm{\Sigma}$ is the covariance matrix. 
	This distribution is standard in robotics to represent full trajectories \cite{paraschos2013probabilistic}, subparts of trajectories with a hidden Markov model \cite{calinon2016tutorial} or a joint distribution of a phase and robot variables with a mixture \cite{calinon2009statistical}.
	
	\paragraph{\textbf{Matrix Bingham---von Mises---Fisher distributions (BMF)}}
	To handle orientations, for example, represented as a rotation matrix, Matrix Bingham---von Mises---Fisher distribution (BMF) \cite{khatri1977mises} can be used. Its normalizing constant is intractable and requires approximation \cite{kume2013saddlepoint}. In our case, this is not a problem since we integrate over robot configurations.
	Its density
	\begin{equation}
	\pbmf(\Q |\bm{A}, \bm{B}, \bm{C} )\propto 
	\exp\Big(\operatorname {tr} (\bm{C} ^{\trsp}\Q + 
	\bm{B}\,{\Q}^{\trsp} \bm{A}\, \Q)\Big),
	\end{equation}
	has a linear and a quadratic term as a Gaussian. $\bm{A}$ and $\bm{B}$ are often chosen as symmetric and diagonal matrices, respectively.
	By imposing additional constraints on $\bm{A}$ and $\bm{B}$, this density can be written as a Gaussian distribution on vectorized rotation matrices. By setting to zero the derivative of the log-likelihood
	\begin{align}
	\frac{\partial l(\Q)}{\partial \Q} = \bm{C} ^{T} + 2 \bm{A}\Q \bm{B} = 0,
	\end{align}
	and exploiting several properties of the trace and the Kronecker product we can rewrite the density as proportional to
	\begin{equation}
	\pmvn\Big(\mathrm{vec}(\Q) \Big|\, \mathrm{vec}(\bm{A}^{-1}\bm{C} ^{T}\bm{B}^{-1}), (\bm{A} \otimes \bm{B})^{-1}\Big).
	\end{equation}
	$\bm{A}$ and $\bm{B}$ should be both invertible and $(\bm{A} \otimes \bm{B})$ positive definite.
	
	In some tasks, it may be interesting to encode correlations between positions and orientations.
	Rewritten as a Gaussian, it is possible to create a joint distribution of position and rotation matrices
	\begin{equation}
	\footnotesize 
	\pmvn\Bigg(
	\begin{bmatrix}
	\q\\ \mathrm{vec}({\Q})
	\end{bmatrix}
	\Bigg|\\
	\begin{bmatrix}
	\bm{\mu}\\
	\mathrm{vec}(\bm{A}^{-1}\bm{C} ^{T}\bm{B}^{-1})
	\end{bmatrix}^{\trsp}\!,\! 	 \begin{bmatrix}
	\bm{\Sigma}_{\bm{xx}} & \bm{\Sigma}_{\bm{xX}}\\
	\bm{\Sigma}_{\bm{Xx}} & (\bm{A} \otimes \bm{B})^{-1}\\
	\end{bmatrix}\Bigg),
	\normalsize
	\label{equ:mvnmbmfjoint}
	\end{equation}
	where $\bm{\Sigma}_{\bm{xX}}\in \mathbb{R}^{3\times9}$ is the covariance between positions and orientations. This joint distribution can be ensured as valid by imposing a constraint on $\bm{\Sigma}_{\bm{xX}}$ with Schur complement condition for positive definiteness
	\begin{align}
	\bm{\Sigma} \succ 0 \iff \bm{\Sigma}_{\bm{xx}} \succ 0, 
	(\bm{A} \otimes \bm{B})^{-1}
	- \bm{\Sigma}_{\bm{Xx}}\bm{\Sigma}_{\bm{xx}}^{-1}\bm{\Sigma}_{\bm{xX}}
	\succ 0.
	\end{align}
	
	This complex parametrization and constraints are not mandatory in the PoE framework, as the experts do not need to be valid and properly normalized distributions themselves. However, it can help to reduce the number of parameters and to stabilize the learning procedure. As an alternative approach, we present an experiment in Sec.~\ref{sec:experiments:pos_rot_correlation} in which a joint distribution of positions and orientations is learned with a low-rank structure on the covariance. 
	
	\paragraph{\textbf{Matrix normal distribution}}
	Matrix valued transformations can be encoded with a matrix normal distribution
	\begin{multline}
	\pmn(\Q |\bm{M}, \bm{U}, \bm{V}){}\propto\\
	\exp\Big(-\frac{1}{2}\operatorname{tr}\big[\bm{V^{-1}}(\Q-\bm{M})^{\trsp}
	\bm{U}^{-1}(\Q-\bm{M})\big]\Big),
	\end{multline}
	where $\bm{M}\in\mathbb{R}^{n\times p}$ and $\bm{U}$ are $\bm{V}$ $n\times n$ and $p\times p$ positive definite matrices, respectively.
	It can also be written as a distribution of vectorized matrix as
	\begin{equation}
	\operatorname{vec}(\Q) \propto \pmvn\big(\!\operatorname{vec}(\bm{M}), (\bm{V} \otimes \bm{U})^{-1}\big),
	\end{equation}
	where the covariance matrix has fewer parameters than in a Gaussian.

	\paragraph{\textbf{Probabilistic movement primitives}}
	So far, the considered distributions targeted static configurations. Probabilistic movement primitives (ProMP) is a way to build Gaussian distributions of trajectories \cite{paraschos2013probabilistic}. 

	An observation $\q_t$ (position or joint angles) follows a Gaussian distribution
	\begin{align}
	p_{\textsc{ProMP}}(\q_t | \bm{\Phi}_t, \bm{w}, \bm{\Sigma_x})&= \pmvn(\q_t | \bm{\Psi}_t^\trsp \bm{w}, \bm{\Sigma_x}),
	\end{align}
	where $\bm{w}$ is a weight vector and
	$\bm{\Phi}_t$ a time-dependent basis matrix. The weight vector also follows a Gaussian distribution. The marginal distribution of the observation given the parameters of the model becomes
	\begin{multline}
	\small
	p_{\textsc{ProMP}}(\q_t | \bm{\Phi}_t, \bm{\mu}_{\bm{w}}, \bm{\Sigma}_{\bm{w}}, \bm{\Sigma_x}) = \\ 
	\int \pmvn(\q_t | \bm{\Psi}_t^\trsp \bm{w}, \bm{\Sigma_x}) \,
	\pmvn(\bm{w} | \bm{\mu}_{\bm{w}}, \bm{\Sigma}_{\bm{w}})d\bm{w} = \\
	\pmvn(\q_t | \bm{\Psi}_t^\trsp \bm{\mu}_{\bm{w}}, \bm{\Psi}_t^\trsp \bm{\Sigma}_{\bm{w}}\bm{\Psi}_t +\bm{\Sigma_x}).
	\end{multline}
	
\begin{figure}
		\centering
		\includegraphics[width=0.7\linewidth]{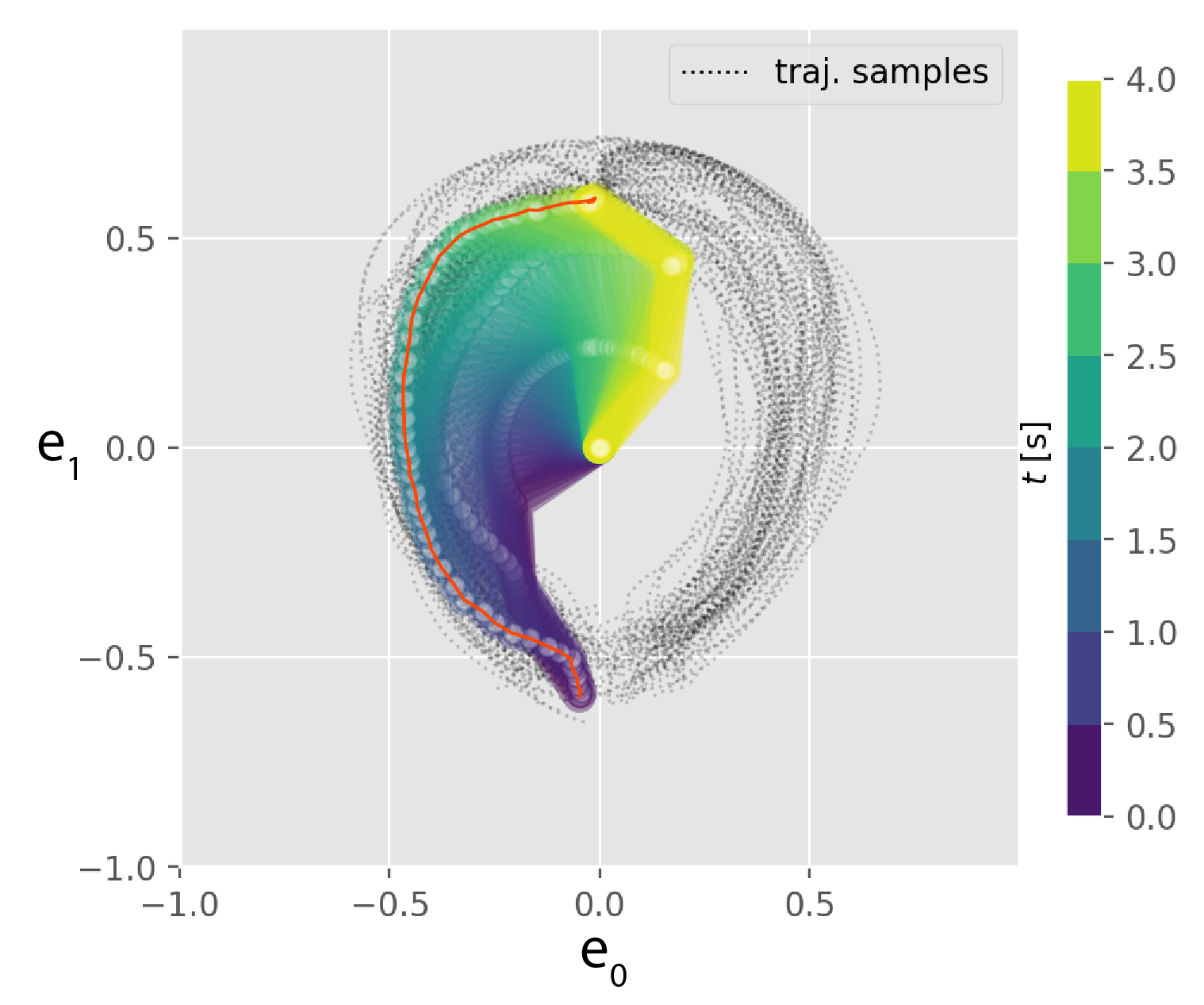}
		\caption{ To encode distributions of trajectories, probabilistic movement primitives (ProMP) can be used as experts in a PoE. In this example, a product of two ProMP experts is considered: one defines a joint angle trajectory, while the other defines the trajectory of the end-effector. A 3-DoF planar robot is used in this example. The product of ProMPs is approximated using variational inference as a mixture of ProMPs in configuration space. Samples of trajectories of the end-effector are displayed, as well as a full sequence of configurations. Even if each expert is Gaussian, the product of ProMPs is multimodal because of the planar robot kinematics.}
		\label{fig:popromp001}
	\end{figure}

	ProMPs are fully compatible with our framework. We can consider, for example, a product of ProMPs in multiple task spaces and configuration space. For approximating the product of ProMPs, it is also possible to use variational inference with a mixture of Gaussians, as proposed in Sec.~\ref{sec:variational_inference}. Instead of using full location and full covariance, the mixture components can be a ProMP as well. In Fig.~\ref{fig:popromp001}, a mixture of ProMPs in configuration space is used to approximate a product of a ProMPs in task space and in configuration space. A 3-DoF planar robot is used, which induced the multimodality of the product.
	
	\paragraph{\textbf{Approximate distributions in the tangent space of manifolds}}
	Another approach compatible with the PoE framework is to consider Gaussian distributions in the tangent space of manifolds, as a way to encode orientations \cite{zeestraten2017approach} or manipulability ellipsoids \cite{Jaquier20IJRR}. 
It has the advantages of being generic to many different types of data. Also, the distribution is approximately normalized in the tangent space. It provides an easier way to initialize this expert individually (using MLE) than if it was unnormalized. The only restriction is that the logarithmic map $\logm{\q}{\bm{\mu}}$, mapping elements from the manifold to the tangent space, should be differentiable. The density is given by
	\begin{equation}
	\pmvn(\q |\bm{\mu}, \bm{\Sigma})\propto
	\exp\Big(-\frac{1}{2}\logm{\q}{\bm{\mu}}^{\trsp}
	\,\bm{\Sigma}^{-1}\,\logm{\q}{\bm{\mu}}\Big).
	\end{equation}
	
	\paragraph{\textbf{Cumulative distribution functions (CDF)}}
	Inequality constraints, such as static equilibrium, obstacles or joint limits can be learned using cumulative distribution function
	\begin{align}
	p(x \leq b),\quad \text{with}\quad x \sim \mathcal{N}\Big(\trans{}{}(\q), \sigma^2\,\Big),
	\label{equ:cdf}
	\end{align}
	where $\trans{}{}(\q)$ is a scalar. For example, for half-plane constraints, $\trans{}{}(\q)$ could be $\bm{w}^\trsp\q$, or for joint limits on first joint $\trans{}{}(\q)=\q_0$.
	
	The use of the CDF makes the objectives continuous and allows safety margin determined by $\sigma$ to be considered.
	
	Obstacles constraints might be impossible to compute exactly and require collision checking techniques (\cite{elbanhawi2014sampling}). With stochastic optimization, our approach is compatible with a stochastic approximation of the collision related cost, which might speed up computation substantially.

	\paragraph{\textbf{Uni-Gauss distributions}}
	In Sec.~\ref{sec:poe_nullspace}, we showed how a PoE is compatible with standard nullspace approaches to represent hierarchy between multiple tasks. Another way to address this problem is to use uni-Gauss experts, as proposed in \cite{hinton1999products}. These experts combine the distribution defining a non-primary objective $p_m$ with a uniform distribution
	\begin{align}
	p_{\mathrm{UG},m}(\q) = \pi_m p_m(\q) + \frac{1-\pi_m}{c},
	\label{equ:unigauss}
	\end{align}
	which means that each objective has a probability $p_m$ to be fulfilled and $1-\pi_m$ not to be fulfilled.
	
	Classical prioritized approaches exploit redundancies of the robot to achieve multiple tasks simultaneously \cite{nakamura1987task}. A nullspace projection matrix is used such that commands required to solve a secondary task do not influence the primary task.
	Using uni-Gauss experts is a less strict approach that does not necessarily require redundancies. Even if there is no redundancy to solve a secondary task without influencing the first one, there may be some mass at the intersection between the different objectives.
	
	There are two possible ways of estimating $\up(\q)$ in the case of Uni-Gauss experts. If the number of tasks is small, we can introduce, for each task $m$, a binary random variable indicating if the task is fulfilled or not. For each combination of these variables, we can then compute $\up(\q)$. The ELBO can be used to estimate the relative mass of each of these combinations, as done in model selection. For example, if the tasks are not compatible, their product would have a very small mass, as compared to the primary task. In the case of numerous objectives, this approach becomes computationally expensive because of the growing number of combinations. We can instead marginalize these variables and we fall back on \eqref{equ:unigauss}. For practical reasons of flat gradients, the uniform distribution can be implemented as a distribution of the same family as $p_m$, but with a higher variance.
	

	\section{Control}
	\label{sec:applications}
	In this section, we present two control strategies that can be used together with PoEs. In the first, the PoE defines the preferred configurations of the robot. The robot should go to these configurations and stay there, facing perturbations.
	The negative log density $-\log p(\q | \bm{\theta}_1, \dots, \bm{\theta}_M)$ of the PoE is used as a cost function in an optimal control problem. This approach is illustrated in Fig.~\ref{fig:poe_manipulator}. In the second scenario, the PoE defines a distribution of configurations that the robot should actively visit.
	%
	
	\subsection{Optimal control}
	In optimal control, control commands $\bm{u}$ are computed with the aim of minimizing a cost based on the control commands and on the states of the system $\bm{\xi}$. The discrete time linear quadratic tracker (LQT) is a popular tractable subproblem of optimal control, where the dynamics of the system are linear
	\begin{align}
	\bm{\xi}_{t+1} = \bm{A}_t \bm{\xi}_t + \bm{B}_t \bm{u}_t,
	\end{align}
	with $\bm{A}_t$ and $\bm{B}_t$ being the time-dependent parameters of the system. In LQT, the cost is quadratic, given as
	\begin{align}
	J = 
	\frac{1}{2} \sum_{t=1}^{N-1} \Big((\bm{\xi}_t-\bm{z}_t)^\trsp \bm{Q}_t(\bm{\xi}_k-\bm{z}_t) + \bm{u}_t^\trsp\bm{R}\bm{u}_t \Big),
	\label{equ:lqr_cost}
	\end{align}
	where $\bm{z}_t$ is a desired state, $\bm{Q}_t$ is a matrix weighting the deviation from the desired state and $\bm{R}$ is a matrix weighting the penalization of the control commands. The minimization of the control commands have multiple underlying goals, such as reducing energy consumption, producing smooth movements through small accelerations, or ensuring safety through the use of small forces. The resulting optimal controller is linear, taking the form
	\begin{align}
	\bm{u}_t = -\bm{K}_t\bm{\xi}_t + \bm{K}^v_t \bm{v}_{t+1},
	\label{equ:lqr_controller}
	\end{align}
	where $\bm{K}_t\bm{\xi}_t$ is a feedback term and $\bm{K}^v_t \bm{v}_{t+1}$ is a feedforward term. More details about the derivations can be found in \cite{bohner2011linear}.
	
	We can use this formulation to compute a controller that would stay in regions of high density of the PoE. Let us consider that we have a linearized model of the manipulator
	\begin{align}	
	\begin{bmatrix}
	\q_{t+1}\\
	\dot{\q}_{t+1}\\
	\end{bmatrix} = \bm{A}_t \begin{bmatrix}
	\q_{t}\\
	\dot{\q}_{t}\\
	\end{bmatrix} + \bm{B}_t \bm{u}_t,
	\end{align}	
	where $\q$ is the configuration of the robot and $\bm{u}_t$ can be joint accelerations or torques, depending on the controller used on the robot. The state can be augmented with the different experts transformations
	\begin{align}
	\setlength\arraycolsep{1pt}
	\bm{\xi}_t =
	\begin{bmatrix}
	\trans{}{1}(\q_t)\\
	\vdots\\
	\trans{}{M}(\q_t)\\\hline
	\q_{t}\\
	\dot{\q}_{t}
	\end{bmatrix} 
	=\begin{bmatrix}
	\bm{y}_{1, t}\\
	\vdots\\
	\bm{y}_{M, t}\\ \hline
	\q_{t}\\
	\dot{\q}_{t}\\
	\end{bmatrix}.
	\label{equ:augmented_state_lqr}
	\end{align}
	The dynamics of the augmented system are linearized using the Jacobian
	$\bm{J}_m(\q) = \frac{\partial \trans{}{m}(\q)}{\partial \q}$ of each transformation $m$ 
	\begin{align}
	\bm{y}_{m, t+1} &\approx \bm{y}_{m, t} + \bm{J}_m(\q_t)\,(\q_{t+1} - \q_{t})\notag\\
	&\approx \bm{y}_{m, t} + \bm{J}_m(\q_t)\,\dot{\q}_t\, \Delta t,
	\end{align}	
	where $\Delta t$ is the discretization of time. The complete system can then be written in matrix form as 
	\begin{align}	
	\setlength\arraycolsep{1pt}
	\overbrace{\begin{bmatrix}
		\bm{y}_{1, t+1}\\
		\vdots\\
		\bm{y}_{M, t+1}\\
		\hline
		\q_{t+1}\\
		\dot{\q}_{t+1}\\
		\end{bmatrix}}^{\bm{\xi}_{t+1}} = 
	\overbrace{\left[\begin{array}{ccc|cc}
		\bm{I}_{d_1}& \cdots &\bm{0} & \,\bm{0} & \,\Delta t\hat{\bm{J}}_1 \\
		\vdots & \ddots &\vdots  & \vdots & \vdots\\
		\bm{0} &  \cdots &  \bm{I}_{d_M} & \,\bm{0} & \,\Delta t\hat{\bm{J}}_M\\
		\hline
		\bm{0} & \cdots& \bm{0} &\multicolumn{2}{c}{\bm{A}_t}
		\end{array}\right]}^{\tilde{\bm{A}_t}}
	\overbrace{\begin{bmatrix}
		\bm{y}_{1, t}\\
		\vdots\\
		\bm{y}_{M, t}\\ \hline
		\q_{t}\\
		\dot{\q}_{t}\\
		\end{bmatrix}}^{\bm{\xi}_t} +\notag \\ 
		\setlength\arraycolsep{1pt}
		\overbrace{\begin{bmatrix}
		\begin{bmatrix}
		\bm{0} & \hat{\bm{J}}_1 
		\end{bmatrix}\bm{B}_t \\ \vdots \\
		\begin{bmatrix}	
		\bm{0} & \hat{\bm{J}}_M 
		\end{bmatrix}\bm{B}_t\\ \hline
		\bm{B}_t\end{bmatrix}}^{\tilde{\bm{B}}_t} \bm{u}_t.
	\label{equ:augmented_system_lqr}
	\end{align}	
	The system $\tilde{\bm{A}_t}$ and $\tilde{\bm{B}_t}$ should not depend on the state; for this reason, the Jacobians should be evaluated as $\hat{\bm{J}}_m$. There could be evaluated either at the local maximum of the PoE density where the robot would converge or at the current state of the robot. The cost is defined as
	\begin{align}
	J = \sum_{t=1}^{N-1} \Big(-\sum_{m=1}^M \log p_m(\bm{y}_{m, t}) + \frac{1}{2} \dot{\q}_t^\trsp\tilde{\bm{Q}}\dot{\q}_t + \frac{1}{2}\,  \bm{u}_t^\trsp\bm{R}\bm{u}_t \Big),
	\label{lqr_poe_cost}
	\end{align}
	where $\tilde{\bm{Q}}$ penalizes the velocities. With the weight control matrix $\bm{R}$, they can be chosen as a diagonal matrix
	\begin{align}
	\tilde{\bm{Q}}=\begin{bmatrix}
	\varsigma_1^{-2} & \cdots & 0\\
	\vdots & \ddots & \vdots \\
	0 & \cdots & \varsigma_P^{-2} \\
	\end{bmatrix},\;
	\bm{R}=\begin{bmatrix}
	\tau_1^{-2} & \cdots & 0 \\
	\vdots & \ddots & \vdots\\
	0 & \cdots & \tau_P^{-2} \\
	\end{bmatrix},
	\label{equ:weight_control_lqr}
	\end{align}
	where $\varsigma_p$ and $\tau_p$ are respectively the range of velocities and the forces (or accelerations) allowed on joint $p$. They both correspond to the definition a desired Gaussian distribution of velocities and forces of standard deviation $\varsigma_p$ and $\tau_p$ on joint $p$.
	If all the experts $p_m$ are Gaussian, this cost can be rewritten exactly as \eqref{equ:lqr_cost} with
	\begin{align}
	\bm{Q} = \begin{bmatrix}
	\bm{\Sigma}_1^{-1} & \cdots & \bm{0} & \bm{0} \\
	\vdots & \ddots &\vdots  & \vdots\\
	\bm{0} & \cdots & \bm{\Sigma}_M^{-1} & \bm{0} \\
	\bm{0} & \cdots &\bm{0}& \tilde{\bm{Q}}\\
	\end{bmatrix},\; \bm{z_t} = \begin{bmatrix}
	\bm{\mu}_1\\
	\vdots\\
	\bm{\mu}_M\\
	\bm{0}
	\end{bmatrix},
	\label{equ:exp_cost_lqr}
	\end{align}
	since the log-likelihood is a quadratic function. In the other case, a quadratic approximation can be used as in iterative LQR, see \cite{li2004iterative}.
	The resulting controller is a feedback controller as in \eqref{equ:lqr_controller}, where the feedback is executed on the different expert transformations
	\begin{align}
	\bm{u}_t = \sum_{m=1}^{M} -\bm{K}_{m, t} \trans{}{m}(\bm{\xi}_t) + \bm{K}^v_{m, t} \bm{v}_{m, t+1}.
	\label{equ:lqr_controller_poe}
	\end{align}
	The gains $\bm{K}_{m,t}$ obtained for each expert $m$ depend on the required precision (the higher the precision, the higher the gain).
	
	If the robot is controlled by torque commands, the whole procedure of training a PoE and controlling the robot can be used to set up automatic virtual guides, as shown in Fig.~\ref{fig:poe_manipulator}. Multiple constraints, such as keeping an orientation, staying on a plane, or pointing towards an object can be learned from demonstration, and reproduced by the robot using the proposed feedback controller. The robot will then be free to be manipulated along directions that are not encoded by the experts (or that have a high variance).
	
\begin{figure}
		\centering
		\includegraphics[width=1.\linewidth]{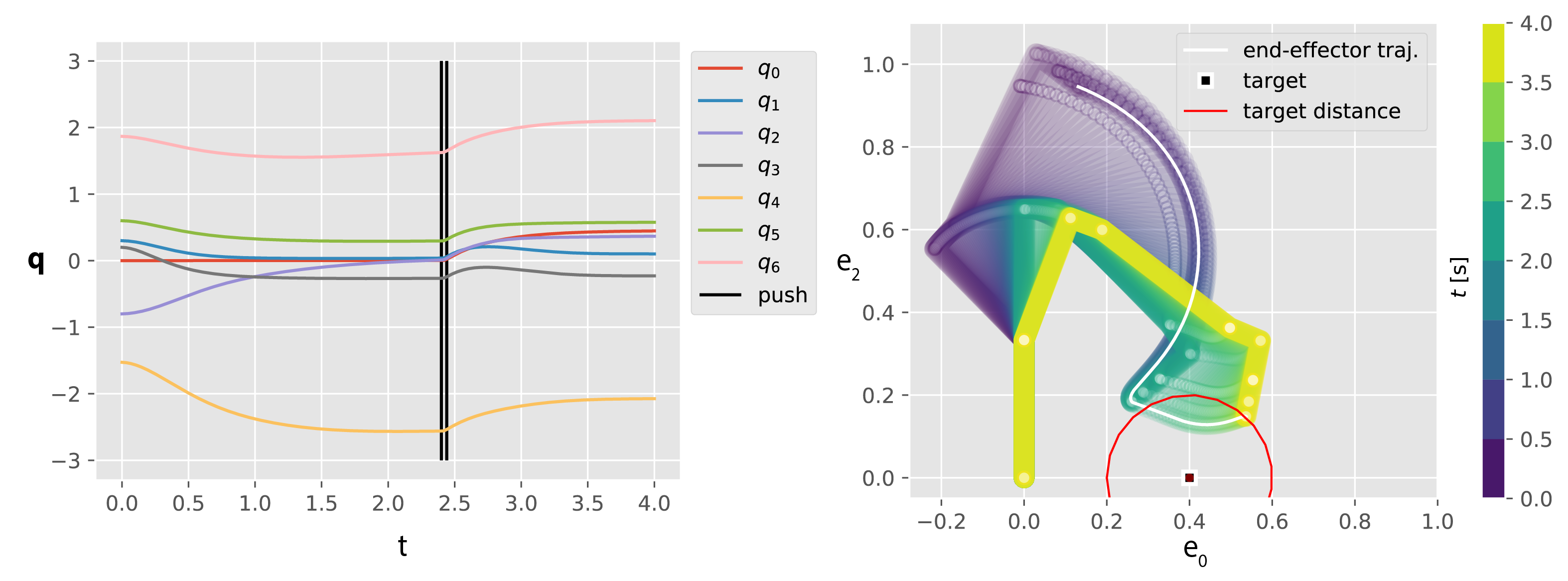}
		\caption{Example of 7-DoF manipulator controlled with an LQT. To show the time evolution, only the kinematic chain is displayed. The PoE is the intersection between a vertical plane ($e_1=0$) and a sphere. The two experts act on the end-effector. They are defined by a Gaussian distribution on $e_1$ around $0$ and by a Gaussian distribution on the log-distance to a target (black square). On the left, the evolution of the values of the joint angles is displayed over time. At $t=2.35[s]$, a perturbation in configuration space occurs. The robot converges to a new local mode of the PoE.}
		\label{fig:lqr_poe}
	\end{figure}

	Fig.~\ref{fig:lqr_poe} shows a simulation of 7-DoF manipulator controlled with this strategy, with a perturbation occurring at $t=2.35[s]$. 
	
	This controller can easily be implemented in a robotic manipulator with two separate loops. A high-level loop solves the LQT problem while a low-level control loop executes the last computed controller until a new one is available. These two loops are described in Alg.~\ref{code:python_loop} and \ref{code:cpp_loop}.
	\begin{algorithm}[h]
		set $\bm{R}$ and $\tilde{\bm{Q}}$ as in \eqref{equ:weight_control_lqr}\\
		compute $\bm{Q}$ and $\bm{z}$, LQT cost from the expert distributions as in \eqref{equ:exp_cost_lqr}\\
		\While{controlling}{
			(\textit{loop duration : $2 - 10[\mathrm{ms}]$})\\
			compute $\bm{A}$ and $\bm{B}$, linearized model of the manipulator around current state\\
			get $\q$ current configuration of the robot\\
			\ForEach{expert $m$}{compute $\hat{\bm{J}}_m(\q)$}
			compute $\hat{\bm{A}}$ and $\hat{\bm{B}}$, augmented linear model as in \eqref{equ:augmented_system_lqr}\\
			solve finite or infinite horizon LQT with constant parameters over the horizon\\
			publish $\bm{K}_1$, $\bm{K}^v_1$, $\bm{v}_{2}$, LQT controller parameters\\
		}
		\caption{High-level loop (coded in Python)}
		\label{code:python_loop}
	\end{algorithm}
	\begin{algorithm}[h]
		\While{controlling}{
			(\textit{loop duration : $<1[\mathrm{ms}]$})\\
			read $\bm{K}_1$, $\bm{K}^v_1$, $\bm{v}_{2}$, current  LQT controller parameters\\
			get $\q$, $\dot{\q}$ current configuration and velocity of the robot\\
			\ForEach{expert $m$}{compute $\trans{}{m}(\q)$
				(call Python functions)} 
			compute $\bm{\xi}$ as in \eqref{equ:augmented_state_lqr} \\
			compute and apply torques or accelerations as $\bm{u} = -\bm{K}_t\bm{\xi}_1 + \bm{K}^v_1 \bm{v}_{2}$\\
		}
		\caption{Low-level control loop (coded in C++)}
		\label{code:cpp_loop}
	\end{algorithm}
	
	\subsection{Ergodic control}
	In the second control strategy, a trajectory should be planned to visit the PoE density. This problem is referred to as ergodic coverage in \cite{mathew2011metrics} and has multiple applications, such as surveillance or target localization. Closer to the considered manipulation tasks, it can be used to discover objects, learn dynamics or polish/paint surfaces.
	%
	In \cite{ayvali2017ergodic}, the ergodic coverage objective is formulated as a KL divergence between the density to visit $p(\q)$ and the time-average statistics of the trajectory $\Gamma(\q)$
	\begin{align}
	\Dkl{\Gamma}{p} &= \int_{\q} \Gamma(\q) \log \frac{\Gamma(\q)}{p(\q)} d\q.
	\label{equ:ergodic_objective}
	\end{align}
	The time-average statistics defines the density covered by the trajectory of the robot. Using the formulation with a KL divergence, particular sensors or areas of influence can be taken into account. For example, a Gaussian can be used to model the area around the robot perceived by its sensors. The time-average statistics become a mixture of Gaussians
	\begin{align}
	\Gamma(\q) \approx \frac{1}{N}\sum_{t=0}^{N-1}\mathcal{N}(\q| \q_t, \bm{\Sigma}),
	\end{align}
	where $\q_t$ is the configuration of the robot at time $t$ (for a trajectory of $N-1$ timesteps). The covariance $\bm{\Sigma}$ can model the coverage of the sensor. A small $\bm{\Sigma}$ means a short-distance coverage and implies finer trajectories. The objective \eqref{equ:ergodic_objective} is the same as in variational inference and can be optimized with the ELBO objective in \eqref{equ:elbo_1}. Thus, it is not required to have a properly normalized density to visit. Given that the covered zone is modeled by a tractable density (from which we can sample, and properly normalized), the ergodic objective can be optimized easily with stochastic gradient descent. Compared to the ELBO objective in \eqref{equ:elbo_1}, an additional constraint has to be added for limiting velocities and accelerations so that $\{\q_t\}_{t=1}^{N-1}$ is a consistent trajectory.
	
\begin{figure}
		\centering
		\includegraphics[width=0.9\linewidth]{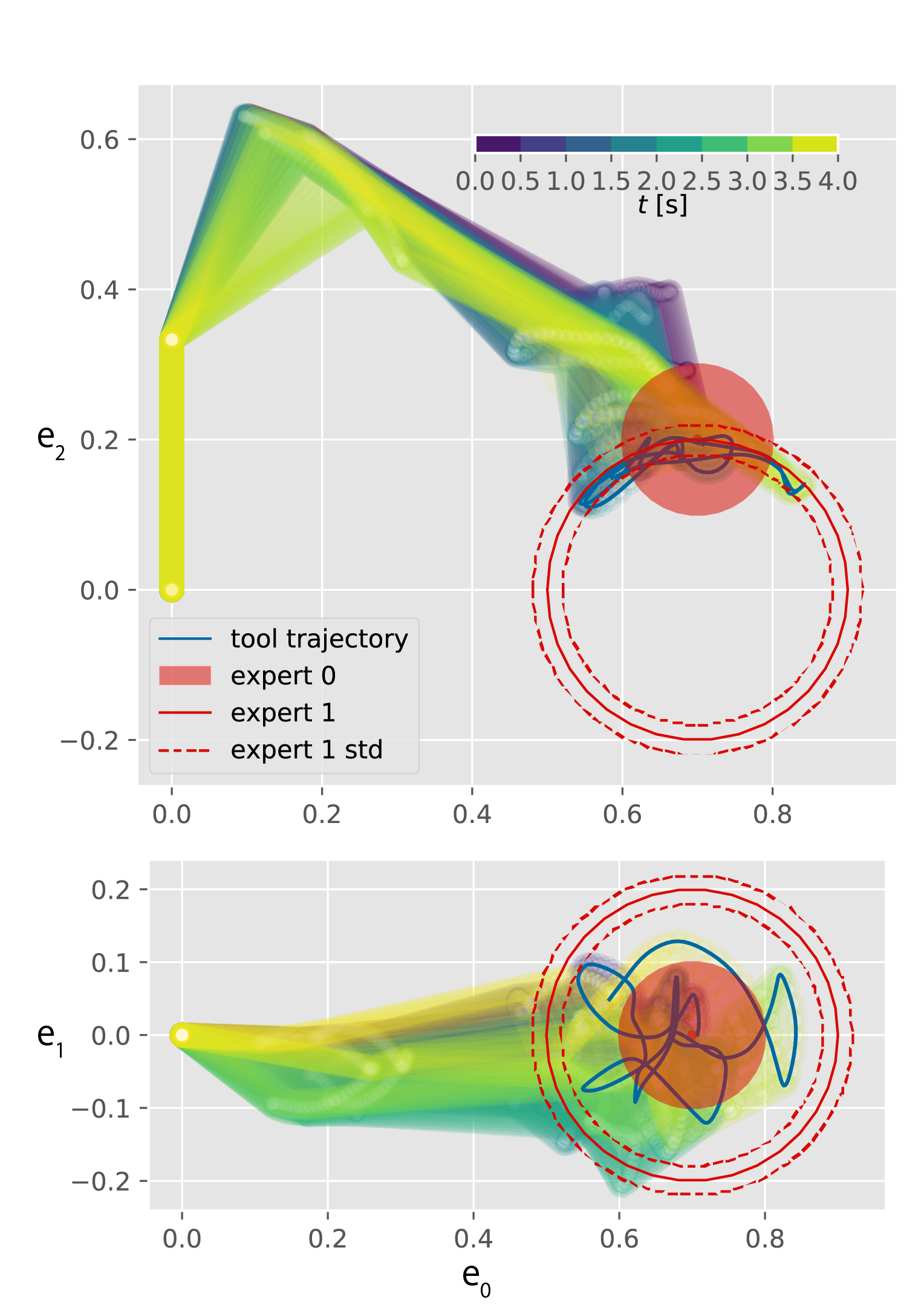}
		\caption{The trajectory of the robot is optimized to cover the PoE. The density is defined by a position constraint of the end-effector (expert 0) and a distance constraint (expert 1).}
		\label{fig:ergodic_poe}
	\end{figure}

	Fig.~\ref{fig:ergodic_poe} shows a trajectory optimized to cover the PoE. The robot is assumed to be controlled by velocity commands. Velocities and accelerations are minimized, and the time-averaged statistics is a mixture of Gaussians in task space.
	
	\section{Experiments}
	\newcommand{\expstate}{\mathrm{state:}\quad}
	\newcommand{\exptrans}{\mathrm{transformations:}\quad}
	\newcommand{\expexp}{\mathrm{experts:}\quad}
	\newcommand{\expparam}{\mathrm{parameters\, to\, learn:}\quad}
	\label{sec:experiments}
	\subsection{Multimodal distributions}
	\label{sec:experiments:multimodal_distribution}
	\begin{figure}
		\centering
		\includegraphics[width=1.\linewidth]{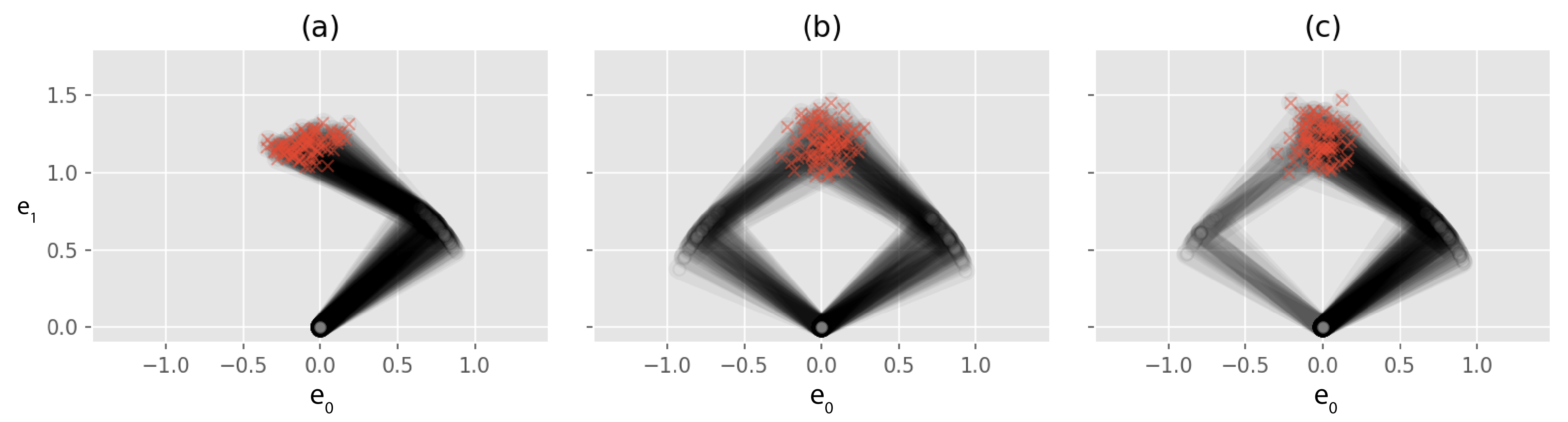}
		\caption{Samples of configurations for three tasks that illustrate the advantages of using variational inference with a mixture model for training a PoE. A 2-DoF planar robot is considered. \textit{(a)} Joint angle target. \textit{(b)} Operational space target \textit{(c)} Operational space target and joint angle target. }
		\label{fig:poetasks001}
	\end{figure}
	
	In a first experiment, we show the advantages of using variational inference to approximate the derivative of the normalizing constant.
	We consider a 2-DoF planar robot that should learn to distinguish between an operational-space or a configuration-space objective. Three datasets are used and shown in Fig.~\ref{fig:poetasks001}. They correspond to the following subtasks: (a) only configuration space target (b) only task space target, involving a multimodal PoE (c) a task-space target with a joint angle preference, involving two modes of unequal weights.
	The PoE model is defined by the following transformations and distributions:
	\begin{align*}
	\expstate
	&\q \in \mathbb{R}^2,\\
	\exptrans
	&\bm{y}_1 = \trans{}{1}(\q)= \bm{F}_{\bm{x}}(\q) \in \mathbb{R}^2,\\
	&\bm{y}_2 = \trans{}{2}(\q)= \q \in \mathbb{R}^2,\\
	\expexp
	&\bm{y}_1 \sim \mathcal{N}(\bm{\mu}_1, \sigma_1 \bm{I}),\\
	&\bm{y}_2 \sim \mathcal{N}(\bm{\mu}_2, \sigma_2 \bm{I}),\\
	\expparam
	&\bm{\mu}_1, \sigma_1, \bm{\mu}_2, \sigma_2,
	\end{align*}
	where $\bm{F}_{\bm{x}}(\q)$ is the position of the end-effector of the planar robot. The first and second experts analyzes task-space objectives and configuration-space objectives, respectively.
	Given three datasets of $N=30$ configurations, the corresponding models are learned either with contrastive divergence (as proposed in \cite{hinton2002training}) or using variational inference (as proposed in Sec.~\ref{sec:training_poe}). The process is repeated multiple times with random initializations of the parameters.
	
	Quantitative evaluations are performed by computing the alpha-divergence with $\alpha=1/2$ between the ground-truth density from which the dataset was sampled and the density learned with the different techniques. This divergence is also related to the Bhattacharyya coefficient as
	\begin{align}
	D_{\alpha=1/2}(p||\approxq) = -2 \log \int \sqrt{p(\q)\approxq(\q)} d\q.
	\end{align}
	This integral was evaluated by discretizing the space. Results are reported in Table \ref{table:cd_vs_variational}. In all cases, the variational approximation performs better. The performance of contrastive divergence is especially poor in case (a), showing high variations. The explanation comes from the incapability of standard sampling methods to jump across distant modes. Only the existence of a second mode, as in case (b) and (c) can help to distinguish between a configuration space and a task space target. A Markov chain initialized on the data never discovers the second mode that would be implied by a task space target. This potential waste of probability mass goes unnoticed by contrastive divergence. In contrast, variational inference with a mixture of Gaussians can localize this region of probability. 
The high variance of the contrastive divergence is due to the nonexistent gradient between the relative standard deviations $\sigma_1$ and $\sigma_2$ of the two experts. The final results are thus determined mostly by the random initialization.
	
	\begin{table}
		\begin{center}
			\caption{Quantitative evaluations of the learned distribution using contrastive divergence (CD) and our approach based on variational inference with a Gaussian mixture model (VI). The table shows alpha-divergence $D_{\alpha=1/2}$ measures between the different approximations and the ground-truth distribution, computed for the three tasks shown in Fig.~\ref{fig:poetasks001}.}
			\begin{tabular}{llll} 
				\toprule Task 
				& (a) & (b) & (c)\\ 
				\midrule 
				CD
				& 0.837 $\pm$ 0.597& 0.039 $\pm$ 0.030& 0.063 $\pm$ 0.026\\ 
				\midrule 
				\textbf{VI}
				& \textbf{0.067 $\pm$ 0.060} & \textbf{0.016 $\pm$ 0.022} & \textbf{0.014 $\pm$ 0.004}\\ 
				\bottomrule 
				\label{table:cd_vs_variational}
			\end{tabular}
		\end{center}
	\end{table}
	
	\subsection{Hierarchical tasks}
	\label{sec:experiments:hierarchical_tasks}
	In this set of experiments, the robot has to learn two competitive objectives, where one has a higher priority than the other.
	The secondary objective is masked by the resolution of the first objective and has to be uncovered.

\begin{figure}
		\centering
		\includegraphics[width=1.\linewidth]{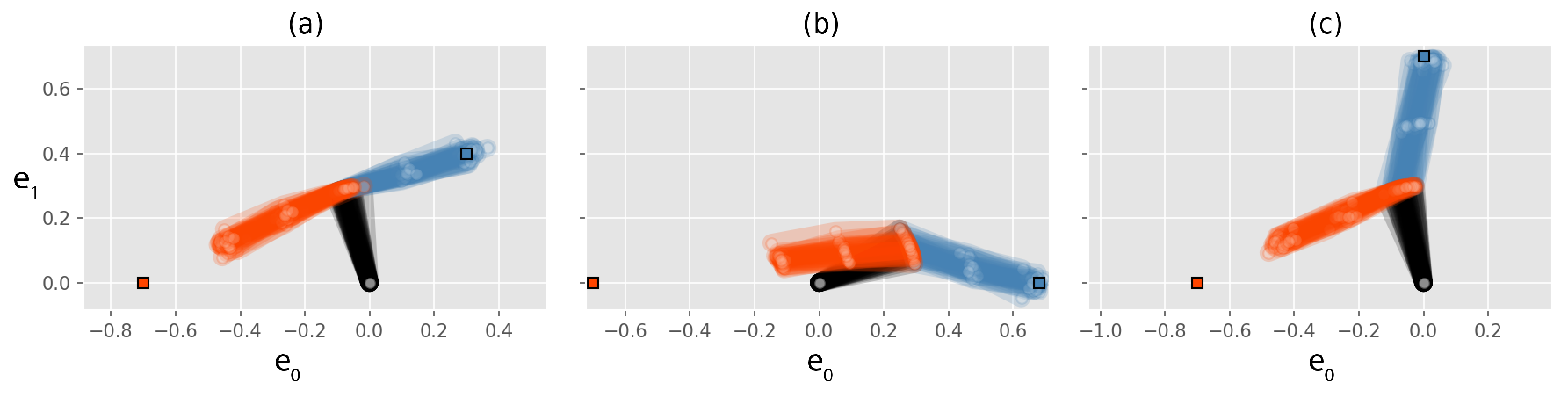}
		\caption{Dataset for the bimanual task with a hierarchy. Three situations are given, where each end-effector should track the target of the same color (displayed as a square).}
		\label{fig:bimanualnullspacedata}
	\end{figure}

	\paragraph{\textbf{Planar robot}}
	We first evaluate our approach in simpler cases, with planar robots. In the first task, we consider a 5-DoF planar robot with two dependent arms, as illustrated in Fig.~\ref{fig:bimanualnullspacedata}. Each end-effector should track its own target, which should be recovered from demonstrations. The end-effector with the highest priority task has three different targets $\{\bm{\mu}_1^{(i)}\}_{i = 0,\dots, 2}$ and the other end-effector has a fixed target $\bm{\mu}_2$. For each of the three cases $i=0,\dots, 2$, $N=30$ independent samples $\samples{\q}_n$ are given. The samples are generated by approximating a ground-truth product of experts with a mixture of $K=50$ Gaussian components. The end-effectors try to follow a normal distribution with a standard deviation $\sigma_1=\sigma_2=0.02$ around their respective targets. In each situation, the secondary target is not reachable.
	\begin{figure}
		\centering
		\includegraphics[width=1.\linewidth]{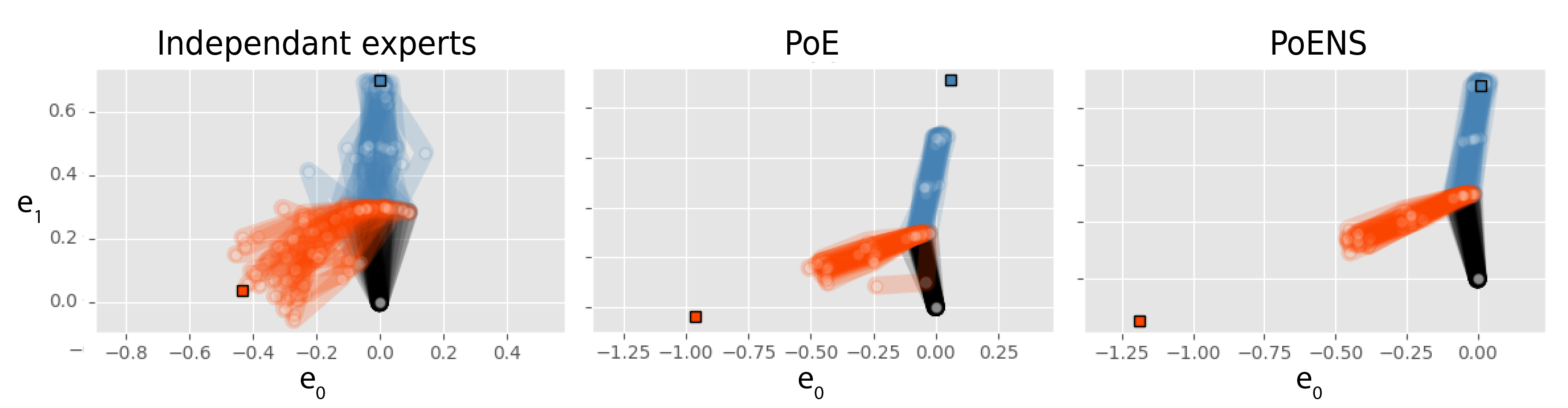}
		\caption{Samples from the different models learned for the bimanual task with a hierarchy. \emph{Left:} Using independent experts, the secondary task is not well understood and its variance is over-estimated. \emph{Center:} Using a PoE, the target of the primary task is inferred to be further away. \emph{Right:} Using a PoE with a nullspace structure, the targets are well recovered.}
		\label{fig:bimanualnullspacecomp}
	\end{figure}
	The PoE model is defined as:
	\begin{align*}
	\expstate
	&\q \in \mathbb{R}^5,\\
	\exptrans
	&\bm{y}_1 = \trans{}{1}(\q)= \bm{F}_{R, \bm{x}}(\q) \in \mathbb{R}^2,\\
	&\bm{y}_2 = \trans{}{2}(\q)= \bm{F}_{L,\bm{x}}(\q) \in \mathbb{R}^2,\\
	\expexp
	&\bm{y}_1 \sim \mathcal{N}(\bm{\mu}_1^{(i)}, \sigma_1 \bm{I})\,|\, \textit{i }=  0, \dots, 2,\\
	&\bm{y}_2 \sim \mathcal{N}(\bm{\mu}_2, \sigma_2 \bm{I}),\\
	\expparam
	&\{\bm{\mu}_1^{(i)}\}_{i = 0,\dots, 2}, \bm{\mu}_2, \sigma_1, \sigma_2.
	\end{align*}
	$\bm{F}_{R, \bm{x}}$ and $\bm{F}_{L, \bm{x}}$ denote respectively the forward kinematics (position only) of the right (blue) and left (orange) arm. From the set of samples $\samples{\q}_n$, the position of the targets $\{\bm{\mu}_1^{(i)}\}_{i = 0,\dots, 2}, \bm{\mu}_2$ and their standard deviation $\sigma_1$ and $\sigma_2$ should be retrieved.
	
	We compare three approaches to learn the model. The first consists of maximum likelihood estimation of each expert separately, after applying the transformations to the dataset.
	With this approach, the main task is well understood. It is not the case of the secondary task, which is masked. The left end-effector displays a large variance, as shown in Fig.~\ref{fig:bimanualnullspacecomp}(a). This variance cannot be explained without taking into account the dependence between the two tasks and their prioritization.
	The second approach is to use a PoE with no prioritization between the objectives. The dependence between the tasks is taken into account, which results in a better understanding of the secondary objective. The prioritization is still not taken into account, which has two negative effects. As shown in Fig.~\ref{fig:bimanualnullspacecomp}(b). the targets of the right arm $\{\bm{\mu}_1^{(i)}\}_{i = 0,\dots, 2}$ are inferred further than they are, and the standard deviation of the secondary task $\sigma_2$ is slightly exaggerated.
	These two misinterpretations compensate the missing hierarchical structure by artificially increasing the importance of the primary task.
	In the third approach, the information of hierarchy is restored by using a product of experts with the nullspace structure (PoENS) as presented in Sec.~\ref{sec:poe_nullspace}. This time, the standard deviations of the two tasks as well as their respective targets are well retrieved.

	Quantitative results are produced in a slightly different manner than for the previous experiment. As only the gradient of the unnormalized log-likelihood of the PoENS is defined, they are first approximated using variational inference with a mixture of $K=50$ Gaussians.
    Results are reported in Table \ref{table:planar_nullspace} under task (a). As noticed in Fig.~\ref{fig:bimanualnullspacecomp}, the divergence is the smallest with the PoENS.
	
\begin{figure}
		\centering
		\includegraphics[width=1.\linewidth]{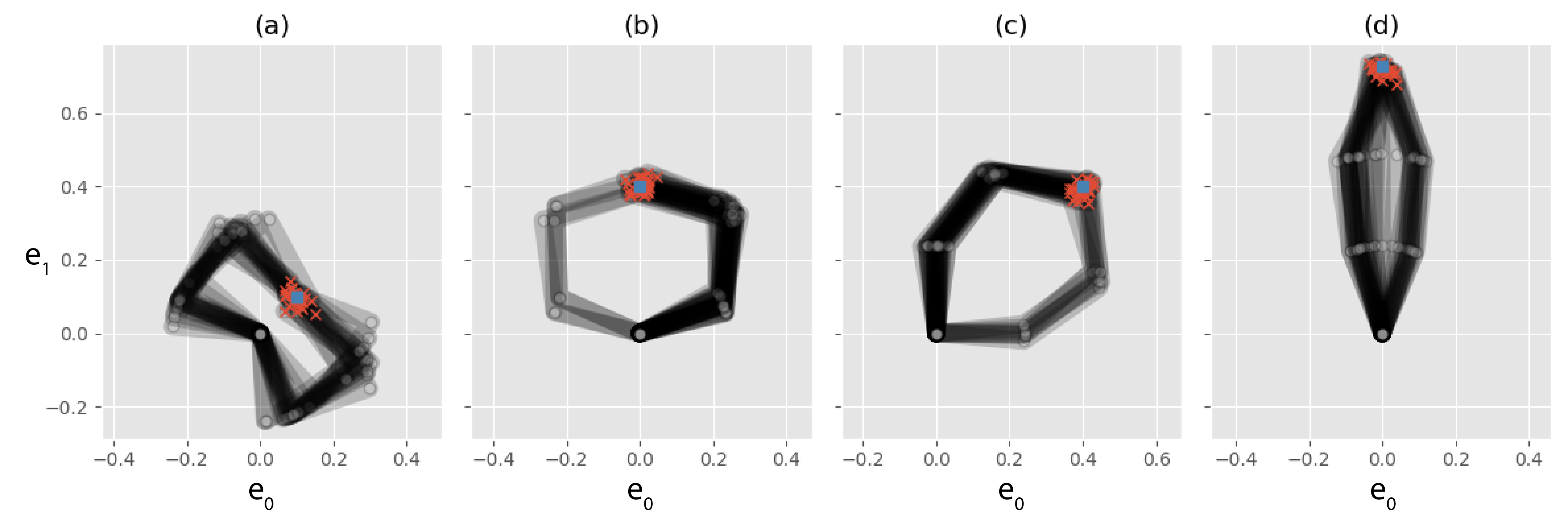}
		\caption{Dataset for the hierarchical task with a manipulability objective. The robot should track the target (displayed as a blue square) and maximize a manipulability measure. Four different targets are given.}
		\label{fig:mannullspacedata}
	\end{figure}

	In the second task with planar robots, we are interested in the manipulability measure presented in Sec.~\ref{sec:transformations}.
	The considered robot has three joints and its principal task is to track a point with its end-effector. One degree of freedom remains, which should be used to maximize the manipulability measure. This objective is set as a log-normal distribution on the determinant of the manipulability matrix (i.e., Gaussian distribution on the log). It defines which manipulability value $\mu_2$ to track, together with the allowed variation $\sigma_2$.
	In the same way as in the previous experiment, the end-effector should track four different targets $\{\bm{\mu}_1^{(i)}\}_{i=0,\dots, 3}$.
	For each of the four cases $i=0,\dots, 3$, $N=30$ independent samples $\samples{\q}_n$ are given, which are shown in Fig.~\ref{fig:mannullspacedata}.
	The PoE model is defined as:
	\begin{align*}
	\expstate
	&\q \in \mathbb{R}^3,\\
	\exptrans
	&\bm{y}_1 = \trans{}{1}(\q)= \bm{F}_{\bm{x}}(\q) \in \mathbb{R}^2,\\
	&y_2 = \trans{}{2}(\q)= \\ &\qquad \log\det\big(\bm{J}(\q)\bm{J}(\q)^{\trsp}\big) \in \mathbb{R}^+,\\
	\expexp
	&\bm{y}_1 \sim \mathcal{N}(\bm{\mu}_1^{(i)}, \sigma_1 \bm{I})\,|\, i = 0, \dots, 3,\\
	&y_2 \sim \mathcal{N}(\mu_2, \sigma_2),\\
	\expparam
	&\{\bm{\mu}_1^{(i)}\}_{i = 0,\dots, 3}, \mu_2,  \sigma_1, \sigma_2,
	\end{align*}
	where $\bm{F}(x)$ is the position of the final link.
	
	As before, we compare three ways of learning the model. As a baseline, we follow the approach of \cite{Jaquier20IJRR} by considering independent training of the experts. 
	We employ a less elaborated description of the manipulability task than in \cite{Jaquier20IJRR}, by considering the volume of the manipulability ellipsoid instead of the full ellipsoid. For our study, this simplified scalar descriptor is sufficient to showcase the advantages of PoENS over an independent training of the experts. 
	
\begin{figure}
		\centering
		\includegraphics[width=1.\linewidth]{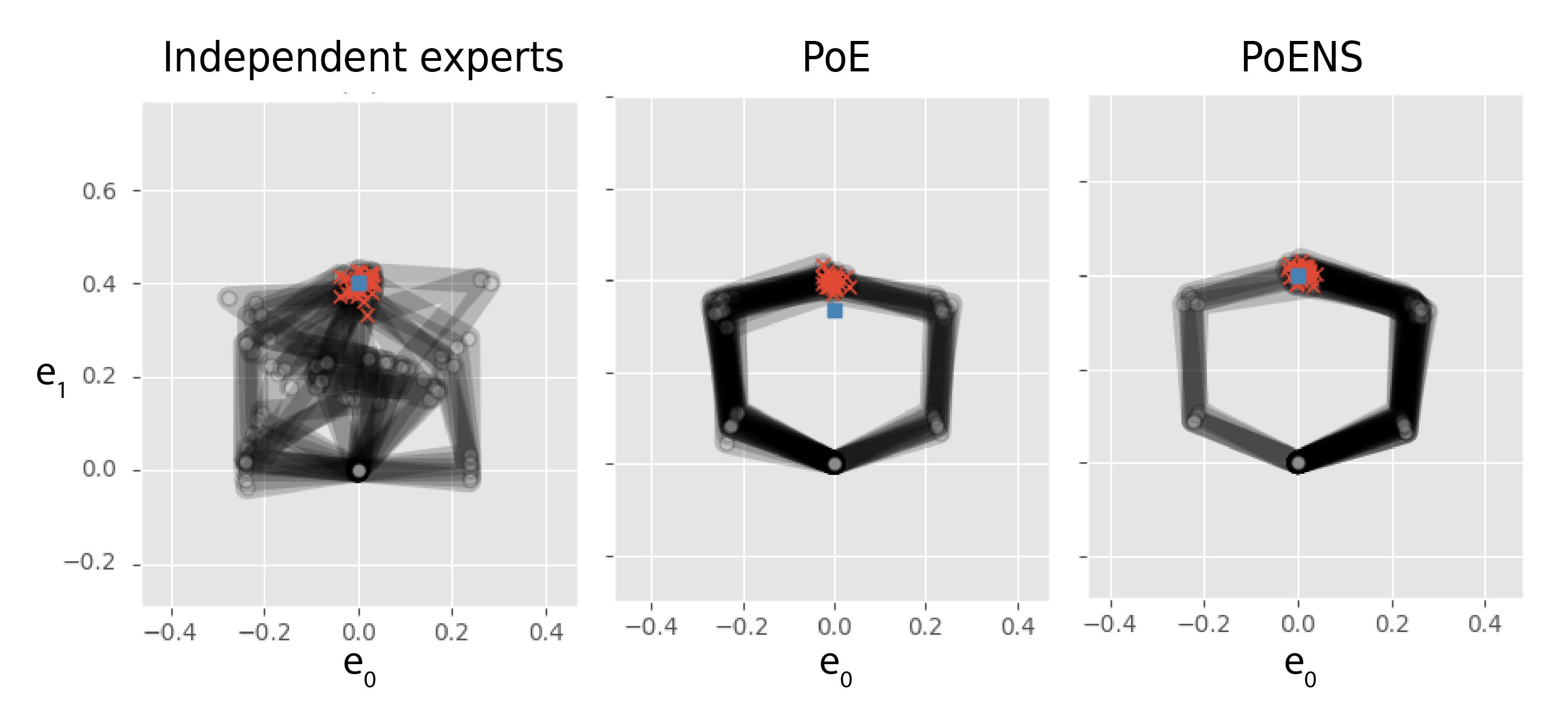}
		\caption{Samples from the different models learned for the task with a manipulability objective. \emph{Left:} Using independent experts, the secondary task is not well understood (the manipulability measure has a big variance in the dataset). \emph{Center:} Using a PoE, the estimation of the principal objective is biased. \emph{Right:} By providing the hierarchical structure, the two objectives are well understood.}
		\label{fig:mannullspacecomp}
	\end{figure}

	Fig.~\ref{fig:mannullspacecomp} presents samples generated from the different models (for a target as in Fig.~\ref{fig:mannullspacecomp}-\emph{(b)}). When the experts are learned independently (\emph{left}), the manipulability is not well understood. The samples are tracking a manipulability that is lower than expected, and the variance is overestimated. When training the PoE without the nullspace structure (\emph{center}), the distribution of generated samples better matches the dataset. However, the task-space target is not well estimated (it is inferred closer to the base of the robot to counterbalance the manipulability objective). In contrast, the samples retrieved from the PoENS (\emph{right}) show that the task-space target is well understood.

	Quantitative evaluations for a bimanual robot are also reported in Table \ref{table:planar_nullspace}.

	These results show that considering independent training of the experts has two limitations: (1) it does not exploit the dependence between the task-space position $\bm{F}(x)$ and the manipulability; and (2) it does not exploit the hierarchical structure that often relegates the manipulability objective to a secondary objective. When this structure is not exploited, the manipulability objective cannot be understood, as it is not directly observed. By maximum likelihood estimation of the log-normal distribution on the dataset, the variance is high and the mean value does not reflect the fact that we were targeting the maximum of manipulability. For example, the demonstration data in Fig.~\ref{fig:mannullspacedata}-\emph{(d)} are characterized by very small manipulability ellipsoids, which would be interpreted erroneously if modeled independently from the tracking task.
	
	\begin{table}
		\begin{center}
			\caption{Quantitative evaluations of the quality of the learned distribution for the two planar tasks. The table shows alpha-divergence $D_{\alpha=1/2}$ measures between the different approximations and the ground-truth distribution are computed. \textit{(a)} Bimanual task \textit{(b)} Manipulability objective}
			\begin{tabular}{p{25mm}ll}
				\toprule Task 
				& (a) & (b)\\ 
				\midrule 
				Independent experts
				& 1.814 $\pm$ 0.055& 0.812 $\pm$ 0.117\\ 
				\midrule 
				PoE 
				& 0.258 $\pm$ 0.101& 0.630 $\pm$ 0.086\\ 
				\midrule 
				\textbf{PoENS}
				& \textbf{0.094 $\pm$ 0.024} & \textbf{0.202 $\pm$ 0.067}\\ 				
				\bottomrule 
				\label{table:planar_nullspace}
			\end{tabular}
		\end{center}
	\end{table}
	
\begin{figure*}
		\centering
		\includegraphics[width=1.\linewidth]{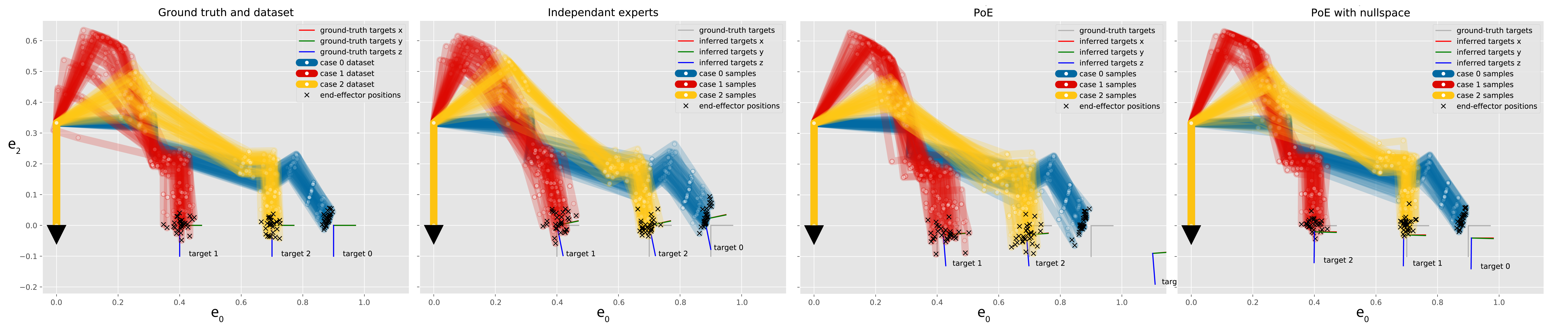}
		\caption{Dataset and retrieved samples for the hierarchical task with the 7-DoF Panda robot (welding task). Only the kinematic chain of the robot is displayed, such that multiple samples can be shown. The dataset provides three different situations (displayed in blue, yellow and red) in which the robot should track a different target with its end-effector. In each situation, the orientation should be held vertical as a secondary task.}
		\label{fig:orientationnullspace001}
	\end{figure*}

	\paragraph{\textbf{Welding task}}
	We conduct a similar experiment with a 7-DoF Panda robot. The dataset mimics a welding task, in which the position of the end-effector is more important than its orientation. The robot is supposed to track the position of the component to weld while preferably keeping its orientation vertical. We give three sets of $N=30$ independent samples $\samples{\q}_n$ for three different target positions $\{\bm{\mu}_{x}^{(i)}\}_{i = 0, \dots, 3}$ as shown in Fig.~\ref{fig:orientationnullspace001}. The orientation should be kept fixed (the same in the three cases).
	The PoE model is defined as:
	\begin{align*}
	\expstate
	&\q \in \mathbb{R}^7,\\
	\exptrans
	&\bm{y}_1 = \trans{}{1}(\q)= \bm{F}_{x}(\q) \in \mathbb{R}^3,\\
	&\bm{y}_2 = \trans{}{2}(\q)= \mathrm{vec}(\bm{F}_{R}(\q)) \in \mathbb{R}^9,\\
	&\bm{y}_3 = \trans{}{3}(\q)= \q \in \mathbb{R}^7,\\
	\expexp
	&\bm{y}_1 \sim \mathcal{N}(\bm{\mu}_{x}^{(i)}, \sigma_{x} \bm{I})\,|\, i = 0, \dots, 3,\\
	&\bm{y}_2 \sim \mathcal{N}(\bm{\mu}_{R}, \sigma_R \bm{I}),\\
	&\bm{y}_3 \sim \mathcal{N}(\bm{\mu}_{\q}, \sigma_{\q} \bm{I}),\\
	\expparam
	&\{\bm{\mu}_{x}^{(i)}\}_{i = 0, \dots, 3}, \sigma_{x}, \bm{\mu}_R, \sigma_R, \bm{\mu}_{\q}, \sigma_{\q}
	\end{align*}
	where $\bm{F}_{x}(\q)$ is the position of the end-effector and $\mathrm{vec}(\bm{F}_{R}(\q))$ its vectorized rotation matrix. We choose a Gaussian distribution for the vectorized rotation matrix with an isotropic covariance matrix $\sigma_R \bm{I}$. 
	As the renormalization arises at the joint level in a PoE, this choice is acceptable even if the rotation expert does not have the proper support.

	We compared the same three ways of training the model. As expected, with independent experts, a mean orientation is computed, resulting in biased estimation of this objective (see Fig.~\ref{fig:orientationnullspace001} second column). The PoE without hierarchy performs a bit better, as it can understand the dependence between the position and the orientation induced by the kinematics. The estimation of the orientation is still biased (see Fig.~\ref{fig:orientationnullspace001} third column). As in previous experiments, the PoENS performs the best. The secondary objective of orientation is clearly understood.
	
	For the quantitative experiments, we compare the dataset with the learned distributions. This comparison is done in each of the 3 situations and by using the maximum mean discrepancy\footnote{with an unbiased estimate of the kernel mean, it is possible to get negative values. Negative values indicates a close to zero discrepancy; their absolute value can be discarded.} $\mathrm{MMD}_{u}^2$, see \cite{gretton2012kernel}.
	The discrepancy is computed with 500 samples and a RBF kernel with $\gamma=0.1$. The results are reported in Table \ref{table:mmd_position_orientation} and in Fig.~\ref{fig:orientationnullspace001}, where PoENS shows much better results.
	\begin{table}
		\begin{center}
			\caption{Quantitative evaluations of the quality of the learned distribution for the 7-DoF Panda robot (welding task). The table shows maximum mean discrepancy $\mathrm{MMD}_{u}^2$ measures between the dataset and different models for the 3 cases displayed in Fig.~\ref{fig:orientationnullspace001}.}
			\begin{tabular}{p{25mm}lll}
				\toprule Case
				& (0) & (1) & (2)\\ 
				\midrule 
				Independent experts &
				$\expnumber{1.3}{-3}$  & $\expnumber{1.3}{-2}$ &  $\expnumber{8.2}{-3}$\\ 
				\midrule 
				PoE  &
				$\expnumber{7.0}{-4}$
				& $\expnumber{1.7}{-3}$ & $\expnumber{1.4}{-3}$\\ 
				\midrule 
				\textbf{PoENS} &
				$\bm{\expnumber{-9.8}{-6}}$ & $\bm{\expnumber{-8.5}{-5}}$ & $\bm{\expnumber{4.5}{-4}}$ \\ 				
				\bottomrule 
				\label{table:mmd_position_orientation}
			\end{tabular}
		\end{center}
	\end{table}
	
\begin{figure}
		\centering
		\includegraphics[width=0.8\linewidth, trim=0 50.0 0 90.0, clip]{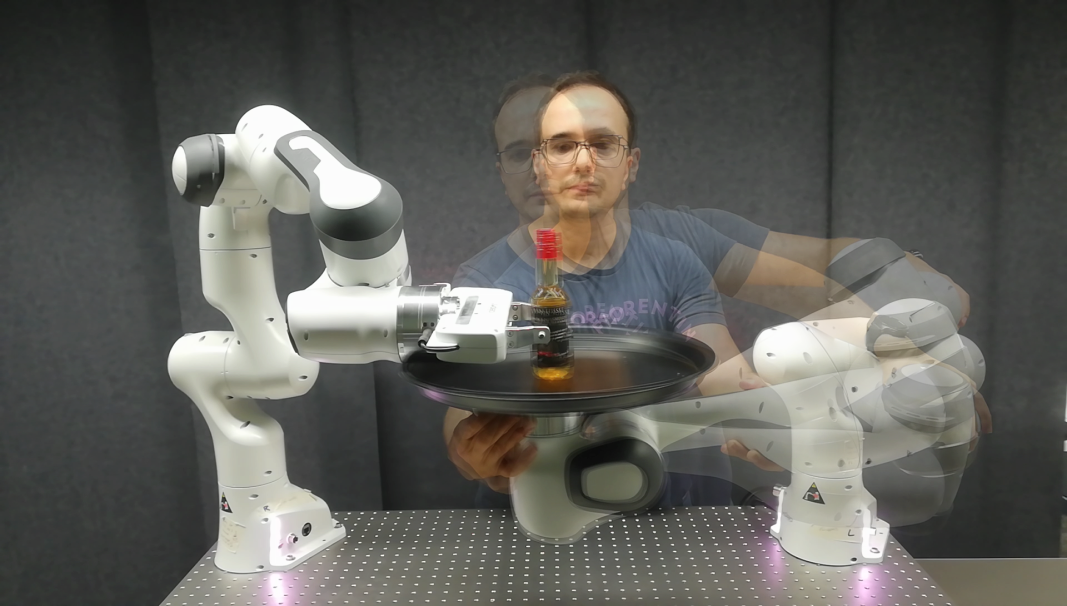}
		\vspace{0.1cm}
		\includegraphics[width=0.8\linewidth, trim=0 50.0 0 90.0, clip]{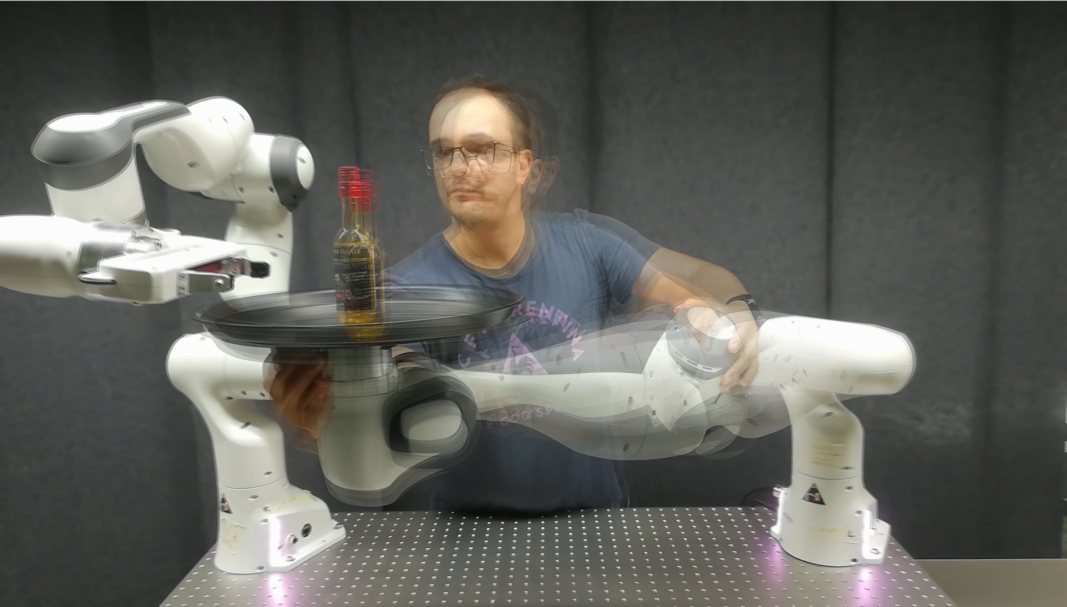}
		\caption{Several demonstrations of the desired priority behavior are given.\textbf{Top:} Both position and orientation are achievable. \textbf{Bottom:} Position is only partially fulfilled, to allow the tray to remain parallel to the floor. The left robot arm is playing the role of a person who should grasp the bottle.}
		\label{fig:waiter-robot-demos}
	\end{figure}

	\paragraph{\textbf{Robot waiter task}}
	In this experiment we consider the task of serving a drink on a tray (Fig. \ref{fig:waiter-robot-demos}). The robot needs to move the tray in response to the position of a person such that it stays parallel to the floor (\textit{orientation sub-task}) and the bottle is within the person's hand (\textit{position sub-task}). However, in order to avoid spilling the drink, the position sub-task should only be performed if the orientation sub-task is fulfilled. 
	Here we show that, by using PoENS, both  priorities and  task references can be learned from demonstrations. This improves on a similar experiment conducted in \cite{silverio2018learning} in which we could teach the strict priority hierarchies but the references (desired position and orientation of the end-effector) were taught separately. For evaluation purposes, we employ an additional gravity-compensated robot to play the role of the person (the left arm in Fig. \ref{fig:waiter-robot-demos}).
	
	We collected samples $\hat{\bm{q}}_n$ for three different positions of the left hand, with $N=10$. In two of the cases, both position and orientation sub-tasks can be achieved (Fig. \ref{fig:waiter-robot-demos}, top). In the other one, only the orientation task is fulfilled (Fig. \ref{fig:waiter-robot-demos}, bottom), with the robot doing its best to track the gripper position. The used PoE model was the same as in the previous \textit{welding task} experiment, except for the expert $\bm{y}_1$ that is now the end-effector position with respect to the gripper coordinate system.

\begin{figure*}
		\centering
		\includegraphics[width=1.\linewidth]{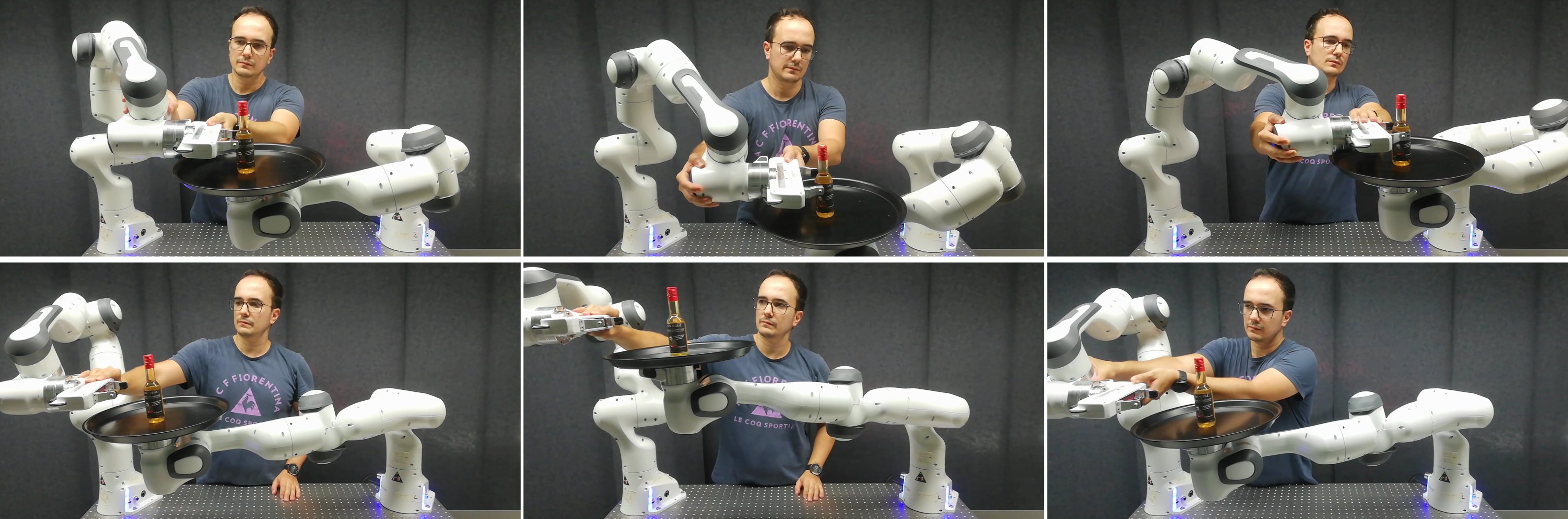}
		\caption{The right arm should keep the tray horizontal and the bottle between the gripper fingers. With a PoENS model, the robot learns \textit{i)} the tray pose with respect to the gripper and \textit{ii)} that the tray orientation has a higher priority than the position. \textbf{Top row:} Reproduction when both position and orientation can be fulfilled. \textbf{Bottom row:} Reproduction when position reference is too far so the robot prioritizes orientation.}
		\label{fig:waiter-robot-repros}
	\end{figure*}

	Figure \ref{fig:waiter-robot-repros} shows the obtained experimental results (a video of the complete experiment is available at \url{https://sites.google.com/view/poexperts}). We observe that the robot respects the demonstrated priorities and learns the end-effector pose correctly. The controller was computed with \eqref{equ:lqr_controller_poe}, using the dynamical system \eqref{equ:augmented_system_lqr} augmented with the null space projection matrix of the identified hierarchy.
	
	
	\subsection{Learning transformations and conditional PoEs}
	\label{sec:experiments:generalization}
	In these experiments, we compare PoEs with less structured ways to learn distributions. 
	We show that PoEs, by providing more structure, require fewer samples to be trained.
	They also allow finding simple explanations for complex distribution, leading to quicker generalization.
	Unlike the previous experiments, the experts transformations are this time only partially known.
	
\begin{figure}
		\centering
		\includegraphics[width=1.\linewidth]{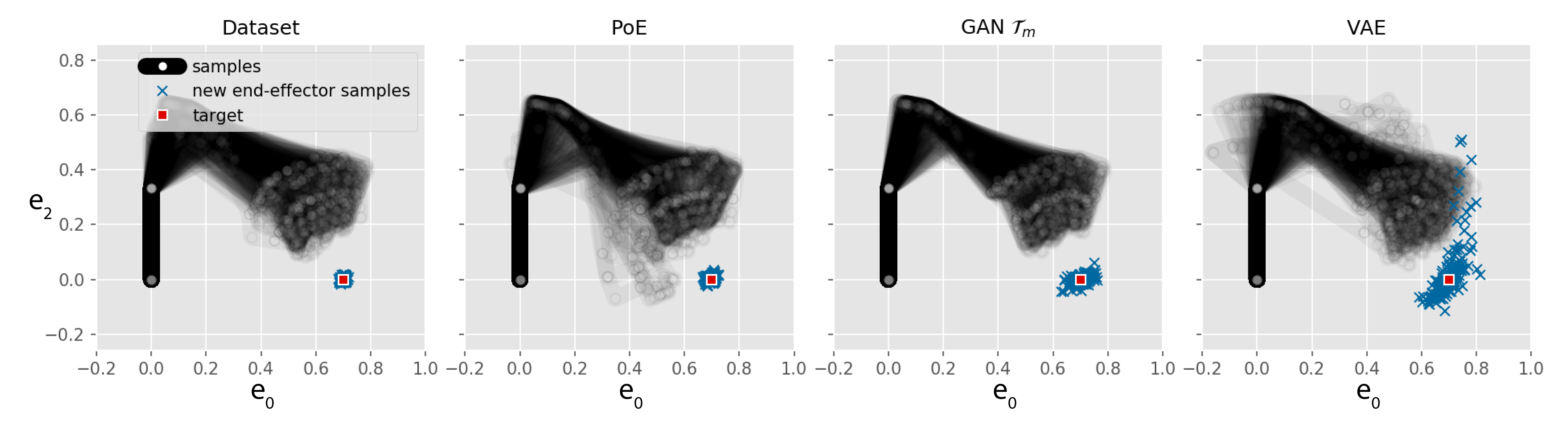}
		\caption{Dataset and samples from the different models for a task involving an unknown end-effector. In this illustration, the models were trained with 3000 samples.}
		\label{fig:new_ee_comp}
	\end{figure}

	\paragraph{\textbf{New end-effector}}
	We consider an experiment with the 7-DoF Panda robot in which a new end-effector is defined. The new end-effector has a fixed position $\bm{d}$ in the coordinate system of the known end-effector. 
	We want to learn the distribution of configurations where the new end-effector follows a Gaussian distribution around $\bm{\mu}_{x}$, a given task-space position target. To compare the data-efficiency of the models, the datasets are composed of $N \in \{3, 10, 30, 100, 3000\}$ independent samples, as illustrated in Fig.~\ref{fig:new_ee_comp}-\emph{left}.
	The PoE model is defined as:
	\begin{align*}
	\expstate
	&\q \in \mathbb{R}^7,\\
	\exptrans
	&\bm{y}_1 = \trans{}{1}(\q) = \bm{F}_{R}(\q) \bm{d} + \bm{F}_{x}(\q) \\
	&\hspace{16.5mm} = \bm{F}_{d}(\q) \in \mathbb{R}^3,\\
	&\bm{y}_2 = \trans{}{2}(\q)= \q\in \mathbb{R}^7,\\
	\expexp
	&\bm{y}_1 \sim \mathcal{N}(\bm{\mu}_{x}, \sigma_{x} \bm{I}),\\
	&\bm{y}_2 \sim \mathcal{N}(\bm{\mu}_{\q}, \sigma_{\q} \bm{I}),\\
	\expparam
	&\bm{d}, \bm{\mu}_{x}, \sigma_{x}, \bm{\mu}_{\q}, \sigma_{\q},
	\end{align*}
	where $\bm{F}_{R}(\q)$ is the rotation matrix and $\bm{F}_{x}(\q)$ the position of the known end-effector. The second expert learns a preferred joint configuration.
	We compare the PoE with five other techniques to learn distributions. The first is a variational autoencoder \cite{kingma2013auto}. 
As architecture for both the encoder and the decoder, we used 2 layers of 20 fully connected hidden units and $\tanh$ activation. The latent space was of 4 dimensions, which corresponds to the remaining DoFs of the robot, after constraining 3 DoFs with the position of the new end-effector. The second technique considered is a Dirichlet process Gaussian mixture model \cite{bishop2006pattern}. 
In this model, the number of Gaussian components (with full covariances) is learned according to the number of datapoints and the complexity of the distribution. Another model tested is a transformed distribution using an invertible neural network (NVP) as in \cite{dinh2017density}. As architecture, four layers of 108 hidden units with $\mathrm{relu}$ activation were used. The two last models are GANs \cite{goodfellow2014generative}. In these two cases, both the generator and discriminators are multilayer perceptrons with 3 layers of 50 hidden units each and sigmoid activation. The latent space of the generator is of 10 dimensions. In the second GAN, that we denote "GAN $\trans{}{m}$", the discriminator is helped by accessing the samples and demonstrations through the different task-space transformations.
	
	\begin{figure}
		\centering
		\includegraphics[width=1.\linewidth]{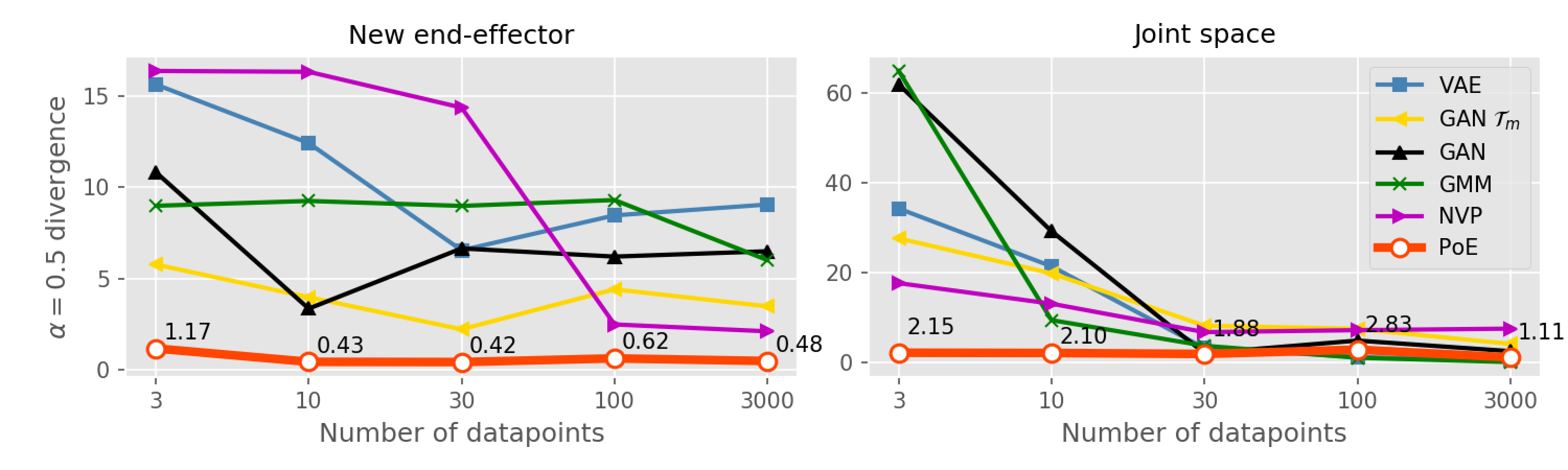}
		\caption{Quantitative results for a task involving an unknown end-effector. The graphs show alpha-divergence measures (with $\alpha=1/2$) between the dataset and the different models for the different sizes of the dataset.}
		\label{fig:new_ee_graph}
	\end{figure}
	
	For each of the six techniques, the model was learned with a different number of samples. Samples from the PoE, the VAE and the GAN $\trans{}{m} $are shown in Fig.~\ref{fig:new_ee_comp}, where the whole dataset was used to train the models. For each model, 500 samples were generated, for which the position of the new end-effector was computed. We compared the distribution of end-effector positions between each model and the dataset using maximum mean discrepancy $\mathrm{MMD}_{u}^2$. The results are reported in Fig.~\ref{fig:new_ee_graph}. The PoE performs better than the others for each number of datapoints. Its performance is also less influenced by this number. Its advantage can be explained by the difficulty to represent the distribution in configuration space, which has a complex shape. Since the target is also very precise, this distribution is close to a 4 dimensional manifold embedded in a 7-dimensional space, which makes it particularly difficult to be encoded as a mixture of Gaussians. A Gaussian approximating a part of this manifold will have an important portion of its density outside of the manifold, resulting in an imprecise tracking.
	Only a high number of Gaussian components can approximate well this distribution.
	Moreover, training such a model requires huge datasets to cover well the entire manifold. 
	
	It is interesting to notice that GAN $\trans{}{m}$ performs quite well. In Fig.~\ref{fig:new_ee_comp}, samples from the new end-effector are tracking the target better with GAN $\trans{}{m}$ than with VAE. The standard GAN was not displayed here but has very similar performance as the VAE. Unlike VAEs, GANs are very appropriate to exploit the existence of task spaces as hand-engineered features of the discriminator. To our knowledge, they are the best alternative to PoEs if more data is available.

	\paragraph{\textbf{Conditional distributions}}
	This experiment is similar to the previous one, but requires a generalization to different targets for the new end-effector. The dataset is split into 3 cases with $N=1000$ datapoints for each. Unlike in the previous experiment, the targets $\{\bm{\mu}_{x}^{(i)}\}_{i=0}^{2}$ are given.
	\begin{figure}
		\centering
		\includegraphics[width=1.\linewidth]{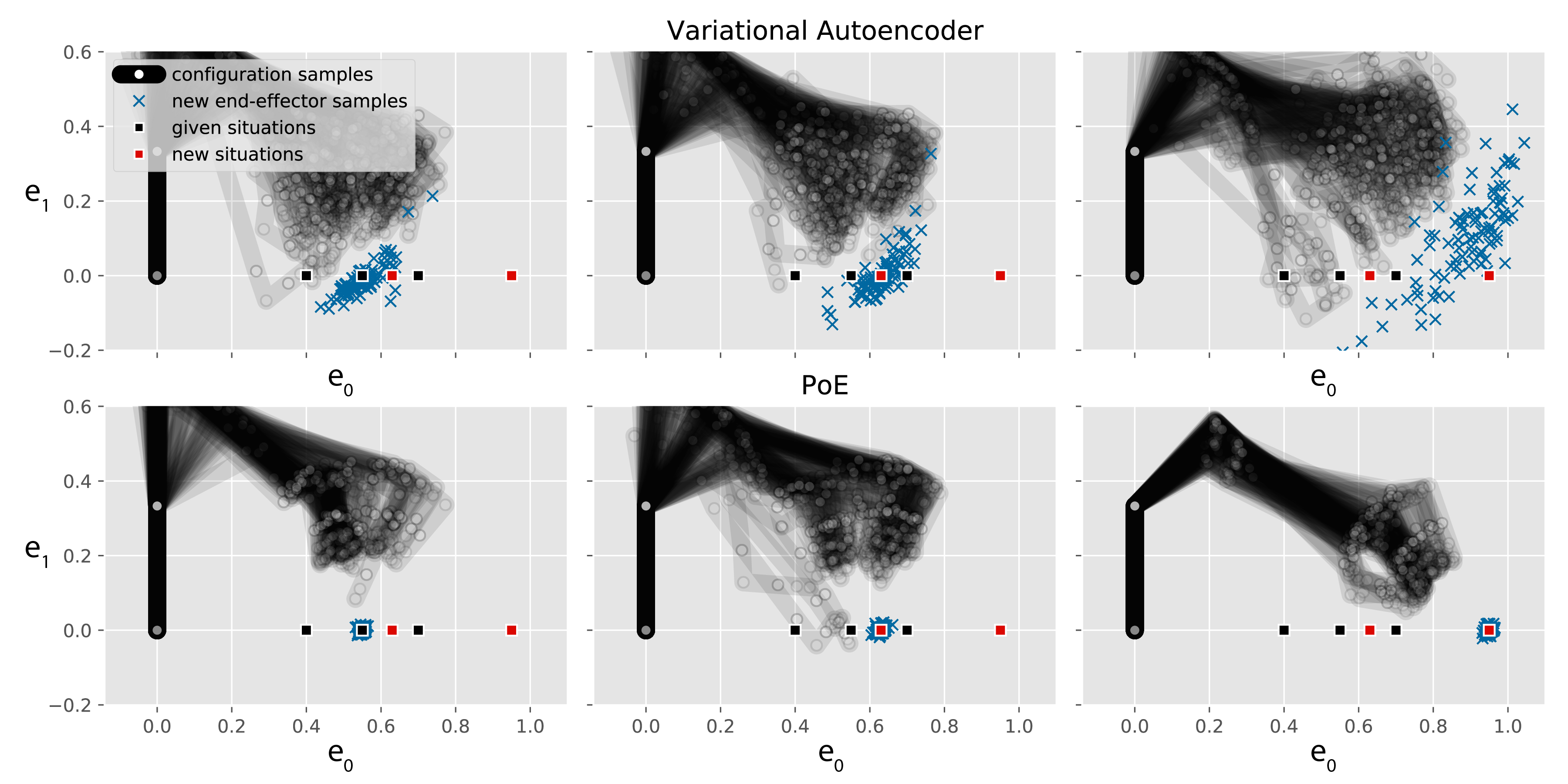}
		\caption{Dataset and samples from the different models for a task involving an unknown end-effector. Three cases are reproduced. \emph{Left column:} The distribution is sampled on a known target. \emph{Middle column:} The target is in-between two given targets. \emph{Right column:} The target is far from the given targets. The first and second rows show results for a conditional variational autoencoder and a PoE, respectively.}
		\label{fig:neweevaepoecond}
	\end{figure}
	The PoE model is defined as before, with:
	\begin{align*}
	\expstate
	&\q \in \mathbb{R}^7,\\
	\exptrans
	&\bm{y}_1 = \trans{}{1}(\q)= \bm{F}_{R}(\q) \bm{d} + \bm{F}_{x}(\q) \\
	&\hspace{16.5mm} = \bm{F}_{d}(\q) \in \mathbb{R}^3,\\
	&\bm{y}_2 = \trans{}{2}(\q)= \q\in \mathbb{R}^7,\\
	\expexp
	&\bm{y}_1 \sim \mathcal{N}(\bm{\mu}_{x}^{(i)}, \sigma_{x} \bm{I})\,|\, i = 0, \dots, 3,\\
	&\bm{y}_2 \sim \mathcal{N}(\bm{\mu}_{\bm{q}}, \sigma_{\bm{q}} \bm{I}),\\
	\expparam
	&\bm{d}, \sigma_{x}, \bm{\mu}_{\bm{q}}, \sigma_{\bm{q}}.
	\end{align*}
	
	For the evaluations, we used a conditional variational autoencoder (VAE), where the targets  $\{\bm{\mu}_{x}^{(i)}\}_{i=0}^{2}$ are concatenated to the corresponding datapoints at the entrance of the encoder and also concatenated to the latent variables at the entrance of the decoder. For the Gaussian mixture model, we encode the joint distribution of configurations $\q$ and targets and used conditioning over new targets to retrieve the corresponding distribution of states, similarly to Gaussian mixture regression (GMR). We evaluate the quality of the distribution with $\mathrm{MMD}_{u}^2$ in three cases. In case (a) (Fig.~\ref{fig:neweevaepoecond}-\emph{left}), the target is one already given in the dataset. In case (b) (Fig.~\ref{fig:neweevaepoecond}-\emph{center}), the target is between two given targets. In (a) and (b), VAE performs quite well, with the same limitations as in the previous experiments.
	In the last case (Fig.~\ref{fig:neweevaepoecond}-\emph{right}), the target is far outside and the performance of VAE further reduced. The PoE performs very well in the three cases, as shown also in the quantitative evaluation reported in Table \ref{table:new_ee_eval_cond}. The only drawback is that it requires some time to approximate the distribution given the new target, as only the unnormalized density of the PoE is directly accessible. Note that this is not a problem if the log-pdf is used as a reward function in optimal control, as proposed in Sec.~\ref{sec:applications}.
	
	\begin{table}
		\begin{center}
			\caption{Quantitative results for the task with the new end-effector. The table shows maximum mean discrepancy $\mathrm{MMD}_{u}^2$ measures between the dataset and the different models for the different cases. (a) The distribution is sampled on a known target. (b) The target is in-between two given targets. (c) Generalization with respect to a new target far from the given targets.}
			\begin{tabular}{p{20mm}lll}
				\toprule Case
				& (a) & (b) & (c) \\ 
				\midrule 
				VAE &
				$\expnumber{8.8}{-4}$ &
				$\expnumber{2.2}{-4}$ &
				$\expnumber{1.0}{-2}$ \\
				\midrule 
				GMR &
				$\expnumber{2.7}{-6}$ &
				$\expnumber{5.1}{-6}$ &
				$\expnumber{9.0}{-2}$ \\
				\midrule 
				\textbf{PoE}  &
				$\bm{\expnumber{3.5}{-9}}$ & 
				$\bm{\expnumber{-2.2}{-7}}$ &
				$\bm{\expnumber{3.4}{-7}}$ \\ 
				\bottomrule 
				\label{table:new_ee_eval_cond}
			\end{tabular}
		\end{center}
	\end{table}
	

	\subsection{End-effector position and rotation correlations}
	\label{sec:experiments:pos_rot_correlation}
	In this last experiment, we show how our framework can help to cope with orientation statistics. Typical distributions, such as matrix Bingham---von Mises---Fisher distribution (BMF) are hard to use, because of their intractable normalizing constant. In PoEs, the normalization happens in the configuration space, which makes the use of intractable density possible. Another tempting approach is to use a multivariate normal distribution of vectorized rotation matrices. When training with maximum likelihood estimation, this leads to bad approximations. Indeed, the wrong normalizing constant is considered because the space of rotation matrices is discarded. We show that using such vectorized distributions is fine with the PoE as the normalizing constant is computed in the product space. Also, in some cases, the Bingham---von Mises---Fisher distribution can be computed as a Gaussian on the vectorized matrix (see Sec.~\ref{subsec:distributions}).
	
\begin{figure}
		\centering
		\includegraphics[width=0.9\linewidth]{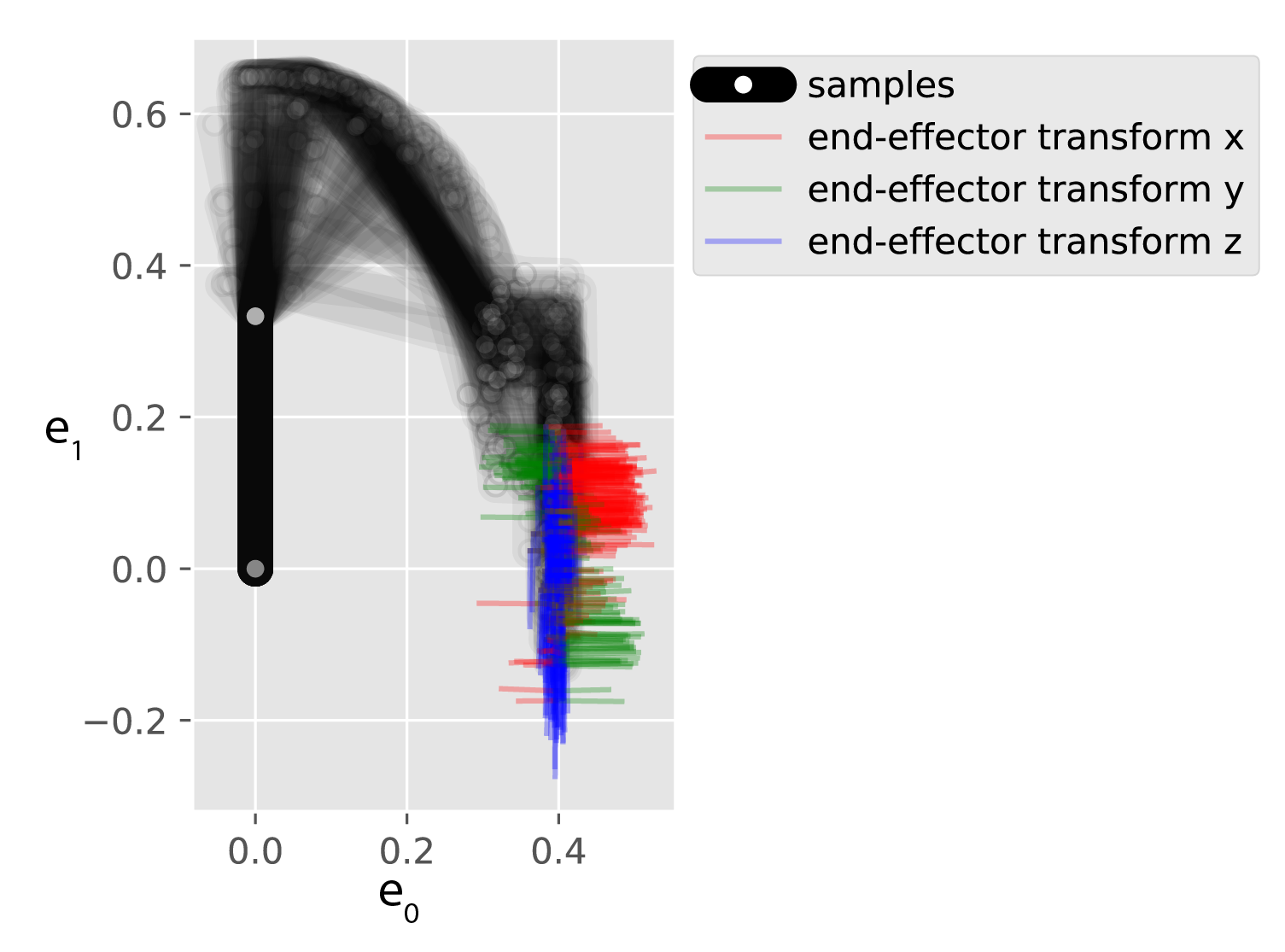}
		\caption{Samples where the height of the end-effector is correlated with its rotation along the $e_1$-axis. The transformation matrix of the end-effector is displayed.}
		\label{fig:poe_rot_loc_corr}
	\end{figure}

	We produced three datasets, composed each one of $N=3000$ independent samples from a ground-truth PoE model. The number of samples is much higher than required to understand the task but reduces the variance of the estimation of the parameters to provide precise comparisons. The PoE model used to sample has a correlation between the height of the end-effector ($e_2$) and the rotation along this axis. In the first case (a), the correlation is induced by penalizing deviations from a linear relationship between elements of the rotation matrix and the height of the end-effector. This dataset is shown in Fig.~\ref{fig:poe_rot_loc_corr}.
	In the two other cases (b) and (c), the correlation is induced with the $z$ Euler angle instead of the rotation matrix. In (b) and (c), the standard deviations of this angle is $0.2$ and $1.$, respectively.
	The PoE model is defined as: 
	\begin{align*}
	\expstate
	&\q \in \mathbb{R}^7,\\
	\exptrans
	&\bm{y}_1 = \trans{}{1}(\q)= \bm{F}_{\bm{x}, \vect{R}}(\q) \in \mathbb{R}^{12},\\
	\expexp
	&\bm{y}_1 \sim \mathcal{N}(\bm{\mu}, \mathrm{diag}(\bm{ \sigma}) + \bm{L}\bm{L}^T),\\
	\expparam
	&\bm{\mu}, \sigma, \bm{L},
	\end{align*}
	where $\bm{y}_1$ is a concatenation of the position of the end-effector and its vectorized rotation matrix. The expert is a Gaussian with a structured covariance matrix. The covariance is composed of a diagonal component $\mathrm{diag}(\bm{\sigma})$,  whose diagonal is the vector $\bm{\sigma}$ and a low-rank component $\bm{L}\in \mathbb{R}^{12 \times n_{\mathrm{axis}}}$ that encodes correlations. In this experiment, $n_{\mathrm{axis}}$ is chosen to be 1 because there is only one axis of covariance. The covariance matrix and the mean $\bm{\mu}$ can also be parametrized to have a marginal distribution of rotations in the form of a matrix Bingham---von Mises---Fisher distribution, a marginal distribution of position as a Gaussian and a covariance between the two (see \eqref{equ:mvnmbmfjoint}).
	
\begin{figure}
		\centering
		\includegraphics[width=1.\linewidth]{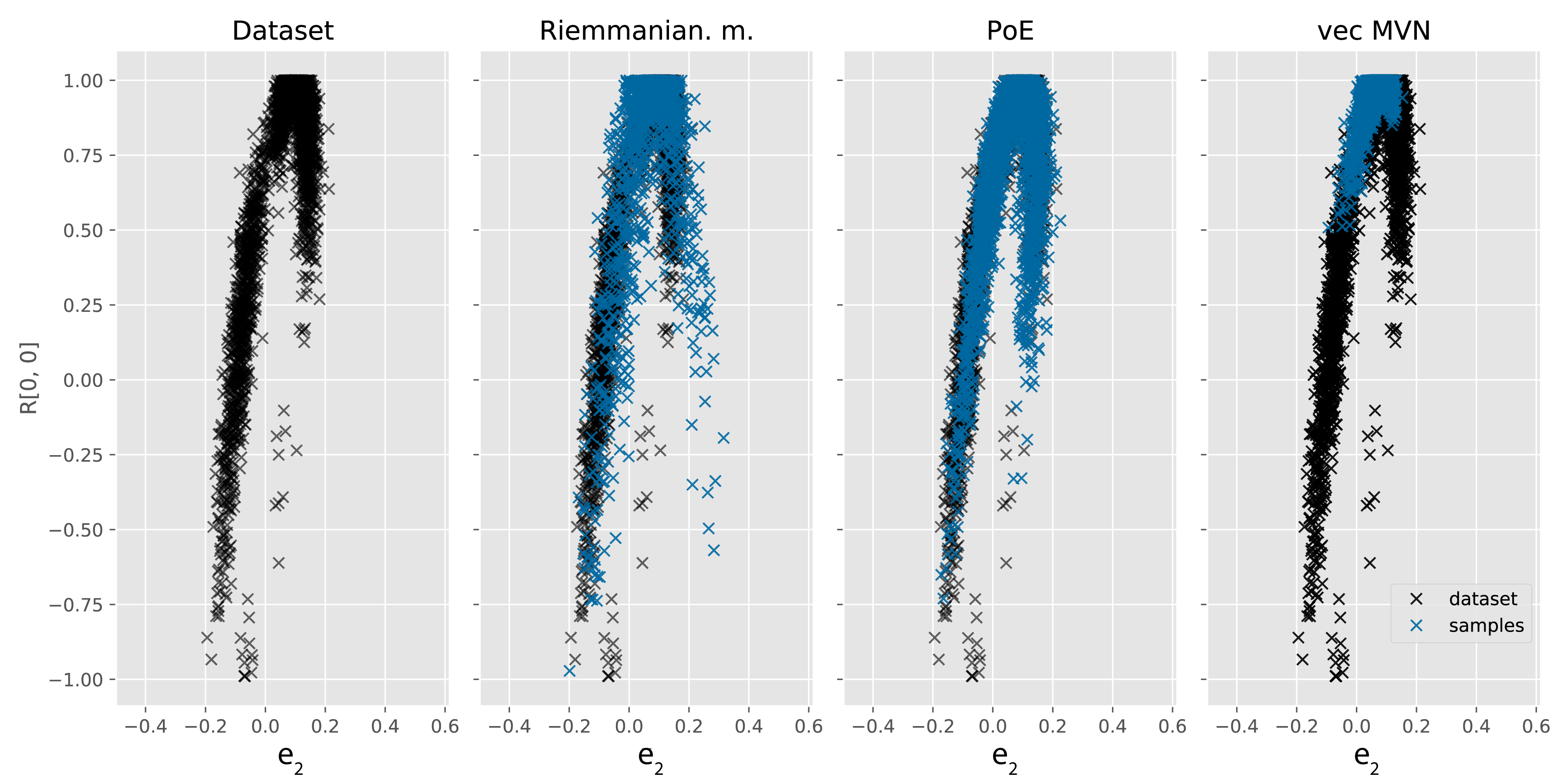}
		\caption{Scatter plot showing the correlation between the height of the end-effector and the first element of the first row of the rotation matrix for case (a). The dataset is shown in black and the samples from the different models in blue.}
		\label{fig:poe_rot_loc_corr_scatt}
	\end{figure}

	We compare our approach with two others, where only the concatenated position and orientation $\bm{y}_1$ is taken into account and without exploiting the kinematic structure of the robot. In the first alternative, we perform maximum likelihood estimation of a Gaussian with $\bm{y}_1$. In the second, we use a Gaussian on a Riemannian manifold as in \cite{zeestraten2017approach}. The rotation matrices are converted to quaternions and the considered manifold is the product between a Euclidean and spherical 3-manifold. For the quantitative evaluation, we sampled 500 points from each model and computed $\mathrm{MMD}_{u}^2$ with the dataset. Results are reported in Table \ref{table:rot_loc_corr_comp}. The vectorized Gaussian is the worst-performing model. Its best performance is in case (b), when the standard deviation of the angle is small, making the Euclidean approximation more valid. The PoE performs better in all cases. 
As expected, its advantage is the biggest in case (a), where the data was actually generated with a correlation between elements of the rotation matrix and height of the end-effector. Figure \ref{fig:poe_rot_loc_corr_scatt} shows the correlation between the height of the end-effector and the first element of the first row of the rotation matrix for case (a).
	
	\begin{table*}
		\begin{center}
			\caption{Quantitative results for the task with a correlation between position and orientation. The table shows $p_\mathrm{value}$ and maximum mean discrepancy $\mathrm{MMD}_{u}^2$ measures between the dataset and the different models for the different cases.}
			\begin{tabular}{p{35mm}ll|ll|ll}
				\toprule Case
				& \multicolumn{2}{c}{(a)} & \multicolumn{2}{c}{(b)} & \multicolumn{2}{c}{(c)}\\ 
				& $p_\mathrm{value}$ & $\mathrm{MMD}_{u}^2$ & $p_\mathrm{value}$ & $\mathrm{MMD}_{u}^2$ & $p_\mathrm{value}$ & $\mathrm{MMD}_{u}^2$  \\ 
				\midrule 
				Vectorized Gaussian
				& 0.0001 & $\expnumber{2.0}{-2}$ &  0.10 & $\expnumber{2.15}{-4}$ & 0.0011 &  $\expnumber{1.0}{-2}$\\ 
				\midrule 
				Riemannian manifold 
				& 0.10 & $\expnumber{1.03}{-3}$ & 0.17 & $\expnumber{1.5}{-4}$ & 0.29 & $\expnumber{2.03}{-4}$\\ 
				\midrule 
				\textbf{PoE}
				& $\bm{0.44}$ & $\bm{\expnumber{1.5}{-3}}$ & $\bm{0.38}$ & $\bm{\expnumber{-3.9}{-5}}$ & $\bm{0.68}$ & $\bm{\expnumber{-1.04}{-3}}$ \\ 
				\bottomrule 
				\label{table:rot_loc_corr_comp}
			\end{tabular}
		\end{center}
		\label{table:poe_nullspace}
	\end{table*}
	
	%
	%
	%
	%
	
	\section{Conclusion}
	We proposed a framework based on products of experts to encode distributions in robotics. We demonstrated the pertinence of the model in various applications, and showed that this framework can be linked to many existing learning from demonstration representations and methods. 
	By incorporating robot knowledge as transformations, we showed that such approach is more data-efficient than general approaches like variational autoencoder. 
	We also discussed the promises that such approach hold to tackle wide-ranging data problems in robotics, by providing a framework that can start learning from few trials or demonstrations, but that it rich enough to progress once more data are available. Indeed, since PoEs offer the flexibility to learn transformations as neural networks, the approach can span a wide range of problems, from small datasets with significant a priori knowledge to bigger datasets with less structured models. 
	In the experiments, we validated that PoEs offer substantial improvements over approaches in which models are learned separately, emphasizing the capability of the approach to uncover tasks masked by kinematic limitations or by the resolutions of higher-level objectives. 
	
	\begin{acks}
		The research leading to these results has received funding from the European Commission's Horizon 2020 Programme through the MEMMO Project (Memory of Motion, http://www.memmo-project.eu/, grant agreement 780684) and COLLABORATE project (https://collaborate-project.eu/, grant agreement 820767).
	\end{acks}
	\bibliographystyle{SageH}
	\bibliography{article_sage.bbl}

\begin{thebibliography}{69}
\providecommand{\natexlab}[1]{#1}
\providecommand{\url}[1]{\texttt{#1}}
\providecommand{\urlprefix}{URL }
\expandafter\ifx\csname urlstyle\endcsname\relax
  \providecommand{\doi}[1]{DOI:\discretionary{}{}{}#1}\else
  \providecommand{\doi}{DOI:\discretionary{}{}{}\begingroup
  \urlstyle{rm}\Url}\fi

\bibitem[{Abadi et~al.(2016)Abadi, Barham, Chen, Chen, Davis, Dean, Devin,
  Ghemawat, Irving, Isard, Kudlur, Levenberg, Monga, Moore, Murray, Steiner,
  Tucker, Vasudevan, Warden, Wicke, Yu and Zheng}]{abadi2016tensorflow}
Abadi M, Barham P, Chen J, Chen Z, Davis A, Dean J, Devin M, Ghemawat S, Irving
  G, Isard M, Kudlur M, Levenberg J, Monga R, Moore S, Murray DG, Steiner B,
  Tucker P, Vasudevan V, Warden P, Wicke M, Yu Y and Zheng X (2016) Tensorflow:
  A system for large-scale machine learning.
\newblock In: \emph{12th {USENIX} Symposium on Operating Systems Design and
  Implementation ({OSDI} 16)}. pp. 265--283.

\bibitem[{Amato and Wu(1996)}]{amato1996randomized}
Amato NM and Wu Y (1996) A randomized roadmap method for path and manipulation
  planning.
\newblock In: \emph{Proc.\ {IEEE} Intl Conf.\ on Robotics and Automation
  ({ICRA})}, volume~1. IEEE, pp. 113--120.

\bibitem[{Andrieu et~al.(2003)Andrieu, De~Freitas, Doucet and
  Jordan}]{andrieu2003introduction}
Andrieu C, De~Freitas N, Doucet A and Jordan MI (2003) An introduction to mcmc
  for machine learning.
\newblock \emph{Machine learning} 50(1-2): 5--43.

\bibitem[{Arenz et~al.(2018)Arenz, Neumann and Zhong}]{arenz2018efficient}
Arenz O, Neumann G and Zhong M (2018) Efficient gradient-free variational
  inference using policy search.
\newblock In: \emph{Proc.\ Intl Conf.\ on Machine Learning ({ICML})},
  volume~80. PMLR, pp. 234--243.

\bibitem[{Ayvali et~al.(2017)Ayvali, Salman and Choset}]{ayvali2017ergodic}
Ayvali E, Salman H and Choset H (2017) Ergodic coverage in constrained
  environments using stochastic trajectory optimization.
\newblock In: \emph{Proc.\ {IEEE/RSJ} Intl Conf.\ on Intelligent Robots and
  Systems ({IROS})}. IEEE, pp. 5204--5210.

\bibitem[{Bengio et~al.(2014)Bengio, Laufer, Alain and
  Yosinski}]{bengio2014deep}
Bengio Y, Laufer E, Alain G and Yosinski J (2014) Deep generative stochastic
  networks trainable by backprop.
\newblock In: \emph{Proc.\ Intl Conf.\ on Machine Learning ({ICML})}. pp.
  226--234.

\bibitem[{Berenson et~al.(2011)Berenson, Srinivasa and
  Kuffner}]{berenson2011task}
Berenson D, Srinivasa S and Kuffner J (2011) Task space regions: A framework
  for pose-constrained manipulation planning.
\newblock \emph{The International Journal of Robotics Research} 30(12):
  1435--1460.

\bibitem[{Bishop(2006)}]{bishop2006pattern}
Bishop CM (2006) \emph{Pattern Recognition and Machine Learning (Information
  Science and Statistics)}.
\newblock Secaucus, NJ, USA: Springer.

\bibitem[{Bishop et~al.(1998)Bishop, Lawrence, Jaakkola and
  Jordan}]{bishop1998approximating}
Bishop CM, Lawrence ND, Jaakkola T and Jordan MI (1998) Approximating posterior
  distributions in belief networks using mixtures.
\newblock In: \emph{Advances in Neural Information Processing Systems
  ({NIPS})}. pp. 416--422.

\bibitem[{Bohner and Wintz(2011)}]{bohner2011linear}
Bohner M and Wintz N (2011) The linear quadratic tracker on time scales.
\newblock \emph{International Journal of Dynamical Systems and Differential
  Equations} 3(4): 423--447.

\bibitem[{Calinon(2016)}]{calinon2016tutorial}
Calinon S (2016) A tutorial on task-parameterized movement learning and
  retrieval.
\newblock \emph{Intelligent Service Robotics} 9(1): 1--29.

\bibitem[{Calinon and Billard(2009)}]{calinon2009statistical}
Calinon S and Billard A (2009) Statistical learning by imitation of competing
  constraints in joint space and task space.
\newblock \emph{Advanced Robotics} 23(15): 2059--2076.

\bibitem[{Chen et~al.(2014)Chen, Fox and Guestrin}]{chen2014stochastic}
Chen T, Fox E and Guestrin C (2014) Stochastic gradient {H}amiltonian monte
  carlo.
\newblock In: \emph{Proc.\ Intl Conf.\ on Machine Learning ({ICML})}. pp.
  1683--1691.

\bibitem[{Dempster et~al.(1977)Dempster, Laird and Rubin}]{dempster1977maximum}
Dempster AP, Laird NM and Rubin DB (1977) Maximum likelihood from incomplete
  data via the em algorithm.
\newblock \emph{Journal of the Royal Statistical Society: Series B
  (Methodological)} 39(1): 1--22.

\bibitem[{Dinh et~al.(2017)Dinh, Sohl{-}Dickstein and Bengio}]{dinh2017density}
Dinh L, Sohl{-}Dickstein J and Bengio S (2017) Density estimation using {R}eal
  {NVP}.
\newblock In: \emph{Proc.\ Intl Conf.\ on Learning Representation({ICLR})}.

\bibitem[{Dixon(2006)}]{dixon2006speech}
Dixon PR (2006) \emph{Speech Pattern Processing Using Products of Experts}.
\newblock PhD Thesis, University of Birmingham.

\bibitem[{Duane et~al.(1987)Duane, Kennedy, Pendleton and
  Roweth}]{duane1987hybrid}
Duane S, Kennedy AD, Pendleton BJ and Roweth D (1987) Hybrid {M}onte {C}arlo.
\newblock \emph{Physics letters B} 195(2): 216--222.

\bibitem[{Elbanhawi and Simic(2014)}]{elbanhawi2014sampling}
Elbanhawi M and Simic M (2014) Sampling-based robot motion planning: A review.
\newblock \emph{{IEEE} access} 2: 56--77.

\bibitem[{Finn et~al.(2016)Finn, Levine and Abbeel}]{finn2016guided}
Finn C, Levine S and Abbeel P (2016) Guided cost learning: Deep inverse optimal
  control via policy optimization.
\newblock In: \emph{Proc.\ Intl Conf.\ on Machine Learning ({ICML})}. pp.
  49--58.

\bibitem[{Goodfellow et~al.(2014)Goodfellow, Pouget-Abadie, Mirza, Xu,
  Warde-Farley, Ozair, Courville and Bengio}]{goodfellow2014generative}
Goodfellow I, Pouget-Abadie J, Mirza M, Xu B, Warde-Farley D, Ozair S,
  Courville A and Bengio Y (2014) Generative adversarial nets.
\newblock In: \emph{Advances in Neural Information Processing Systems
  ({NIPS})}. pp. 2672--2680.

\bibitem[{Gretton et~al.(2012)Gretton, Borgwardt, Rasch, Sch{\"o}lkopf and
  Smola}]{gretton2012kernel}
Gretton A, Borgwardt KM, Rasch MJ, Sch{\"o}lkopf B and Smola A (2012) A kernel
  two-sample test.
\newblock \emph{Journal of Machine Learning Research} 13(Mar): 723--773.

\bibitem[{Guo et~al.(2016)Guo, Wang, Fan, Broderick and
  Dunson}]{guo2016boosting}
Guo F, Wang X, Fan K, Broderick T and Dunson DB (2016) Boosting variational
  inference.
\newblock \emph{Advances in Neural Information Processing Systems ({NIPS})} .

\bibitem[{Hinton(1999)}]{hinton1999products}
Hinton GE (1999) Products of experts.
\newblock \emph{Proc.\ Intl Conf.\ on Artificial Neural Networks. ({ICANN})} .

\bibitem[{Hinton(2002)}]{hinton2002training}
Hinton GE (2002) Training products of experts by minimizing contrastive
  divergence.
\newblock \emph{Neural computation} 14(8): 1771--1800.

\bibitem[{Hinton et~al.(2006)Hinton, Osindero and Teh}]{hinton2006fast}
Hinton GE, Osindero S and Teh YW (2006) A fast learning algorithm for deep
  belief nets.
\newblock \emph{Neural computation} 18(7): 1527--1554.

\bibitem[{Ichter et~al.(2018)Ichter, Harrison and Pavone}]{ichter2018learning}
Ichter B, Harrison J and Pavone M (2018) Learning sampling distributions for
  robot motion planning.
\newblock In: \emph{Proc.\ {IEEE} Intl Conf.\ on Robotics and Automation
  ({ICRA})}. IEEE, pp. 7087--7094.

\bibitem[{Jaquier et~al.(2020)Jaquier, Rozo, Caldwell and
  Calinon}]{Jaquier20IJRR}
Jaquier N, Rozo L, Caldwell DG and Calinon S (2020) Geometry-aware
  manipulability learning, tracking and transfer.
\newblock \emph{International Journal of Robotics Research ({IJRR})} .

\bibitem[{Jetchev and Toussaint(2011)}]{jetchev2011task}
Jetchev N and Toussaint M (2011) Task space retrieval using inverse feedback
  control.
\newblock In: \emph{Proc.\ Intl Conf.\ on Machine Learning ({ICML})}. pp.
  449--456.

\bibitem[{Johnson et~al.(2017)Johnson, Keller and
  Weintraut}]{johnson2017learning}
Johnson DD, Keller RM and Weintraut N (2017) Learning to create jazz melodies
  using a product of experts.
\newblock In: \emph{International Conference on Computational Creativity}. pp.
  151--158.

\bibitem[{Kalakrishnan et~al.(2013)Kalakrishnan, Pastor, Righetti and
  Schaal}]{kalakrishnan2013learning}
Kalakrishnan M, Pastor P, Righetti L and Schaal S (2013) Learning objective
  functions for manipulation.
\newblock In: \emph{Proc.\ {IEEE} Intl Conf.\ on Robotics and Automation
  ({ICRA})}. IEEE, pp. 1331--1336.

\bibitem[{Khatri and Mardia(1977)}]{khatri1977mises}
Khatri C and Mardia K (1977) The von {M}ises-{F}isher matrix distribution in
  orientation statistics.
\newblock \emph{Journal of the Royal Statistical Society. Series B
  (Methodological)} : 95--106.

\bibitem[{Kingma and Ba(2015)}]{kingma2014adam}
Kingma DP and Ba J (2015) {A}dam: A method for stochastic optimization.
\newblock \emph{Proc.\ Intl Conf.\ on Learning Representation({ICLR})} .

\bibitem[{Kingma and Welling(2013)}]{kingma2013auto}
Kingma DP and Welling M (2013) Auto-encoding variational bayes.
\newblock \emph{Proc.\ Intl Conf.\ on Learning Representation({ICLR})} .

\bibitem[{Kopicki et~al.(2017)Kopicki, Zurek, Stolkin, Moerwald and
  Wyatt}]{kopicki2017learning}
Kopicki M, Zurek S, Stolkin R, Moerwald T and Wyatt JL (2017) Learning modular
  and transferable forward models of the motions of push manipulated objects.
\newblock \emph{Autonomous Robots} 41(5): 1061--1082.

\bibitem[{Kume et~al.(2013)Kume, Preston and Wood}]{kume2013saddlepoint}
Kume A, Preston S and Wood AT (2013) Saddlepoint approximations for the
  normalizing constant of {F}isher--{B}ingham distributions on products of
  spheres and stiefel manifolds.
\newblock \emph{Biometrika} 100(4): 971--984.

\bibitem[{Lehner and Albu-Sch{\"a}ffer(2017)}]{lehner2017repetition}
Lehner P and Albu-Sch{\"a}ffer A (2017) Repetition sampling for efficiently
  planning similar constrained manipulation tasks.
\newblock In: \emph{Proc.\ {IEEE/RSJ} Intl Conf.\ on Intelligent Robots and
  Systems ({IROS})}. IEEE, pp. 2851--2856.

\bibitem[{Li and Todorov(2004)}]{li2004iterative}
Li W and Todorov E (2004) Iterative linear quadratic regulator design for
  nonlinear biological movement systems.
\newblock In: \emph{ICINCO}. pp. 222--229.

\bibitem[{Lin et~al.(2015)Lin, Howard and Vijayakumar}]{Lin15}
Lin HC, Howard M and Vijayakumar S (2015) Learning null space projections.
\newblock In: \emph{Proc.\ {IEEE} Intl Conf.\ on Robotics and Automation
  ({ICRA})}. Seattle, WA, USA, pp. 2613--2619.

\bibitem[{Lober et~al.(2015)Lober, Padois and Sigaud}]{lober2015variance}
Lober R, Padois V and Sigaud O (2015) Variance modulated task prioritization in
  whole-body control.
\newblock In: \emph{Proc.\ {IEEE/RSJ} Intl Conf.\ on Intelligent Robots and
  Systems ({IROS})}. IEEE, pp. 3944--3949.

\bibitem[{Mathew and Mezi{\'c}(2011)}]{mathew2011metrics}
Mathew G and Mezi{\'c} I (2011) Metrics for ergodicity and design of ergodic
  dynamics for multi-agent systems.
\newblock \emph{Physica D: Nonlinear Phenomena} 240(4-5): 432--442.

\bibitem[{Miller et~al.(2017)Miller, Foti and Adams}]{miller2017variational}
Miller AC, Foti NJ and Adams RP (2017) Variational boosting: Iteratively
  refining posterior approximations.
\newblock In: \emph{Proc.\ Intl Conf.\ on Machine Learning ({ICML})}. pp.
  2420--2429.

\bibitem[{M{\"u}hlig et~al.(2009)M{\"u}hlig, Gienger, Steil and
  Goerick}]{muhlig2009automatic}
M{\"u}hlig M, Gienger M, Steil JJ and Goerick C (2009) Automatic selection of
  task spaces for imitation learning.
\newblock In: \emph{Proc.\ {IEEE/RSJ} Intl Conf.\ on Intelligent Robots and
  Systems ({IROS})}. IEEE, pp. 4996--5002.

\bibitem[{Nakamura et~al.(1987)Nakamura, Hanafusa and
  Yoshikawa}]{nakamura1987task}
Nakamura Y, Hanafusa H and Yoshikawa T (1987) Task-priority based redundancy
  control of robot manipulators.
\newblock \emph{The International Journal of Robotics Research} 6(2): 3--15.

\bibitem[{Ng and Russell(2000)}]{ng2000algorithms}
Ng AY and Russell SJ (2000) Algorithms for inverse reinforcement learning.
\newblock In: \emph{Proc.\ Intl Conf.\ on Machine Learning ({ICML})}, volume~1.
  pp. 663--670.

\bibitem[{Niekum et~al.(2015)Niekum, Osentoski, Konidaris, Chitta, Marthi and
  Barto}]{niekum2015learning}
Niekum S, Osentoski S, Konidaris G, Chitta S, Marthi B and Barto AG (2015)
  Learning grounded finite-state representations from unstructured
  demonstrations.
\newblock \emph{The International Journal of Robotics Research} 34(2):
  131--157.

\bibitem[{Opper and Archambeau(2009)}]{opper2009variational}
Opper M and Archambeau C (2009) The variational {G}aussian approximation
  revisited.
\newblock \emph{Neural computation} 21(3): 786--792.

\bibitem[{Paraschos et~al.(2013)Paraschos, Daniel, Peters and
  Neumann}]{paraschos2013probabilistic}
Paraschos A, Daniel C, Peters JR and Neumann G (2013) Probabilistic movement
  primitives.
\newblock In: \emph{Advances in Neural Information Processing Systems
  ({NIPS})}. pp. 2616--2624.

\bibitem[{Paraschos et~al.(2017)Paraschos, Lioutikov, Peters and
  Neumann}]{paraschos2017probabilistic}
Paraschos A, Lioutikov R, Peters J and Neumann G (2017) Probabilistic
  prioritization of movement primitives.
\newblock \emph{IEEE Robotics and Automation Letters} 2(4): 2294--2301.

\bibitem[{Parzen(1962)}]{parzen1962estimation}
Parzen E (1962) On estimation of a probability density function and mode.
\newblock \emph{The annals of mathematical statistics} 33(3): 1065--1076.

\bibitem[{Pini(1991)}]{pini1991invexity}
Pini R (1991) Invexity and generalized convexity.
\newblock \emph{Optimization} 22(4): 513--525.

\bibitem[{Pradalier et~al.(2003)Pradalier, Colas and
  Bessiere}]{pradalier2003expressing}
Pradalier C, Colas F and Bessiere P (2003) Expressing bayesian fusion as a
  product of distributions: application in robotics.
\newblock In: \emph{Proc.\ {IEEE/RSJ} Intl Conf.\ on Intelligent Robots and
  Systems ({IROS})}. pp. 1851--1856.

\bibitem[{Raiola et~al.(2015)Raiola, Lamy and Stulp}]{raiola2015co}
Raiola G, Lamy X and Stulp F (2015) Co-manipulation with multiple probabilistic
  virtual guides.
\newblock In: \emph{Proc.\ {IEEE/RSJ} Intl Conf.\ on Intelligent Robots and
  Systems ({IROS})}. IEEE, pp. 7--13.

\bibitem[{Ranganath et~al.(2014)Ranganath, Gerrish and
  Blei}]{ranganath2014black}
Ranganath R, Gerrish S and Blei D (2014) Black box variational inference.
\newblock In: \emph{Artificial Intelligence and Statistics}. pp. 814--822.

\bibitem[{Rezende and Mohamed(2015)}]{rezende2015variational}
Rezende DJ and Mohamed S (2015) Variational inference with normalizing flows.
\newblock In: \emph{Proc.\ Intl Conf.\ on Machine Learning ({ICML})}. JMLR.
  org, pp. 1530--1538.

\bibitem[{Salakhutdinov and Hinton(2009)}]{salakhutdinov2009deep}
Salakhutdinov R and Hinton G (2009) Deep {B}oltzmann machines.
\newblock In: \emph{Artificial intelligence and statistics}. pp. 448--455.

\bibitem[{Salimans and Knowles(2013)}]{salimans2013fixed}
Salimans T and Knowles DA (2013) Fixed-form variational posterior approximation
  through stochastic linear regression.
\newblock \emph{Bayesian Analysis} 8(4): 837--882.

\bibitem[{Silv{\'e}rio et~al.(2018)Silv{\'e}rio, Calinon, Rozo and
  Caldwell}]{silverio2018learning}
Silv{\'e}rio J, Calinon S, Rozo L and Caldwell DG (2018) Learning task
  priorities from demonstrations.
\newblock \emph{IEEE Transactions on Robotics} 35(1): 78--94.

\bibitem[{Sminchisescu et~al.(2003)Sminchisescu, Welling and
  Hinton}]{sminchisescu2003mode}
Sminchisescu C, Welling M and Hinton G (2003) A mode-hopping {MCMC} sampler.
\newblock Technical report, University of Toronto.

\bibitem[{Sohn et~al.(2015)Sohn, Lee and Yan}]{sohn2015learning}
Sohn K, Lee H and Yan X (2015) Learning structured output representation using
  deep conditional generative models.
\newblock In: \emph{Advances in Neural Information Processing Systems
  ({NIPS})}. pp. 3483--3491.

\bibitem[{Towell et~al.(2010)Towell, Howard and Vijayakumar}]{Towell10}
Towell C, Howard M and Vijayakumar S (2010) Learning nullspace policies.
\newblock In: \emph{Proc.\ {IEEE/RSJ} Intl Conf.\ on Intelligent Robots and
  Systems ({IROS})}. Taipei, Taiwan, pp. 241--248.

\bibitem[{Voss et~al.(2017)Voss, Moll and Kavraki}]{voss2017atlas+}
Voss C, Moll M and Kavraki LE (2017) {A}tlas+ x: Sampling-based planners on
  constraint manifolds.
\newblock Technical report, RICE computer science.

\bibitem[{Wainwright and Jordan(2008)}]{wainwright2008graphical}
Wainwright MJ and Jordan MI (2008) Graphical models, exponential families, and
  variational inference.
\newblock \emph{Foundations and Trends in Machine Learning} 1(1--2): 1--305.

\bibitem[{Welling et~al.(2004)Welling, Mnih and Hinton}]{welling2004wormholes}
Welling M, Mnih A and Hinton GE (2004) Wormholes improve contrastive
  divergence.
\newblock In: \emph{Advances in Neural Information Processing Systems
  ({NIPS})}. pp. 417--424.

\bibitem[{Yang et~al.(2015)Yang, Ivan and Vijayakumar}]{yang2015real}
Yang Y, Ivan V and Vijayakumar S (2015) Real-time motion adaptation using
  relative distance space representation.
\newblock In: \emph{Proc.\ Intl Conf.\ on Advanced Robotics ({ICAR})}. IEEE,
  pp. 21--27.

\bibitem[{Yoshikawa(1985)}]{yoshikawa1985manipulability}
Yoshikawa T (1985) Manipulability of robotic mechanisms.
\newblock \emph{The international journal of Robotics Research} 4(2): 3--9.

\bibitem[{Zeestraten et~al.(2017)Zeestraten, Havoutis, Silverio, Calinon and
  Caldwell}]{zeestraten2017approach}
Zeestraten MJ, Havoutis I, Silverio J, Calinon S and Caldwell DG (2017) An
  approach for imitation learning on riemannian manifolds.
\newblock \emph{IEEE Robotics and Automation Letters} 2(3): 1240--1247.

\bibitem[{Zen et~al.(2012)Zen, Gales, Nankaku and Tokuda}]{zen2012product}
Zen H, Gales MJ, Nankaku Y and Tokuda K (2012) Product of experts for
  statistical parametric speech synthesis.
\newblock \emph{IEEE Transactions on Audio, Speech, and Language Processing}
  20(3): 794--805.

\bibitem[{Zhang et~al.(2013)Zhang, Hauser and Luo}]{zhang2013unbiased}
Zhang Y, Hauser K and Luo J (2013) Unbiased, scalable sampling of closed
  kinematic chains.
\newblock In: \emph{Proc.\ {IEEE} Intl Conf.\ on Robotics and Automation
  ({ICRA})}. IEEE, pp. 2459--2464.

\bibitem[{Ziebart et~al.(2008)Ziebart, Maas, Bagnell and
  Dey}]{ziebart2008maximum}
Ziebart BD, Maas AL, Bagnell JA and Dey AK (2008) Maximum entropy inverse
  reinforcement learning.
\newblock In: \emph{AAAI}, volume~8. Chicago, IL, USA, pp. 1433--1438.

\end{thebibliography}
	
\end{document}